\pdfoutput=1

\documentclass[11pt]{article}

\PassOptionsToPackage{dvipsnames,table}{xcolor}

\usepackage[final]{acl}
\usepackage{amssymb}
\usepackage{pifont}

\usepackage{times}
\usepackage{latexsym}
\usepackage{svg}
\usepackage{listings}
\usepackage[most]{tcolorbox}
\tcbuselibrary{listings,breakable,skins}

\lstdefinestyle{promptstyle}{
  basicstyle=\ttfamily\footnotesize,
  columns=fullflexible,
  breaklines=true,
  breakatwhitespace=true,
  keepspaces=true,
  showstringspaces=false,
  upquote=true
}

\newtcblisting{promptbox}[1][]{%
  listing only,
  listing options={style=promptstyle},
  breakable,
  enhanced,
  colback=white,
  colframe=black!40,
  coltitle=black,        
  boxrule=0.6pt,
  arc=2pt,
  left=6pt,right=6pt,top=6pt,bottom=6pt,
  before skip=0.8em,
  after skip=0.8em,
  fonttitle=\bfseries,
  colbacktitle=black!5,
  #1
}

\usepackage[T1]{fontenc}

\usepackage[utf8]{inputenc}

\usepackage{microtype}

\usepackage{threeparttable}
\usepackage{booktabs}
\usepackage{enumitem}
\usepackage{hyperref}
\usepackage{booktabs}
\usepackage{pifont}
\usepackage{float}
\usepackage{graphicx}

\usepackage{booktabs}
\usepackage{tabularx}
\usepackage{threeparttable}
\usepackage{graphicx}
\usepackage{subcaption}

\usepackage{booktabs}
\usepackage{pifont}
\usepackage{array}          
\usepackage{booktabs}
\usepackage{tabularx}
\usepackage{array}
\usepackage{footnotehyper}
\makesavenoteenv{table*}

\usepackage[most]{tcolorbox}
\usepackage{enumitem}

\definecolor{webblue}{RGB}{76, 111, 255} 
\definecolor{webbg}{RGB}{248, 249, 253}  
\definecolor{webtext}{RGB}{50, 50, 80}
\definecolor{slotcolor}{RGB}{220, 50, 50} 
\usepackage{multirow}

\tcbset{
    webpage/.style={
        colframe=gray!30,
        colback=white,
        fonttitle=\sffamily\bfseries,
        coltitle=black,
        sharp corners=south,
        boxrule=0.5mm,
        drop shadow medium
    },
    webbutton/.style={
        colback=webblue,
        colframe=webblue,
        coltext=white,
        halign=center,
        valign=center,
        fontupper=\sffamily\bfseries\small,
        arc=3mm,
        auto outer arc,
        boxsep=2pt,
        width=4cm
    }
}
\author{
Alaa Elsetohy$^{1,*}$, 
Sama Hadhoud$^{1}$, 
Haryo Akbarianto Wibowo$^{1}$, 
Chenxi Whitehouse$^{2}$, \\
\textbf{Genta Indra Winata$^{3}$, 
Fajri Koto$^{1}$, 
Alham Fikri Aji$^{1,*}$} \\
[0.5em]
$^{1}$MBZUAI \quad
$^{2}$Meta \quad
$^{3}$Capital One \\
[0.5em]
\texttt{\{alaa.elsetohy,alham.fikri\}@mbzuai.ac.ae} \\
[0.3em]
\textit{$^{*}$Corresponding authors}
}

\newcommand{\dataname}{\texttt{Macaron}}

\NewDocumentCommand{\inlineimage}{O{0.5} m}{%
  \raisebox{-0.2\baselineskip}{\includegraphics[height=#1\baselineskip]{#2}}%
  \hspace{-3pt}%
}

\DeclareRobustCommand{\logo}{%
  \raisebox{-0.2\baselineskip}{%
    \includegraphics[height=0.9\baselineskip]{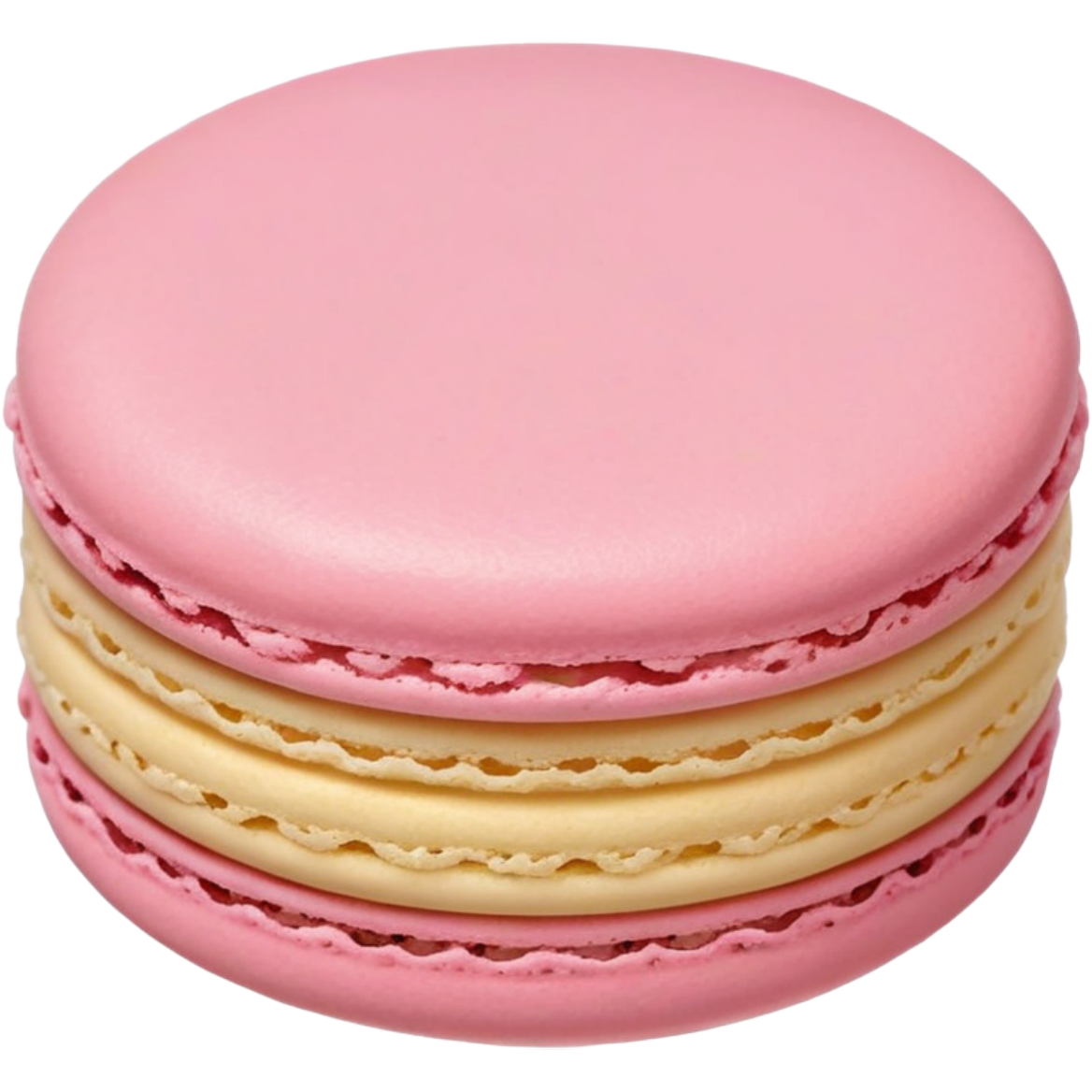}%
  }%
  \hspace{-3pt}%
}
\title{\logo\ \dataname: Controlled, Human-Written Benchmark for Multilingual and Multicultural Reasoning via Template-Filling}

\begin{document}

\maketitle


\begin{abstract}
Multilingual benchmarks rarely test reasoning over culturally grounded premises: translated datasets keep English-centric scenarios, while culture-first datasets often lack control over the reasoning required. We propose \dataname, a template-first benchmark that factorizes reasoning type and cultural aspect across question languages. Using 100 language-agnostic templates that cover 7 reasoning types, 22 cultural aspects, native annotators create scenario-aligned English and local-language multiple-choice questions and systematically derived True/False questions. \dataname\ contains 11{,}862 instances spanning 20 countries/cultural contexts, 10 scripts, and 20 languages and dialects (including low-resource ones like Amharic, Yoruba, Zulu, Kyrgyz, and some Arabic dialects). In zero-shot evaluation of 21 multilingual LLMs, reasoning-mode models achieve the strongest performance (80.8\% overall) and near-parity between English and local languages ($\Delta$MC\,=\,$-$1.3\%), while open-weight models degrade substantially in local languages ($\Delta$MC\,=\,$-$6.8\%) and often approach chance on T/F tasks. Culture-grounded mathematical and counting templates are consistently the hardest. The data can be accessed here \url{https://huggingface.co/datasets/AlaaAhmed2444/Macaron}.

\end{abstract}

\section{Introduction}
With the growing progress of multilingual LLMs, benchmarking them across languages and cultures is equally important. Existing benchmarks pursue this along two complementary directions, each with its own blind spot. 
Translation-parallel benchmarks enable controlled cross-lingual comparison but inherit English-centric scenarios \citep{conneau2018xnli,ponti2020xcopa,artetxe2019xquad,lin2021common,globalmmlu,mmluprox}. Culture-first benchmarks provide locally salient content but lack explicit control over reasoning skills, and scaling them requires new questions from scratch for each culture, which drifts in scope and difficulty \citep{chiu2025culturalbench,myung2024blend,sadallah2025arabculture,hasan2025nativqa,wibowo-etal-2024-copal,romero2024cvqaculturallydiversemultilingualvisual}.

We propose \dataname, a template-first benchmark for multilingual multicultural reasoning that addresses these issues by separating, by construction, the three factors that prior evaluations conflate: the reasoning skill a question requires, the cultural aspect it probes, and the language in which it is presented.
We design 100 language-agnostic templates tagged with 7 reasoning types and 22 cultural aspects, and recruit native annotators to instantiate them with culturally grounded content and produce scenario-aligned English--local versions. Because templates are reusable, extending \dataname\ to new cultures primarily requires instantiation and translation of the same template set, keeping structure and targeted reasoning stable.

From 1,977 bilingual MC scenarios spanning 20 languages and dialects, we derive aligned True/False variants yielding 11,862 evaluation instances. Evaluations across 21 multilingual LLMs show that reasoning-mode models are strongest (80.8\%) and nearly language-robust (Avg. $\Delta$MC = -1.3), while open-weight models lag (58.0\%) and degrade more in local languages (Avg.$\Delta$MC = -6.8), with culture-grounded mathematical and counting questions consistently hardest.
\paragraph{Our contributions are:}
\begin{enumerate}[itemsep=2pt, topsep=2pt, parsep=0pt]
    \item A template-first framework that factorizes \emph{reasoning type} and \emph{cultural aspect} for controlled multilingual cultural reasoning.
    \item \dataname: a scenario-aligned bilingual benchmark with MCQ and derived T/F variants across 20 cultural contexts.
    \item An evaluation of 21 multilingual LLMs with analyses across languages, reasoning categories, and cultural aspects.
\end{enumerate}

\begin{table*}[ht!]
\centering
\resizebox{\textwidth}{!}{%
\setlength{\tabcolsep}{4pt}
\renewcommand{\arraystretch}{1.15}
\begin{tabular}{@{}llcccccccccc@{}}
\toprule
\textbf{Benchmark} & \textbf{Format} & \textbf{\#Eval} & \textbf{\#Langs \&} & \textbf{\#Cultures/} & \textbf{Culture} & \textbf{Native} & \textbf{Bilingual} & \textbf{Template} & \textbf{Reasoning} & \textbf{Culture} & \textbf{Reasoning} \\
 & & \textbf{items} & \textbf{Dialects} & \textbf{Regions} & \textbf{grounded} & \textbf{authored} & \textbf{aligned} & \textbf{based} & \textbf{taxonomy} & \textbf{taxonomy} & $\times$ \textbf{Culture} \\
\midrule
\multicolumn{12}{l}{\textit{Translation-parallel benchmarks}} \\
\midrule
XNLI \cite{conneau2018xnli}                  & NLI       & 5k/lang    & 15 & —  & \texttimes & \texttimes & \checkmark & \texttimes & \texttimes & \texttimes & \texttimes \\
XCOPA \cite{ponti2020xcopa}                  & MCQ       & 500/lang   & 11 & —  & \texttimes & \texttimes & \checkmark & \texttimes & \texttimes & \texttimes & \texttimes \\
Global-MMLU \cite{globalmmlu}               & MCQ       & 14k/lang   & 42 & 42 & \texttimes & \texttimes & \checkmark & \texttimes & \texttimes & \texttimes & \texttimes \\
MMLU-ProX \cite{mmluprox}                   & MCQ       & 11.8k/lang & 29 & 29 & \texttimes & \texttimes & \checkmark & \texttimes & \texttimes & \texttimes & \texttimes \\
\midrule
\multicolumn{12}{l}{\textit{Culture-grounded benchmarks}} \\
\midrule
INCLUDE \cite{romanou2025include}            & MCQ       & 197.2k     & 44 & 52 & \checkmark & \checkmark & \texttimes & \texttimes & \texttimes & \texttimes & \texttimes \\
MILU \cite{verma2024milu}                    & MCQ       & 79.6k      & 11 & 11 & \checkmark & \checkmark & \texttimes & \texttimes & \texttimes & \texttimes & \texttimes \\
BLEnD \cite{myung2024blend}                  & MCQ+Free  & 52.6k      & 13 & 16 & \checkmark & \checkmark & \checkmark & \texttimes & \texttimes & \texttimes & \texttimes \\
CulturalBench \cite{chiu2025culturalbench}   & MCQ+T/F   & 1.7k       &  1 & 45 & \checkmark & \checkmark & \texttimes & \texttimes & \texttimes & \checkmark & \texttimes \\
ArabCulture \cite{sadallah2025arabculture}   & MCQ       & 3.5k       &  1 & 13 & \checkmark & \checkmark & \texttimes & \texttimes & \texttimes & \texttimes & \texttimes \\
NativQA \cite{hasan2025nativqa}              & QA        & 64k        &  7 &  9 & \checkmark & \checkmark & \texttimes & \texttimes & \texttimes & \texttimes & \texttimes \\
WorldCuisines \cite{winata2025worldcuisines} & MCQ+QA    & $\sim$1M   & 30 & —  & \checkmark & \checkmark & \checkmark & \checkmark & \texttimes & \texttimes & \texttimes \\
MultiNRC \cite{fabbri2025multinrcchallengingnativemultilingual}           & Free-text & 1k         &  3 &  3 & \checkmark & \checkmark & \checkmark & \texttimes & \checkmark\textsuperscript{*} & \texttimes & \texttimes \\
\midrule
\multicolumn{12}{l}{\textit{Ours}} \\
\midrule
\textbf{Macaron}                             & \textbf{MCQ+T/F} & $\sim$\textbf{12k} & \textbf{20} & \textbf{20} & \checkmark & \checkmark & \checkmark & \checkmark & \checkmark & \checkmark & \checkmark \\
\bottomrule
\end{tabular}%
}
\caption{Comparison of Macaron with related multilingual and multicultural benchmarks across key design properties. \textbf{\#Cultures/Regions}: number of distinct cultural contexts covered; dashes~(—) indicate translation-parallel benchmarks with no distinct cultural grounding, and WorldCuisines where the cultural unit is cuisine rather than a discrete cultural context. \textbf{Bilingual aligned}: English vs Local language. \textbf{Reasoning $\times$ Culture}: reasoning type and cultural aspect are jointly tagged at the item level. \textsuperscript{*}MultiNRC provides reasoning categories but at a much coarser granularity than Macaron's seven-type taxonomy.}
\label{tab:rw_compare}
\end{table*}

\section{Related Work}
\label{sec:related}

\paragraph{Reasoning and diagnostic evaluation.}English-first benchmarks cover commonsense and plausibility reasoning (HellaSwag, WinoGrande, ARC, CROW)
\citep{zellers2019hellaswagmachinereallyfinish,sakaguchi2019winograndeadversarialwinogradschema,clark2018arc,ismayilzada2023crow}
and exam-style reasoning (BIG-bench, MMLU, MMLU-Pro) \citep{srivastava2022bigbench,mmlu,mmlupro}, with harder diagnostic subsets such as BBH
\citep{BBH}. Controlled-structure datasets such as bAbI and CLUTRR \citep{bAbI,clutrr} motivate template-controlled evaluation. While these resources provide
strong reasoning diagnostics, they are not designed to evaluate reasoning under culturally grounded premises.

\paragraph{Translation-parallel multilingual evaluation.}A common multilingual strategy is to translate English-source datasets to many languages, enabling controlled cross-lingual comparison but inheriting source
framing and assumptions. Examples include XNLI \citep{conneau2018xnli}, XCOPA \citep{ponti2020xcopa}, XQuAD \citep{artetxe2019xquad}, and X--CSR
\citep{lin2021common}. Global-MMLU and MMLU-ProX expand exam-style evaluation across languages and scripts while keeping instances parallel
\citep{globalmmlu,mmluprox}. M3Exam extends this to a multilingual, multimodal, multilevel setting using real exam questions from multiple countries \citep{zhang2023m3exammultilingualmultimodalmultilevel}, but like other translation-parallel benchmarks it does not systematically control reasoning type or pair English and local-language items over identical cultural scenarios.

\paragraph{Culture-grounded and regional benchmarks.}
Regional-sourcing benchmarks such as INCLUDE and MILU draw from local exams or region-specific materials \citep{romanou2025include,verma2024milu}. Culture-first
and native-query resources such as CulturalBench, BLEnD, ArabCulture, and NativQA/MultiNativQA emphasize locally salient content
\citep{chiu2025culturalbench,myung2024blend,sadallah2025arabculture,hasan2025nativqa}. NormAd is complementary for norm and etiquette judgments, but it is not a bilingual, scenario-aligned reasoning test
\citep{normad2024}. MultiNRC adds explicit reasoning categories alongside native-authored questions, but covers fewer languages and does not systematically
cross reasoning types with cultural domains at scale \citep{fabbri2025multinrcchallengingnativemultilingual}.
\citet{notequal} show that multilingual capability and cultural alignment are distinct dimensions in LLMs, motivating benchmarks that explicitly separate the two, a goal our template-first design directly addresses.
\citet{veselovsky2025localizedculturalknowledgeconserved} show from an interpretability perspective that localized cultural knowledge is internally represented and controllable in LLMs, providing a complementary mechanistic view to our behavioral evaluation.

\paragraph{Disentangling language and culture.}
Closest in motivation to ours, \citet{ying2025disentanglinglanguagecultureevaluating} propose a post-hoc framework for decomposing model scores along linguistic and cultural axes, but applied to pre-existing datasets, whereas Macaron enforces this separation by construction through native-annotated, scenario-aligned bilingual items.


Table~\ref{tab:rw_compare} compares our benchmark with previous work. Macaron is the only benchmark to simultaneously satisfy all seven properties: culture-grounded, native-authored, bilingual-aligned, template-based, with a reasoning taxonomy, a culture taxonomy, and their joint coverage at the item level.

\section{Data Curation}
\label{sec:method}

\begin{figure*}[t!]
    \centering
    \includegraphics[width=\textwidth]{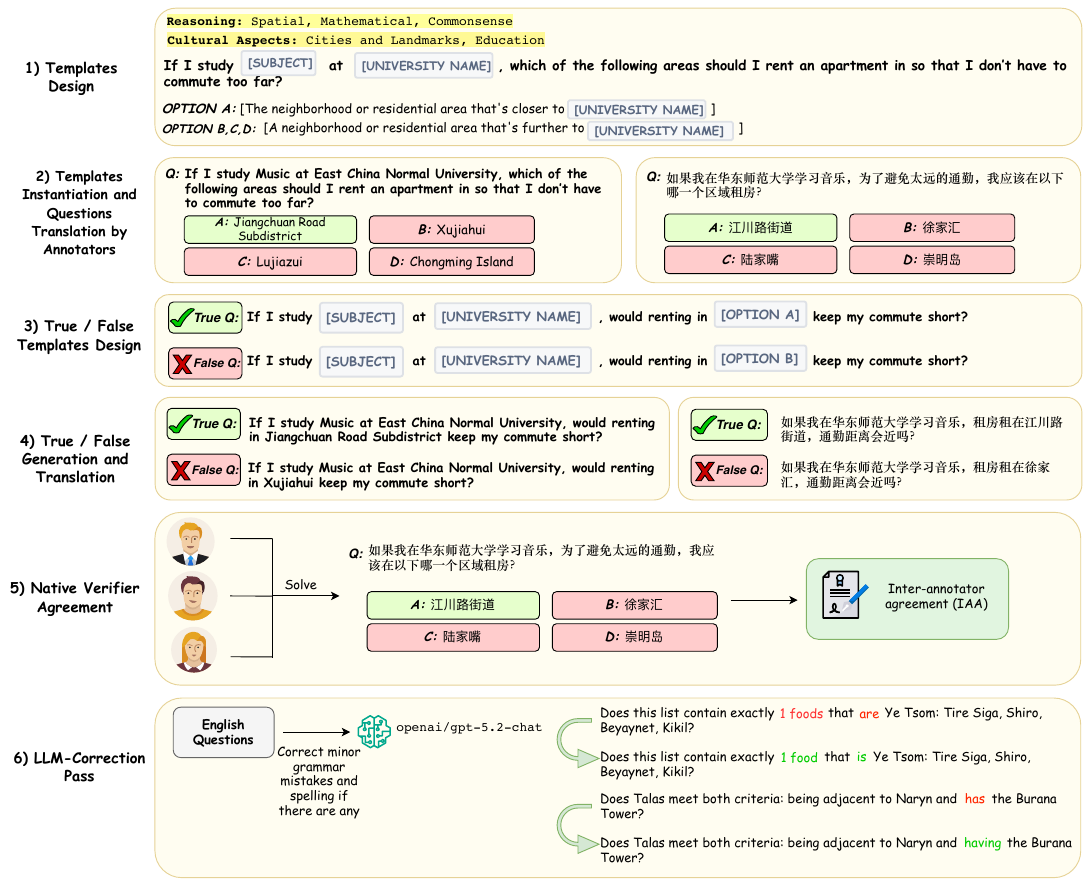}
    \caption{\textbf{{\dataname} Curation Pipeline.} We first design language-agnostic templates tagged with reasoning categories and cultural aspects. Native annotators instantiate each template with culturally grounded content to produce scenario-aligned English MCQs and translate them into local languages. From each MCQ, we derive aligned True/False statements by instantiating the template’s binary child forms with the correct option (True) and a distractor (False), and translate these statements into the local language. Finally, we run an LLM proofreading pass to correct minor spelling/grammar issues while preserving the original meaning and leaving all local-language text unchanged.}
    \label{fig:pipeline}
\end{figure*}

Our goal is to evaluate \emph{multilingual, multicultural reasoning} in a controlled setting.
We operationalize this as (i) multiple-choice question answering and (ii) binary True/False verification over the
\emph{same} culturally grounded scenarios as shown in Figure \ref{fig:pipeline}.
The benchmark is designed to help disentangle three factors that are often confounded in multilingual evaluation:
\emph{language} (English vs.\ local input), \emph{cultural grounding},
and \emph{reasoning} (the inference required to answer).

\subsection{Task Definition}
\label{sec:task}

Let $\mathcal{L}$ denote the set of local languages in the benchmark and $\mathcal{C}_{\text{ctx}}$ the set of cultural contexts
(countries or regions). We construct \emph{base annotations} as bilingual, culturally aligned multiple-choice items.
A base annotation is a tuple
\[
  a = \big(q^{\text{en}}, A^{\text{en}}, q^{\ell}, A^{\ell}, R_a, C_a\big),
\]
where $q^{\text{en}}$ and $q^{\ell}$ are the English and local-language question texts, $A^{\text{en}}$ and $A^{\ell}$ are the corresponding
sets of four answer options with exactly one correct choice, and $\ell \in \mathcal{L}$ is the local language.
We treat both reasoning and culture as explicit (potentially multi-label) metadata:
$R_a \subseteq \mathcal{R}$ is the set of reasoning types targeted by the item, and
$C_a \subseteq \mathcal{C}_{\text{aspect}}$ is the set of cultural aspects it probes.

From each base annotation, we derive four additional binary instances (True/False in English and in local language;
Section~\ref{sec:truefalse}), yielding \textbf{six aligned evaluation instances per cultural scenario}.

\subsection{Reasoning and Cultural Taxonomies}
\label{sec:taxonomies}

\paragraph{Reasoning types.}
We define a taxonomy of seven reasoning types that commonly arise in culturally grounded questions:
\emph{mathematical} (numerical computation and comparison),
\emph{commonsense} (everyday plausibility and typical situations),
\emph{causal} (cause--effect relations),
\emph{temporal} (time, order, calendars),
\emph{logical} (deduction, implication, and analogy),
\emph{spatial} (geographic and spatial relations),
and \emph{multi-hop} (composition of two or more inference steps, e.g., symbol $\rightarrow$ religion $\rightarrow$ practice).
Templates may be tagged with multiple reasoning types when solving requires more than one skill.

\paragraph{Cultural aspects.}
Following \citet{adilazuarda2024towards}, we treat culture as a broad, multifaceted concept that resists direct definition, and instead operationalize it through \emph{proxies of culture}, specifically 22 \emph{semantic proxies} that span the domains of everyday life a community shares.
These aspects serve as our cultural taxonomy:
\emph{Agriculture},
\emph{Brands and Commerce},
\emph{Cities and Landmarks},
\emph{Death and Funerals},
\emph{Education},
\emph{Events and Festivals},
\emph{Famous People},
\emph{Fashion and Media},
\emph{Folklore and Folktales},
\emph{Food and Cuisine},
\emph{Language and Communication},
\emph{Literature and Written works},
\emph{Music and Art},
\emph{Naming},
\emph{Objects and Units},
\emph{Politics and Governance},
\emph{Relationships},
\emph{Social Customs},
\emph{Sports},
\emph{Time},
\emph{Transportation},
and \emph{socio-religious aspects of life}.
Each template is associated with at least one aspect, and some span multiple aspects.
We provide an example template for each aspect in Appendix~\ref{app:templates} (Table~\ref{tab:cultural_aspect_example_templates}).

\subsection{Template Framework}
\label{sec:templates}

To systematically cover the reasoning$\times$culture space, we design a set of 100 language-agnostic templates.
Each template specifies:
\begin{itemize}  
    \item a question skeleton with typed slots (e.g., \texttt{[COUNTRY]}, \texttt{[PERSON]}, \texttt{[FOOD1]}),
    including constraints on valid slot values;
    \item metadata tags indicating the targeted reasoning type(s) and cultural aspect(s);
    \item an expected output format (four-option multiple choice with exactly one correct answer).
\end{itemize}

Templates are authored and iteratively refined by the dataset creators.
During refinement, we remove culturally insensitive or non-portable designs, tighten slot constraints to prevent ambiguity,
and ensure that the intended reasoning path is stable across cultural contexts.
Each template also includes a True/False variant (Section~\ref{sec:truefalse}).

\subsection{Bilingual Annotation Pipeline}
\label{sec:annotation}

\paragraph{Annotators and onboarding}
\label{sec:annotators}

For each cultural context, we recruit two annotators via the Upwork freelancing platform.
Annotators are native speakers of the target local language and have substantial lived experience in the target context
(e.g., having grown up and/or currently residing there).
During onboarding, annotators receive some guidelines and complete a small pilot set of templates with feedback from the dataset creators.
Annotator demographics (gender, age, education, residence duration) are reported in Appendix~\ref{app:annotators}; screenshots of the annotation platform are shown in Appendix~\ref{app:annotation}.

The guidelines emphasize:
\begin{enumerate}
    \item \textbf{Cultural representativeness within templates.} Instantiate each assigned template with content that is locally appropriate
    and commonly recognizable to members of the target culture, aiming for diversity across everyday institutions and practices.
    \item \textbf{Avoiding both stereotypes and obscure trivia.} Prefer culturally salient, everyday knowledge rather than tourist facts or
    internet stereotypes, while avoiding niche or hard-to-verify trivia that most locals would not know.
    \item \textbf{Plausible within-context distractors.} Write distractors that are plausible \emph{within the same cultural sphere} so items
    cannot be solved by eliminating obviously foreign options; ensure exactly one correct answer.
    \item \textbf{Non-applicability and ambiguity flags.} Flag and skip templates that do not meaningfully apply to your cultural context.
\end{enumerate}

\paragraph{Step 1: English multiple-choice instantiation}
\label{sec:step1}

Annotators are assigned a subset of templates. For each template, they:
\begin{enumerate}[itemsep=2pt, topsep=2pt, parsep=0pt]
    \item fill required slots with culturally appropriate content based on lived experience and commonly shared local knowledge;
    \item provide one correct option and three plausible distractors, ensuring exactly one option is correct for the target context.
\end{enumerate}
\paragraph{Step 2: local-language translation}
\label{sec:step2}

After completing the English version, the same annotator translates the question and its options into their native language.
For each base item, they produce local-language text $q^{\ell}$ and $A^{\ell}$ under the constraints that the translation must:
\begin{itemize}[itemsep=2pt, topsep=2pt, parsep=0pt]
    \item preserve the underlying cultural content (same dish, institution, practice, person, etc.);
    \item preserve the reasoning structure and difficulty;
    \item allow only light adaptations for naturalness when needed, as long as the English and local versions still align.
\end{itemize}

Each item inherits the reasoning and cultural-aspect tags from the template metadata, yielding a bilingual base annotation
$a = \big(q^{\text{en}}, A^{\text{en}}, q^{\ell}, A^{\ell}, R_a, C_a\big)$.

\subsection{True/False Variant Generation}
\label{sec:truefalse}

Starting from each base annotation, we construct four additional binary instances: True and False in English and in the local language.
Concretely, for each base item we instantiate the template's binary child forms: a \textbf{True} variant by inserting the \emph{correct} option, yielding a statement whose correct label is True; and a \textbf{False} variant by inserting a carefully chosen \emph{distractor} option, yielding a statement whose correct label is False.

We generate these binary instances in both English and the local language, maintaining scenario-level alignment.
The True and False variants share the same cultural scenario and reasoning requirements as the parent multiple-choice item.
Thus, each base annotation yields six aligned instances: MC--EN, MC--L, T--EN, T--L, F--EN, and F--L.

\subsection{Quality Control}
\label{sec:qc}

To ensure cultural correctness, linguistic clarity, and consistency across annotators and cultural contexts, we apply a combination of human review and automated quality-control procedures.

\paragraph{Native verifier agreement.}
We conduct an independent verification step to assess the factual correctness of cultural content and the quality of local-language translations.
For each cultural context, we recruit \textbf{three independent native verifiers}, distinct from the original annotators, and ask them to answer the translated MCQ version of every question in the dataset.
Verifiers are native speakers with substantial lived experience in the target culture; their responses test both whether the cultural facts are accurate and whether the local-language phrasing is clear and unambiguous.

We compute \textbf{inter-annotator agreement} (IAA) as the proportion of items on which all three verifiers select the same answer.
Table~\ref{tab:iaa} (Appendix~\ref{app:iaa}) reports IAA scores across all 20 cultural contexts; scores range from 87.0\% (South Africa) to 94.8\% (China), with a mean of 90.9\%, indicating high factual correctness and translation clarity across the dataset.
For items where verifiers disagree with the originally labeled correct answer, we review the item through discussions with the annotators and verifiers; if consensus confirms the verifiers' answer is correct, we update the gold label accordingly while keeping the question text unchanged.

\paragraph{LLM-assisted English proofreading.}
Because questions are produced by instantiating shared templates with culture-specific content, small surface-level inconsistencies can arise in the
English text across annotators and contexts (e.g., tense mismatches introduced by adapting a template from a generic present-tense form to a
past event).
To mitigate this template-instantiation noise without altering cultural content or reasoning difficulty, we run a deterministic LLM-assisted
proofreading pass on \emph{English} fields only (multiple-choice questions and options, and the English True/False statements).
For each English field, we query \texttt{openai/gpt-5.2-chat}\footnote{\url{https://cdn.openai.com/gpt-5-system-card.pdf}} to correct \emph{only} spelling and grammatical errors (including agreement, tense
consistency, and capitalization), while preserving the original writing style and word choices; rephrasing or stylistic improvement is explicitly disallowed. All local-language text is left exactly as written by annotators.

\subsection{Dataset Statistics}
\label{sec:stats}
After quality control and expansion, the benchmark contains {11{,}862} total evaluation instances.
Table~\ref{tab:data_stats} summarizes the distribution by cultural context (country), along with the associated local language and script.
Appendix~\ref{more-data} provides additional breakdowns of template coverage across cultural aspects (Figure~\ref{fig:culture})
and reasoning categories (Figure~\ref{fig:reasoning}).

\begin{table}[t]
\centering
\scriptsize
\setlength{\tabcolsep}{3pt}
\renewcommand{\arraystretch}{1.05}
\begin{tabular}{p{1.7cm}p{2.2cm}p{2.0cm}r}
\toprule
\textbf{Country} & \textbf{Language} & \textbf{Script} & \textbf{\#Q} \\
\midrule
Egypt        & Egyptian Arabic      & Arabic     & 594 \\
Philippines  & Tagalog              & Latin      & 594 \\
India        & Hindi                & Devanagari & 600 \\
Ethiopia     & Amharic              & Ge'ez      & 588 \\
Mexico       & Mexican Spanish      & Latin      & 594 \\
Tunisia      & Tunisian Arabic      & Arabic     & 600 \\
Greece       & Greek                & Greek      & 600 \\
Brazil       & Portuguese           & Latin      & 600 \\
Kyrgyzstan   & Kyrgyz               & Cyrillic   & 600 \\
South Africa & Zulu                 & Latin      & 600 \\
Italy        & Italian              & Latin      & 588 \\
Thailand     & Thai                 & Thai       & 594 \\
Turkey       & Turkish              & Latin      & 600 \\
Georgia      & Georgian             & Mkhedruli  & 594 \\
China        & Chinese              & Han (Hans) & 582 \\
Indonesia    & Indonesian           & Latin      & 570 \\
Yemen        & Yemeni Arabic        & Arabic     & 600 \\
Nigeria      & Yoruba               & Latin      & 570 \\
Morocco      & Moroccan Arabic      & Arabic     & 600 \\
Japan        & Japanese             & Japanese   & 594 \\
\midrule
\textbf{Total} &  &  & \textbf{11{,}862} \\
\bottomrule
\end{tabular}
\caption{Dataset statistics and coverage.
Small deviations across contexts reflect items removed in revisions.}
\label{tab:data_stats}
\end{table}

\begin{table*}[t]
\centering
\scriptsize
\setlength{\tabcolsep}{3pt}
\renewcommand{\arraystretch}{1.05}
\begin{tabularx}{\textwidth}{p{2.4cm}Xrrrrrrr}
\toprule
\textbf{Category} & \textbf{Model} & \textbf{MC-EN} & \textbf{MC-L} & \textbf{$\Delta$MC} & \textbf{TF-EN} & \textbf{TF-L} & \textbf{$\Delta$TF} & \textbf{Overall} \\
\midrule
\textbf{Closed (thinking)} & Gemini 3 Flash Preview \citep{gemini3flash2025} (thinking) & 89.5\% & 89.1\% & -0.4\% & 84.5\% & 82.7\% & -1.8\% & 86.5\% \\
 & Gemini 2.5 Pro \citep{comanici2025gemini25pushingfrontier} (thinking) & 87.5\% & 88.3\% & +0.8\% & 81.6\% & 81.2\% & -0.4\% & 84.7\% \\
 & Gemini 2.5 Flash \citep{comanici2025gemini25pushingfrontier} (thinking) & 70.1\% & 65.9\% & -4.2\% & 74.6\% & 74.1\% & -0.5\% & 71.2\% \\
\multicolumn{2}{l}{\textit{\textbf{Average (Closed (thinking))}}} & 82.4\% & 81.1\% & -1.3\% & 80.2\% & 79.3\% & -0.9\% & \textbf{80.8\%} \\
\midrule
\textbf{Closed (standard)} & Gemini 3 Flash Preview \citep{gemini3flash2025} (standard) & 86.8\% & 87.0\% & +0.2\% & 81.9\% & 79.7\% & -2.2\% & 83.9\% \\
 & Claude Opus 4.5 \citep{claude45opus2025} & 81.3\% & 80.2\% & -1.1\% & 76.9\% & 75.2\% & -1.7\% & 78.4\% \\
 & GPT-5 Chat \citep{gpt5systemcard2025} & 79.0\% & 78.2\% & -0.8\% & 77.1\% & 73.5\% & -3.6\% & 77.0\% \\
 & Gemini 2.5 Flash \citep{comanici2025gemini25pushingfrontier} & 80.2\% & 80.2\% & +0.0\% & 71.9\% & 70.5\% & -1.4\% & 75.7\% \\
 & Claude Haiku 4.5 \citep{claude45opus2025} & 70.5\% & 63.8\% & -6.7\% & 68.5\% & 64.8\% & -3.7\% & 66.9\% \\
 & GPT-4o-mini \citep{openai2024gpt4technicalreport} & 70.6\% & 65.3\% & -5.3\% & 67.0\% & 64.2\% & -2.8\% & 66.8\% \\
\multicolumn{2}{l}{\textit{\textbf{Average (Closed (standard))}}} & 78.1\% & 75.8\% & -2.3\% & 73.9\% & 71.3\% & -2.6\% & \textbf{74.8\%} \\
\midrule
\textbf{Open-weight} & DeepSeek-Chat v3.1 \citep{deepseekai2025deepseekv3technicalreport} (thinking) & 76.2\% & 75.6\% & -0.6\% & 76.7\% & 71.7\% & -5.0\% & 75.1\% \\
 & DeepSeek-Chat v3.1 \citep{deepseekai2025deepseekv3technicalreport} & 74.0\% & 68.4\% & -5.6\% & 67.9\% & 64.1\% & -3.8\% & 68.6\% \\
 & Qwen3-235B-A22B \citep{yang2025qwen3technicalreport} & 73.3\% & 68.3\% & -5.0\% & 67.3\% & 65.9\% & -1.4\% & 68.7\% \\
 & Llama 3.3-70B \citep{grattafiori2024llama3herdmodels} & 70.2\% & 62.6\% & -7.6\% & 67.5\% & 62.0\% & -5.5\% & 65.6\% \\
 & Llama 4 Maverick \citep{llama4maverick2025} & 68.7\% & 67.5\% & -1.2\% & 64.1\% & 62.1\% & -2.0\% & 65.6\% \\
 & Llama 3.1-8B \citep{grattafiori2024llama3herdmodels} & 54.2\% & 43.4\% & -10.8\% & 56.7\% & 53.4\% & -3.3\% & 51.9\% \\
 & Qwen3-4B \citep{yang2025qwen3technicalreport} & 52.6\% & 45.6\% & -7.0\% & 55.2\% & 53.7\% & -1.5\% & 51.8\% \\
 & InternLM3-8B \citep{internlm3_8b} & 54.8\% & 40.9\% & -13.9\% & 55.8\% & 52.2\% & -3.6\% & 50.9\% \\
 & Qwen2.5-7B \citep{qwen2025qwen25technicalreport} & 57.0\% & 46.9\% & -10.1\% & 52.6\% & 52.6\% & +0.0\% & 52.3\% \\
 & Aya Expanse-8B \citep{dang2024ayaexpansecombiningresearch} & 52.7\% & 48.7\% & -4.0\% & 51.7\% & 52.5\% & +0.8\% & 51.4\% \\
 & Llama 3.2-3B \citep{grattafiori2024llama3herdmodels} & 47.4\% & 36.3\% & -11.1\% & 54.9\% & 51.6\% & -3.3\% & 47.6\% \\
 & Aya-23-8B \citep{aryabumi2024aya23openweight} & 43.8\% & 39.2\% & -4.6\% & 50.4\% & 50.5\% & +0.1\% & 46.0\% \\
\multicolumn{2}{l}{\textit{\textbf{Average (Open-weight)}}} & 60.4\% & 53.6\% & -6.8\% & 60.1\% & 57.7\% & -2.4\% & \textbf{58.0\%} \\
\midrule
\multicolumn{2}{l}{\textbf{Average (All)}} & 68.6\% & 63.9\% & -4.7\% & 66.9\% & 64.7\% & -2.2\% & \textbf{66.0\%} \\
\bottomrule
\end{tabularx}
\caption{Zero-shot accuracy (\%) of 21 LLMs on \dataname, grouped by model category. \textbf{MC-EN}/\textbf{MC-L}: multiple-choice accuracy in English and local language; \textbf{TF-EN}/\textbf{TF-L}: True/False accuracy; $\boldsymbol{\Delta}$\textbf{MC}/$\boldsymbol{\Delta}$\textbf{TF}: cross-lingual gap (Local$-$English), where negative values indicate degraded local-language performance. \textbf{Overall} averages all four scores.}
\label{tab:main_nohard}
\end{table*}

\section{Experimental Setup}
\label{sec:exp}
We evaluate 21 multilingual LLMs in zero-shot on both multiple-choice (MC) and True/False (T/F) formats, using paired English and local-language versions to isolate the effects of \emph{language}, \emph{cultural grounding}, and \emph{reasoning type}. 

Open-weight models use log-probability scoring; API-only models, including thinking models, output a structured JSON answer on the final line (Appendix~\ref{app:eval-prompts}). Full scoring details and validation against generation-based scoring are in Appendix~\ref{app:logprob-vs-gen}. Answer extraction error rates for generation-based models are near zero across all models and do not meaningfully affect conclusions (Appendix~\ref{app:extraction-errors}).



\begin{table*}[t]
\centering
\scriptsize
\setlength{\tabcolsep}{3pt}
\renewcommand{\arraystretch}{1.12}

\begin{tabular}{r p{6.5cm} || r p{6.5cm}}
\toprule
\multicolumn{2}{c||}{\textbf{Top-5 Easiest Templates (avg.\ over models)}} &
\multicolumn{2}{c}{\textbf{Top-5 Hardest Templates (avg.\ over models)}} \\
\midrule
\textbf{Acc} & \textbf{Template prompt} & \textbf{Acc} & \textbf{Template prompt} \\
\midrule
92.51\% &
\texttt{If my friend has the last name [LAST NAME], which country is most likely their birthplace?}
&
35.76\% &
\texttt{Among all the provinces in [SET OF PROVINCES/LOCATION], how many provinces have an area smaller than [PROVINCE]?}
\\[2pt]

85.66\% &
\texttt{Which of the following authors is traditionally associated with [CULTURE/REGION]'s literature?}
&
37.52\% &
\texttt{[GIFT 1], [GIFT 2], [GIFT 3], and [GIFT 4]. How many of these are suitable gifts to give when attending a [NATIONALITY] friend's wedding?}
\\[2pt]

85.44\% &
\texttt{If I am visiting [PLACE] in [MONTH], what kind of clothes should I pack?}
&
38.31\% &
\texttt{[MATERIAL 1], [MATERIAL 2], [MATERIAL 3], [MATERIAL 4]. How many of these are not typically used when making [TRADITIONAL ITEM]?}
\\[2pt]

84.98\% &
\texttt{In [NATIONALITY OR CULTURE] attire, what determines the [GARMENT FEATURE]?}
&
39.61\% &
\texttt{Among all [POLITICAL POSITION] in [COUNTRY] up until [YEAR], how many of them are [CONDITION]?}
\\[2pt]

84.36\% &
\texttt{Which of these political systems was/is traditionally practiced in [REGION] during the [TIME PERIOD]?}
&
42.30\% &
\texttt{[FOOD 1], [FOOD 2], [FOOD 3], [FOOD 4], how many of these are [SOME CLASSIFICATION]?}
\\
\bottomrule
\end{tabular}
\caption{Template difficulty extremes: easiest prompts are simple cultural associations, while hardest prompts are constraint-heavy exact-count questions.}
\label{tab:template_extremes_models_prompts}
\end{table*}

\section{Results and Discussion}
\label{sec:results}
Table~\ref{tab:main_nohard} reports overall performance on scenario-aligned English and local-language
instances in both multiple-choice (MC) and True/False formats, grouped by model category.
We additionally report cross-lingual gaps $\Delta$MC and $\Delta$TF, computed as (Local $-$ English),
where negative values indicate degraded performance in the local language.
Per-language and per-script breakdowns are provided in Appendix~\ref{app:detailed-benchmark}.

\begin{figure}[t]
  \centering
  \includegraphics[width=\columnwidth]{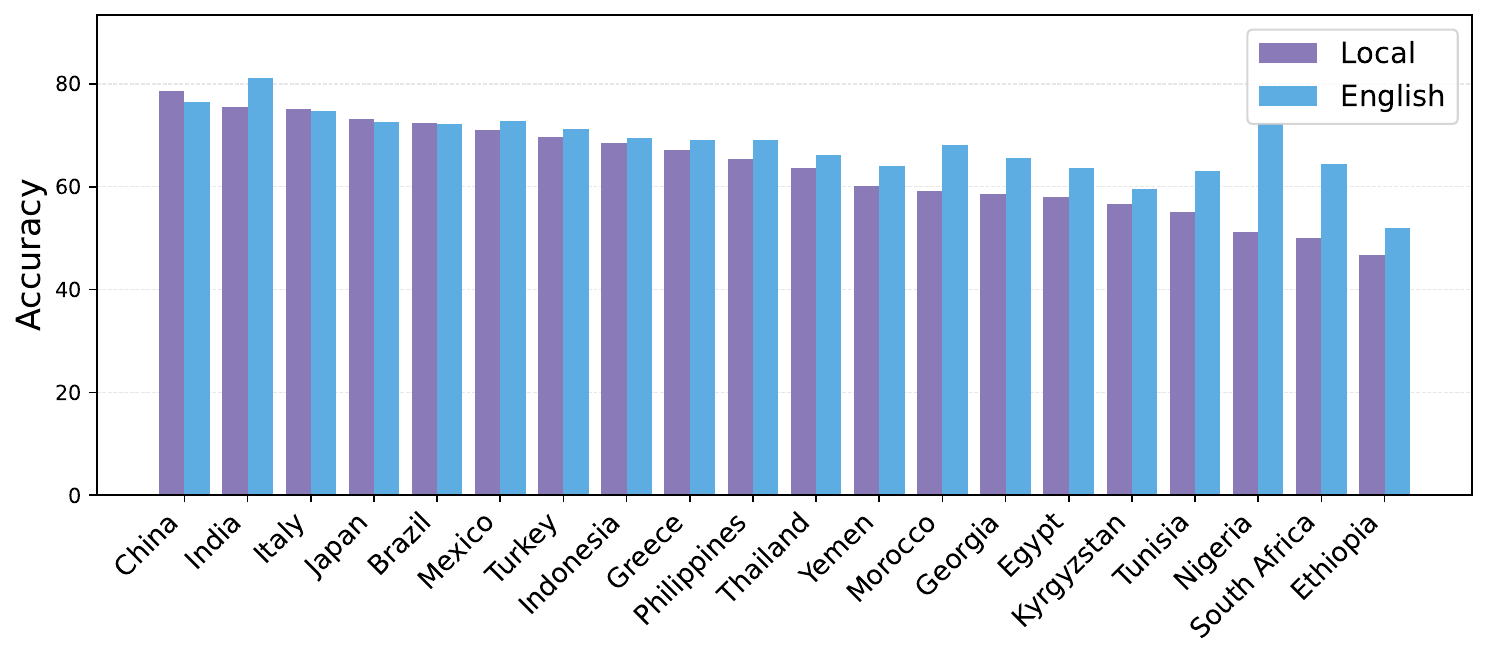}
  \caption{Average performance in English vs.\ local language across cultural contexts.}
  \label{fig:eng_local}
\end{figure}

\begin{figure}[ht!]
  \centering
  \includegraphics[width=\columnwidth]{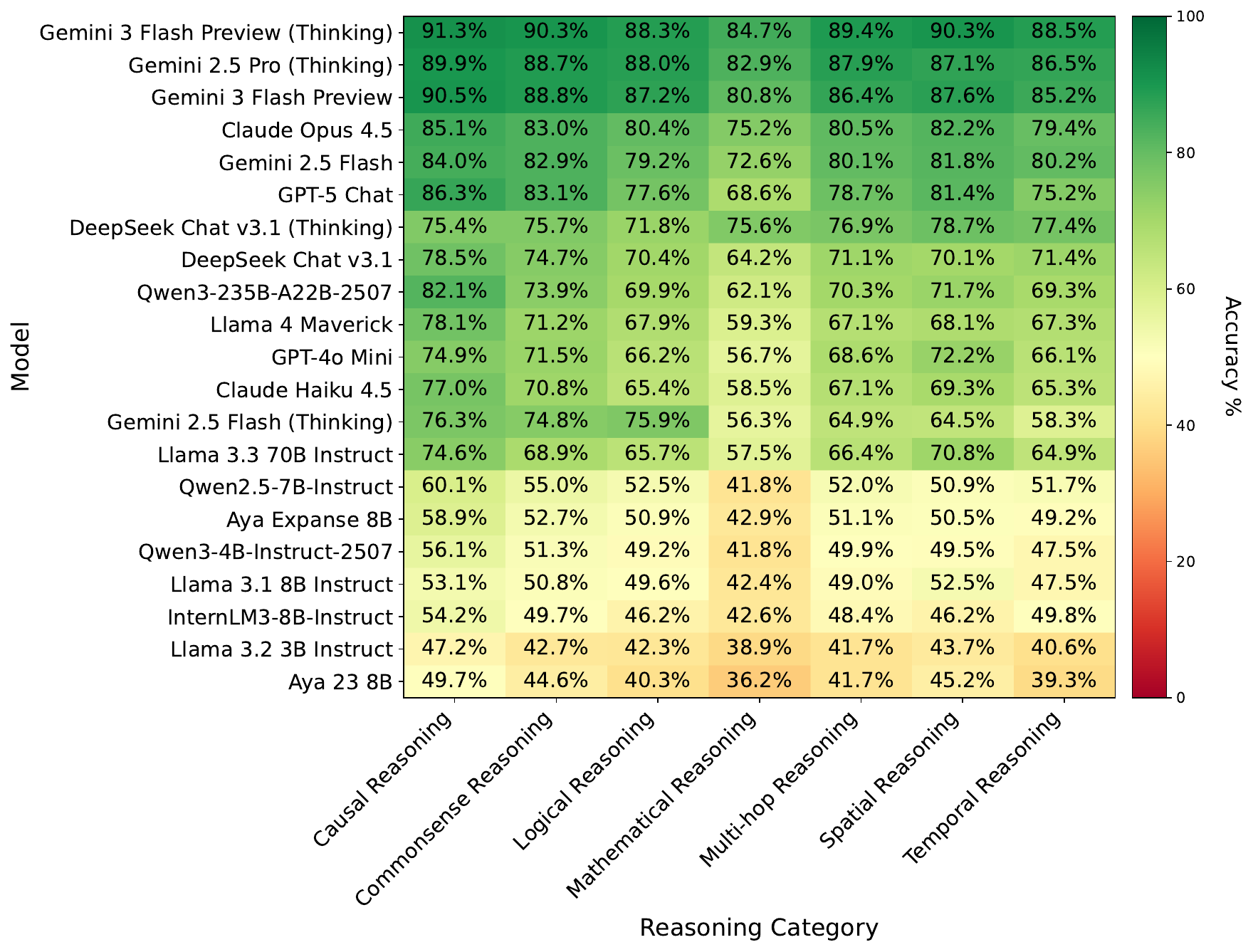}
  \caption{Accuracy by reasoning type across models.}
  \label{fig:reasoning_heatmap}
\end{figure}

\begin{table}[t]
\centering
\scriptsize
\setlength{\tabcolsep}{3pt}
\renewcommand{\arraystretch}{1.05}
\begin{tabular}{p{2.6cm}rrr}
\toprule
\textbf{Category} & \textbf{T/F (avg.\ EN/L)} & \textbf{Paired T/F} & \textbf{Drop (pp)} \\
\midrule
Closed (thinking) & 78.4\% & 62.1\% & 16.3 \\
Closed (standard) & 71.7\% & 48.4\% & 23.3 \\
Open-weight       & 56.6\% & 18.2\% & 38.4 \\
\midrule
\textbf{Average (All)} & \textbf{65.8\%} & \textbf{36.6\%} & \textbf{29.2} \\
\bottomrule
\end{tabular}
\caption{
True/False accuracy (averaged over English and local vs.\ paired True/False accuracy by model category. \textbf{Drop} is the difference in percentage points.
}
\label{tab:tf_vs_paired}
\end{table}

\begin{figure*}[ht!]
  \centering
  \includegraphics[width=2\columnwidth]{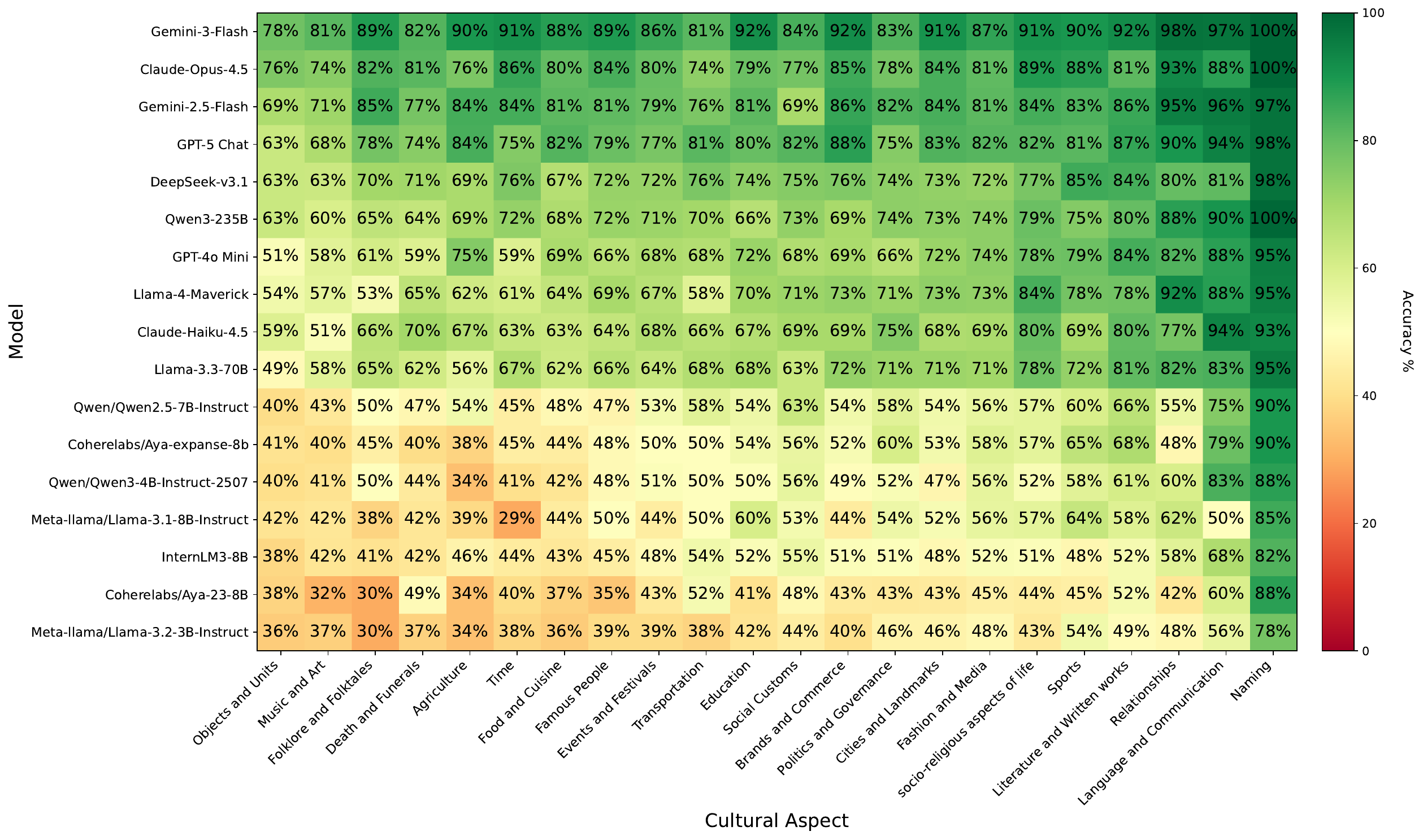}
  \caption{Accuracy (\%) by cultural aspect across models. Each cell aggregates over evaluation instances whose templates are tagged with the corresponding aspect (multi-label; a single instance may contribute to multiple aspects). Aspects are ordered by mean accuracy (hard $\rightarrow$ easy).}
  \label{fig:bycultural_aspects2}
\end{figure*}

\noindent\textbf{Open-weight models show larger English--local gaps and reduced reliability on T/F.}
Closed-source thinking models achieve the highest overall accuracy (\textbf{80.8\%} on average),
outperforming closed (standard) models (\textbf{74.8\%}) and open-weight models (\textbf{58.0\%}).
The English--local gap widens as model capacity decreases: thinking models are near-parity (Avg.\ $\Delta$MC $=-1.3$), whereas open-weight models exhibit much larger drops
(Avg.\ $\Delta$MC $=-6.8$), particularly at 3B--8B scale.
While most models are above random baseline in MC, many open-weight models cluster around
$\sim$50--55\% on T/F, indicating limited reliability
in binary verification of culturally grounded statements.

\noindent\textbf{Paired True/False accuracy exposes scenario-level verification.}
Each MC scenario yields a True/False pair (Section~\ref{sec:truefalse}); we count a scenario correct only if the model answers both correctly. As shown in Table~\ref{tab:tf_vs_paired}, paired accuracy is substantially lower than per-question T/F accuracy across all categories, suggesting single T/F performance is inflated by shallow response tendencies rather than genuine cultural knowledge.

\noindent\textbf{English outperforms local except for China, while the largest gaps concentrate in lower-resource languages.}
Figure~\ref{fig:eng_local} shows China as nearly the only case where local slightly exceeds English, plausibly influenced
by the strong presence of Chinese-focused models (e.g., multiple Qwen variants) in our evaluation.
In contrast, the English--local gap is substantially larger for Amharic, Yoruba, Zulu,
and Arabic dialects, highlighting that cross-lingual brittleness concentrates
in lower-coverage languages and dialects.

\noindent\textbf{Mathematical reasoning is hardest in cultural contexts; causal and temporal reasoning are easier.}
In Figure~\ref{fig:reasoning_heatmap}, Mathematical Reasoning is the lowest-accuracy category for 20/21 models, while Causal (often alongside Commonsense) is typically highest.
We attribute this to a double burden: culture-grounded math questions require both retrieving long-tail, locale-specific numeric facts and performing exact composition (e.g., counting or aggregation), so either retrieval or calculation errors can flip the answer. Moreover, such numeric facts are often sparse in training data and region-specific, raising the risk of confident but incorrect answers. In contrast, causal and commonsense questions are
often supported by broadly shared everyday knowledge that is better covered in pretraining corpora.

\noindent\textbf{Cultural aspect difficulty is consistent across models.}
Figure~\ref{fig:bycultural_aspects2} reports a model$\times$aspect heatmap and reveals a stable hard $\rightarrow$ easy ordering
across model families. Averaged over models, \emph{Naming} is the easiest aspect (92.5\%), followed by
\emph{Language and Communication} (80.6\%), whereas \emph{Objects and Units} (52.9\%) and \emph{Music and Art} (54.0\%) are the
hardest, yielding a $\sim$40-point spread. Most remaining aspects form a broad middle band (roughly 64--70\% mean accuracy).
Beyond mean difficulty, the heatmap highlights where robustness gaps are largest: \emph{Time} varies from 91\% (strongest model)
down to 29\% (smaller open-weight models), with similarly large spreads across other aspects.

\noindent\textbf{``How many'' templates are the main failure mode.}
Table~\ref{tab:template_extremes_models_prompts} shows that the easiest templates are mostly single-step cultural
associations (e.g., last name $\rightarrow$ likely birthplace at 92.51\%), while the hardest are uniformly ``How many\ldots'' prompts
that require enumerating a culturally grounded set, applying a constraint (often with negation/comparison), and producing an exact
count (down to 35.76\%). This pattern is consistent with our earlier finding that mathematical reasoning remains the weakest capability in culturally situated scenarios.
\section{Conclusion}
We introduce a template-first benchmark for multilingual, multicultural reasoning across 20 cultural contexts, languages and dialects, and 10 scripts ({11{,}862} total instances). Our dataset separates language, cultural grounding, and reasoning type using scenario-aligned multiple-choice and True/False items. Zero-shot evaluations show that closed reasoning models achieve near-parity between English and local inputs, while open-weight models lag with significant performance drops. This benchmark supports diagnostic evaluation to motivate more culturally robust model development.

\section*{Limitations}
Despite its breadth, the benchmark has several limitations. First, \textbf{coverage} is necessarily coarse: we include 20 cultural contexts with one primary local language each, which cannot capture within-country cultural diversity, minority languages, or finer-grained dialect continua. As a result, performance within a single country or language should not be interpreted as representative of all local varieties or communities. Second, the \textbf{task format} is intentionally controlled: while multiple-choice and binary verification enable precise alignment and diagnostic evaluation, they do not reflect open-ended dialogue, interactive reasoning, tool use, or real-world information access. Consequently, the benchmark measures culturally grounded reasoning under constrained conditions rather than end-to-end performance in realistic deployment settings.

\section*{Ethical Considerations}

{\dataname} is a human-written benchmark designed to evaluate multilingual reasoning over culturally grounded premises in a controlled, template-first setting.

\paragraph{Annotator compensation.}
We recruited two annotators per cultural context via Upwork and compensated them at a fixed rate of US\$9 per 10 completed template instantiations, where a completion consists of writing the English multiple-choice item and translating the question, options, and T/F variants into the local language. We additionally recruited three independent native verifiers per cultural context to assess factual correctness and translation quality, compensated at a rate of US\$8 per hour. All annotators and verifiers were anonymized, and no personally identifiable information about them is included in the dataset or released publicly.

\paragraph{Cultural sensitivity and representational harms.}
Because items are cultural-based, there is a risk of stereotyping, oversimplifying a country/region into a single culture, or encoding contested practices as universal. We mitigate this through (i) iterative template refinement to remove culturally insensitive/non-portable designs and reduce ambiguity, (ii) annotator guidelines that emphasize culturally representative everyday knowledge while avoiding stereotypes and obscure trivia, and (iii) plausible within-context distractors to reduce “foreign-option elimination.” Annotators may also flag templates as non-applicable when they do not meaningfully transfer.

Coverage is coarse (one primary local language per cultural context), so results should not be interpreted as measuring within-country diversity or dialect variation. As with any benchmark, {\dataname} can be misused for overfitting or for simplistic “ranking” of languages/cultures; we recommend using it as a diagnostic tool and reporting results with the above coverage and format constraints in mind.

\paragraph{LLM assistance in writing.}
We used an LLM as a writing aid (e.g., to improve clarity and correct minor grammar issues) while drafting this manuscript. All technical content, experimental design, analyses, and conclusions were produced and verified by the authors, who take full responsibility for the final paper. We did not provide any sensitive or personally identifying information to the model.




\bibliography{custom}

@misc{zellers2019hellaswagmachinereallyfinish,
      title={HellaSwag: Can a Machine Really Finish Your Sentence?}, 
      author={Rowan Zellers and Ari Holtzman and Yonatan Bisk and Ali Farhadi and Yejin Choi},
      year={2019},
      eprint={1905.07830},
      archivePrefix={arXiv},
      primaryClass={cs.CL},
      url={https://arxiv.org/abs/1905.07830}, 
}

@misc{deepseekai2025deepseekv3technicalreport,
      title={DeepSeek-V3 Technical Report}, 
      author={DeepSeek-AI and Aixin Liu and Bei Feng and Bing Xue and Bingxuan Wang and Bochao Wu and Chengda Lu and Chenggang Zhao and Chengqi Deng and Chenyu Zhang and Chong Ruan and Damai Dai and Daya Guo and Dejian Yang and Deli Chen and Dongjie Ji and Erhang Li and Fangyun Lin and Fucong Dai and Fuli Luo and Guangbo Hao and Guanting Chen and Guowei Li and H. Zhang and Han Bao and Hanwei Xu and Haocheng Wang and Haowei Zhang and Honghui Ding and Huajian Xin and Huazuo Gao and Hui Li and Hui Qu and J. L. Cai and Jian Liang and Jianzhong Guo and Jiaqi Ni and Jiashi Li and Jiawei Wang and Jin Chen and Jingchang Chen and Jingyang Yuan and Junjie Qiu and Junlong Li and Junxiao Song and Kai Dong and Kai Hu and Kaige Gao and Kang Guan and Kexin Huang and Kuai Yu and Lean Wang and Lecong Zhang and Lei Xu and Leyi Xia and Liang Zhao and Litong Wang and Liyue Zhang and Meng Li and Miaojun Wang and Mingchuan Zhang and Minghua Zhang and Minghui Tang and Mingming Li and Ning Tian and Panpan Huang and Peiyi Wang and Peng Zhang and Qiancheng Wang and Qihao Zhu and Qinyu Chen and Qiushi Du and R. J. Chen and R. L. Jin and Ruiqi Ge and Ruisong Zhang and Ruizhe Pan and Runji Wang and Runxin Xu and Ruoyu Zhang and Ruyi Chen and S. S. Li and Shanghao Lu and Shangyan Zhou and Shanhuang Chen and Shaoqing Wu and Shengfeng Ye and Shengfeng Ye and Shirong Ma and Shiyu Wang and Shuang Zhou and Shuiping Yu and Shunfeng Zhou and Shuting Pan and T. Wang and Tao Yun and Tian Pei and Tianyu Sun and W. L. Xiao and Wangding Zeng and Wanjia Zhao and Wei An and Wen Liu and Wenfeng Liang and Wenjun Gao and Wenqin Yu and Wentao Zhang and X. Q. Li and Xiangyue Jin and Xianzu Wang and Xiao Bi and Xiaodong Liu and Xiaohan Wang and Xiaojin Shen and Xiaokang Chen and Xiaokang Zhang and Xiaosha Chen and Xiaotao Nie and Xiaowen Sun and Xiaoxiang Wang and Xin Cheng and Xin Liu and Xin Xie and Xingchao Liu and Xingkai Yu and Xinnan Song and Xinxia Shan and Xinyi Zhou and Xinyu Yang and Xinyuan Li and Xuecheng Su and Xuheng Lin and Y. K. Li and Y. Q. Wang and Y. X. Wei and Y. X. Zhu and Yang Zhang and Yanhong Xu and Yanhong Xu and Yanping Huang and Yao Li and Yao Zhao and Yaofeng Sun and Yaohui Li and Yaohui Wang and Yi Yu and Yi Zheng and Yichao Zhang and Yifan Shi and Yiliang Xiong and Ying He and Ying Tang and Yishi Piao and Yisong Wang and Yixuan Tan and Yiyang Ma and Yiyuan Liu and Yongqiang Guo and Yu Wu and Yuan Ou and Yuchen Zhu and Yuduan Wang and Yue Gong and Yuheng Zou and Yujia He and Yukun Zha and Yunfan Xiong and Yunxian Ma and Yuting Yan and Yuxiang Luo and Yuxiang You and Yuxuan Liu and Yuyang Zhou and Z. F. Wu and Z. Z. Ren and Zehui Ren and Zhangli Sha and Zhe Fu and Zhean Xu and Zhen Huang and Zhen Zhang and Zhenda Xie and Zhengyan Zhang and Zhewen Hao and Zhibin Gou and Zhicheng Ma and Zhigang Yan and Zhihong Shao and Zhipeng Xu and Zhiyu Wu and Zhongyu Zhang and Zhuoshu Li and Zihui Gu and Zijia Zhu and Zijun Liu and Zilin Li and Ziwei Xie and Ziyang Song and Ziyi Gao and Zizheng Pan},
      year={2025},
      eprint={2412.19437},
      archivePrefix={arXiv},
      primaryClass={cs.CL},
      url={https://arxiv.org/abs/2412.19437}, 
}

@misc{comanici2025gemini25pushingfrontier,
      title={Gemini 2.5: Pushing the Frontier with Advanced Reasoning, Multimodality, Long Context, and Next Generation Agentic Capabilities}, 
      author={Gheorghe Comanici and Eric Bieber and Mike Schaekermann and Ice Pasupat and Noveen Sachdeva and Inderjit Dhillon and Marcel Blistein and Ori Ram and Dan Zhang and Evan Rosen and Luke Marris and Sam Petulla and Colin Gaffney and Asaf Aharoni and Nathan Lintz and Tiago Cardal Pais and Henrik Jacobsson and Idan Szpektor and Nan-Jiang Jiang and Krishna Haridasan and Ahmed Omran and Nikunj Saunshi and Dara Bahri and Gaurav Mishra and Eric Chu and Toby Boyd and Brad Hekman and Aaron Parisi and Chaoyi Zhang and Kornraphop Kawintiranon and Tania Bedrax-Weiss and Oliver Wang and Ya Xu and Ollie Purkiss and Uri Mendlovic and Ilaï Deutel and Nam Nguyen and Adam Langley and Flip Korn and Lucia Rossazza and Alexandre Ramé and Sagar Waghmare and Helen Miller and Nathan Byrd and Ashrith Sheshan and Raia Hadsell and Sangnie Bhardwaj and Pawel Janus and Tero Rissa and Dan Horgan and Alvin Abdagic and Lior Belenki and James Allingham and Anima Singh and Theo Guidroz and Srivatsan Srinivasan and Herman Schmit and Kristen Chiafullo and Andre Elisseeff and Nilpa Jha and Prateek Kolhar and Leonard Berrada and Frank Ding and Xiance Si and Shrestha Basu Mallick and Franz Och and Sofia Erell and Eric Ni and Tejasi Latkar and Sherry Yang and Petar Sirkovic and Ziqiang Feng and Robert Leland and Rachel Hornung and Gang Wu and Charles Blundell and Hamidreza Alvari and Po-Sen Huang and Cathy Yip and Sanja Deur and Li Liu and Gabriela Surita and Pablo Duque and Dima Damen and Johnson Jia and Arthur Guez and Markus Mircea and Animesh Sinha and Alberto Magni and Paweł Stradomski and Tal Marian and Vlado Galić and Wenhu Chen and Hisham Husain and Achintya Singhal and Dominik Grewe and François-Xavier Aubet and Shuang Song and Lorenzo Blanco and Leland Rechis and Lewis Ho and Rich Munoz and Kelvin Zheng and Jessica Hamrick and Kevin Mather and Hagai Taitelbaum and Eliza Rutherford and Yun Lei and Kuangyuan Chen and Anand Shukla and Erica Moreira and Eric Doi and Berivan Isik and Nir Shabat and Dominika Rogozińska and Kashyap Kolipaka and Jason Chang and Eugen Vušak and Srinivasan Venkatachary and Shadi Noghabi and Tarun Bharti and Younghoon Jun and Aleksandr Zaks and Simon Green and Jeshwanth Challagundla and William Wong and Muqthar Mohammad and Dean Hirsch and Yong Cheng and Iftekhar Naim and Lev Proleev and Damien Vincent and Aayush Singh and Maxim Krikun and Dilip Krishnan and Zoubin Ghahramani and Aviel Atias and Rajeev Aggarwal and Christo Kirov and Dimitrios Vytiniotis and Christy Koh and Alexandra Chronopoulou and Pawan Dogra and Vlad-Doru Ion and Gladys Tyen and Jason Lee and Felix Weissenberger and Trevor Strohman and Ashwin Balakrishna and Jack Rae and Marko Velic and Raoul de Liedekerke and Oded Elyada and Wentao Yuan and Canoee Liu and Lior Shani and Sergey Kishchenko and Bea Alessio and Yandong Li and Richard Song and Sam Kwei and Orion Jankowski and Aneesh Pappu and Youhei Namiki and Yenai Ma and Nilesh Tripuraneni and Colin Cherry and Marissa Ikonomidis and Yu-Cheng Ling and Colin Ji and Beka Westberg and Auriel Wright and Da Yu and David Parkinson and Swaroop Ramaswamy and Jerome Connor and Soheil Hassas Yeganeh and Snchit Grover and George Kenwright and Lubo Litchev and Chris Apps and Alex Tomala and Felix Halim and Alex Castro-Ros and Zefei Li and Anudhyan Boral and Pauline Sho and Michal Yarom and Eric Malmi and David Klinghoffer and Rebecca Lin and Alan Ansell and Pradeep Kumar S and Shubin Zhao and Siqi Zuo and Adam Santoro and Heng-Tze Cheng and Solomon Demmessie and Yuchi Liu and Nicole Brichtova and Allie Culp and Nathaniel Braun and Dan Graur and Will Ng and Nikhil Mehta and Aaron Phillips and Patrik Sundberg and Varun Godbole and Fangyu Liu and Yash Katariya and David Rim and Mojtaba Seyedhosseini and Sean Ammirati and Jonas Valfridsson and Mahan Malihi and Timothy Knight and Andeep Toor and Thomas Lampe and Abe Ittycheriah and Lewis Chiang and Chak Yeung and Alexandre Fréchette and Jinmeng Rao and Huisheng Wang and Himanshu Srivastava and Richard Zhang and Rocky Rhodes and Ariel Brand and Dean Weesner and Ilya Figotin and Felix Gimeno and Rachana Fellinger and Pierre Marcenac and José Leal and Eyal Marcus and Victor Cotruta and Rodrigo Cabrera and Sheryl Luo and Dan Garrette and Vera Axelrod and Sorin Baltateanu and David Barker and Dongkai Chen and Horia Toma and Ben Ingram and Jason Riesa and Chinmay Kulkarni and Yujing Zhang and Hongbin Liu and Chao Wang and Martin Polacek and Will Wu and Kai Hui and Adrian N Reyes and Yi Su and Megan Barnes and Ishaan Malhi and Anfal Siddiqui and Qixuan Feng and Mihai Damaschin and Daniele Pighin and Andreas Steiner and Samuel Yang and Ramya Sree Boppana and Simeon Ivanov and Arun Kandoor and Aditya Shah and Asier Mujika and Da Huang and Christopher A. Choquette-Choo and Mohak Patel and Tianhe Yu and Toni Creswell and Jerry and Liu and Catarina Barros and Yasaman Razeghi and Aurko Roy and Phil Culliton and Binbin Xiong and Jiaqi Pan and Thomas Strohmann and Tolly Powell and Babi Seal and Doug DeCarlo and Pranav Shyam and Kaan Katircioglu and Xuezhi Wang and Cassidy Hardin and Immanuel Odisho and Josef Broder and Oscar Chang and Arun Nair and Artem Shtefan and Maura O'Brien and Manu Agarwal and Sahitya Potluri and Siddharth Goyal and Amit Jhindal and Saksham Thakur and Yury Stuken and James Lyon and Kristina Toutanova and Fangxiaoyu Feng and Austin Wu and Ben Horn and Alek Wang and Alex Cullum and Gabe Taubman and Disha Shrivastava and Chongyang Shi and Hamish Tomlinson and Roma Patel and Tao Tu and Ada Maksutaj Oflazer and Francesco Pongetti and Mingyao Yang and Adrien Ali Taïga and Vincent Perot and Nuo Wang Pierse and Feng Han and Yoel Drori and Iñaki Iturrate and Ayan Chakrabarti and Legg Yeung and Dave Dopson and Yi-ting Chen and Apoorv Kulshreshtha and Tongfei Guo and Philip Pham and Tal Schuster and Junquan Chen and Alex Polozov and Jinwei Xing and Huanjie Zhou and Praneeth Kacham and Doron Kukliansky and Antoine Miech and Sergey Yaroshenko and Ed Chi and Sholto Douglas and Hongliang Fei and Mathieu Blondel and Preethi Myla and Lior Madmoni and Xing Wu and Daniel Keysers and Kristian Kjems and Isabela Albuquerque and Lijun Yu and Joel D'sa and Michelle Plantan and Vlad Ionescu and Jaume Sanchez Elias and Abhirut Gupta and Manish Reddy Vuyyuru and Fred Alcober and Tong Zhou and Kaiyang Ji and Florian Hartmann and Subha Puttagunta and Hugo Song and Ehsan Amid and Anca Stefanoiu and Andrew Lee and Paul Pucciarelli and Emma Wang and Amit Raul and Slav Petrov and Isaac Tian and Valentin Anklin and Nana Nti and Victor Gomes and Max Schumacher and Grace Vesom and Alex Panagopoulos and Konstantinos Bousmalis and Daniel Andor and Josh Jacob and Yuan Zhang and Bill Rosgen and Matija Kecman and Matthew Tung and Alexandra Belias and Noah Goodman and Paul Covington and Brian Wieder and Nikita Saxena and Elnaz Davoodi and Muhuan Huang and Sharath Maddineni and Vincent Roulet and Folawiyo Campbell-Ajala and Pier Giuseppe Sessa and Xintian and Wu and Guangda Lai and Paul Collins and Alex Haig and Vytenis Sakenas and Xiaowei Xu and Marissa Giustina and Laurent El Shafey and Pichi Charoenpanit and Shefali Garg and Joshua Ainslie and Boone Severson and Montse Gonzalez Arenas and Shreya Pathak and Sujee Rajayogam and Jie Feng and Michiel Bakker and Sheng Li and Nevan Wichers and Jamie Rogers and Xinyang Geng and Yeqing Li and Rolf Jagerman and Chao Jia and Nadav Olmert and David Sharon and Matthew Mauger and Sandeep Mariserla and Hongxu Ma and Megha Mohabey and Kyuyeun Kim and Alek Andreev and Scott Pollom and Juliette Love and Vihan Jain and Priyanka Agrawal and Yannick Schroecker and Alisa Fortin and Manfred Warmuth and Ji Liu and Andrew Leach and Irina Blok and Ganesh Poomal Girirajan and Roee Aharoni and Benigno Uria and Andrei Sozanschi and Dan Goldberg and Lucian Ionita and Marco Tulio Ribeiro and Martin Zlocha and Vighnesh Birodkar and Sami Lachgar and Liangzhe Yuan and Himadri Choudhury and Matt Ginsberg and Fei Zheng and Gregory Dibb and Emily Graves and Swachhand Lokhande and Gabriel Rasskin and George-Cristian Muraru and Corbin Quick and Sandeep Tata and Pierre Sermanet and Aditya Chawla and Itay Karo and Yan Wang and Susan Zhang and Orgad Keller and Anca Dragan and Guolong Su and Ian Chou and Xi Liu and Yiqing Tao and Shruthi Prabhakara and Marc Wilson and Ruibo Liu and Shibo Wang and Georgie Evans and David Du and Alfonso Castaño and Gautam Prasad and Mona El Mahdy and Sebastian Gerlach and Machel Reid and Jarrod Kahn and Amir Zait and Thanumalayan Sankaranarayana Pillai and Thatcher Ulrich and Guanyu Wang and Jan Wassenberg and Efrat Farkash and Kiran Yalasangi and Congchao Wang and Maria Bauza and Simon Bucher and Ting Liu and Jun Yan and Gary Leung and Vikas Sindhwani and Parker Barnes and Avi Singh and Ivan Jurin and Jichuan Chang and Niket Kumar Bhumihar and Sivan Eiger and Gui Citovsky and Ben Withbroe and Zhang Li and Siyang Xue and Niccolò Dal Santo and Georgi Stoyanov and Yves Raimond and Steven Zheng and Yilin Gao and Vít Listík and Sławek Kwasiborski and Rachel Saputro and Adnan Ozturel and Ganesh Mallya and Kushal Majmundar and Ross West and Paul Caron and Jinliang Wei and Lluis Castrejon and Sharad Vikram and Deepak Ramachandran and Nikhil Dhawan and Jiho Park and Sara Smoot and George van den Driessche and Yochai Blau and Chase Malik and Wei Liang and Roy Hirsch and Cicero Nogueira dos Santos and Eugene Weinstein and Aäron van den Oord and Sid Lall and Nicholas FitzGerald and Zixuan Jiang and Xuan Yang and Dale Webster and Ali Elqursh and Aedan Pope and Georges Rotival and David Raposo and Wanzheng Zhu and Jeff Dean and Sami Alabed and Dustin Tran and Arushi Gupta and Zach Gleicher and Jessica Austin and Edouard Rosseel and Megh Umekar and Dipanjan Das and Yinghao Sun and Kai Chen and Karolis Misiunas and Xiang Zhou and Yixian Di and Alyssa Loo and Josh Newlan and Bo Li and Vinay Ramasesh and Ying Xu and Alex Chen and Sudeep Gandhe and Radu Soricut and Nikita Gupta and Shuguang Hu and Seliem El-Sayed and Xavier Garcia and Idan Brusilovsky and Pu-Chin Chen and Andrew Bolt and Lu Huang and Alex Gurney and Zhiying Zhang and Alexander Pritzel and Jarek Wilkiewicz and Bryan Seybold and Bhargav Kanagal Shamanna and Felix Fischer and Josef Dean and Karan Gill and Ross Mcilroy and Abhishek Bhowmick and Jeremy Selier and Antoine Yang and Derek Cheng and Vladimir Magay and Jie Tan and Dhriti Varma and Christian Walder and Tomas Kocisky and Ryo Nakashima and Paul Natsev and Mike Kwong and Ionel Gog and Chiyuan Zhang and Sander Dieleman and Thomas Jimma and Andrey Ryabtsev and Siddhartha Brahma and David Steiner and Dayou Du and Ante Žužul and Mislav Žanić and Mukund Raghavachari and Willi Gierke and Zeyu Zheng and Dessie Petrova and Yann Dauphin and Yuchuan Liu and Ido Kessler and Steven Hand and Chris Duvarney and Seokhwan Kim and Hyo Lee and Léonard Hussenot and Jeffrey Hui and Josh Smith and Deepali Jain and Jiawei Xia and Gaurav Singh Tomar and Keyvan Amiri and Du Phan and Fabian Fuchs and Tobias Weyand and Nenad Tomasev and Alexandra Cordell and Xin Liu and Jonathan Mallinson and Pankaj Joshi and Andy Crawford and Arun Suggala and Steve Chien and Nick Fernando and Mariella Sanchez-Vargas and Duncan Williams and Phil Crone and Xiyang Luo and Igor Karpov and Jyn Shan and Terry Thurk and Robin Strudel and Paul Voigtlaender and Piyush Patil and Tim Dozat and Ali Khodaei and Sahil Singla and Piotr Ambroszczyk and Qiyin Wu and Yifan Chang and Brian Roark and Chaitra Hegde and Tianli Ding and Angelos Filos and Zhongru Wu and André Susano Pinto and Shuang Liu and Saarthak Khanna and Aditya Pandey and Siobhan Mcloughlin and Qiujia Li and Sam Haves and Allan Zhou and Elena Buchatskaya and Isabel Leal and Peter de Boursac and Nami Akazawa and Nina Anderson and Terry Chen and Krishna Somandepalli and Chen Liang and Sheela Goenka and Stephanie Winkler and Alexander Grushetsky and Yifan Ding and Jamie Smith and Fan Ye and Jordi Pont-Tuset and Eric Li and Ruichao Li and Tomer Golany and Dawid Wegner and Tao Jiang and Omer Barak and Yuan Shangguan and Eszter Vértes and Renee Wong and Jörg Bornschein and Alex Tudor and Michele Bevilacqua and Tom Schaul and Ankit Singh Rawat and Yang Zhao and Kyriakos Axiotis and Lei Meng and Cory McLean and Jonathan Lai and Jennifer Beattie and Nate Kushman and Yaxin Liu and Blair Kutzman and Fiona Lang and Jingchen Ye and Praneeth Netrapalli and Pushkar Mishra and Myriam Khan and Megha Goel and Rob Willoughby and David Tian and Honglei Zhuang and JD Chen and Zak Tsai and Tasos Kementsietsidis and Arjun Khare and James Keeling and Keyang Xu and Nathan Waters and Florent Altché and Ashok Popat and Bhavishya Mittal and David Saxton and Dalia El Badawy and Michael Mathieu and Zheng Zheng and Hao Zhou and Nishant Ranka and Richard Shin and Qingnan Duan and Tim Salimans and Ioana Mihailescu and Uri Shaham and Ming-Wei Chang and Yannis Assael and Nishanth Dikkala and Martin Izzard and Vincent Cohen-Addad and Cat Graves and Vlad Feinberg and Grace Chung and DJ Strouse and Danny Karmon and Sahand Sharifzadeh and Zoe Ashwood and Khiem Pham and Jon Blanton and Alex Vasiloff and Jarred Barber and Mark Geller and Aurick Zhou and Fedir Zubach and Tzu-Kuo Huang and Lei Zhang and Himanshu Gupta and Matt Young and Julia Proskurnia and Ronny Votel and Valentin Gabeur and Gabriel Barcik and Aditya Tripathi and Hongkun Yu and Geng Yan and Beer Changpinyo and Filip Pavetić and Amy Coyle and Yasuhisa Fujii and Jorge Gonzalez Mendez and Tianhao Zhou and Harish Rajamani and Blake Hechtman and Eddie Cao and Da-Cheng Juan and Yi-Xuan Tan and Valentin Dalibard and Yilun Du and Natalie Clay and Kaisheng Yao and Wenhao Jia and Dimple Vijaykumar and Yuxiang Zhou and Xinyi Bai and Wei-Chih Hung and Steven Pecht and Georgi Todorov and Nikhil Khadke and Pramod Gupta and Preethi Lahoti and Arnaud Autef and Karthik Duddu and James Lee-Thorp and Alexander Bykovsky and Tautvydas Misiunas and Sebastian Flennerhag and Santhosh Thangaraj and Jed McGiffin and Zack Nado and Markus Kunesch and Andreas Noever and Amir Hertz and Marco Liang and Victor Stone and Evan Palmer and Samira Daruki and Arijit Pramanik and Siim Põder and Austin Kyker and Mina Khan and Evgeny Sluzhaev and Marvin Ritter and Avraham Ruderman and Wenlei Zhou and Chirag Nagpal and Kiran Vodrahalli and George Necula and Paul Barham and Ellie Pavlick and Jay Hartford and Izhak Shafran and Long Zhao and Maciej Mikuła and Tom Eccles and Hidetoshi Shimokawa and Kanav Garg and Luke Vilnis and Hanwen Chen and Ilia Shumailov and Kuang-Huei Lee and Abdelrahman Abdelhamed and Meiyan Xie and Vered Cohen and Ester Hlavnova and Dan Malkin and Chawin Sitawarin and James Lottes and Pauline Coquinot and Tianli Yu and Sandeep Kumar and Jingwei Zhang and Aroma Mahendru and Zafarali Ahmed and James Martens and Tao Chen and Aviel Boag and Daiyi Peng and Coline Devin and Arseniy Klimovskiy and Mary Phuong and Danny Vainstein and Jin Xie and Bhuvana Ramabhadran and Nathan Howard and Xinxin Yu and Gitartha Goswami and Jingyu Cui and Sam Shleifer and Mario Pinto and Chih-Kuan Yeh and Ming-Hsuan Yang and Sara Javanmardi and Dan Ethier and Chace Lee and Jordi Orbay and Suyog Kotecha and Carla Bromberg and Pete Shaw and James Thornton and Adi Gerzi Rosenthal and Shane Gu and Matt Thomas and Ian Gemp and Aditya Ayyar and Asahi Ushio and Aarush Selvan and Joel Wee and Chenxi Liu and Maryam Majzoubi and Weiren Yu and Jake Abernethy and Tyler Liechty and Renke Pan and Hoang Nguyen and Qiong and Hu and Sarah Perrin and Abhinav Arora and Emily Pitler and Weiyi Wang and Kaushik Shivakumar and Flavien Prost and Ben Limonchik and Jing Wang and Yi Gao and Timothee Cour and Shyamal Buch and Huan Gui and Maria Ivanova and Philipp Neubeck and Kelvin Chan and Lucy Kim and Huizhong Chen and Naman Goyal and Da-Woon Chung and Lu Liu and Yao Su and Anastasia Petrushkina and Jiajun Shen and Armand Joulin and Yuanzhong Xu and Stein Xudong Lin and Yana Kulizhskaya and Ciprian Chelba and Shobha Vasudevan and Eli Collins and Vasilisa Bashlovkina and Tony Lu and Doug Fritz and Jongbin Park and Yanqi Zhou and Chen Su and Richard Tanburn and Mikhail Sushkov and Mitchelle Rasquinha and Jinning Li and Jennifer Prendki and Yiming Li and Pallavi LV and Shriya Sharma and Hen Fitoussi and Hui Huang and Andrew Dai and Phuong Dao and Mike Burrows and Henry Prior and Danfeng Qin and Golan Pundak and Lars Lowe Sjoesund and Art Khurshudov and Zhenkai Zhu and Albert Webson and Elizabeth Kemp and Tat Tan and Saurabh Agrawal and Susie Sargsyan and Liqun Cheng and Jim Stephan and Tom Kwiatkowski and David Reid and Arunkumar Byravan and Assaf Hurwitz Michaely and Nicolas Heess and Luowei Zhou and Sonam Goenka and Viral Carpenter and Anselm Levskaya and Bo Wang and Reed Roberts and Rémi Leblond and Sharat Chikkerur and Stav Ginzburg and Max Chang and Robert Riachi and Chuqiao and Xu and Zalán Borsos and Michael Pliskin and Julia Pawar and Morgane Lustman and Hannah Kirkwood and Ankit Anand and Aditi Chaudhary and Norbert Kalb and Kieran Milan and Sean Augenstein and Anna Goldie and Laurel Prince and Karthik Raman and Yanhua Sun and Vivian Xia and Aaron Cohen and Zhouyuan Huo and Josh Camp and Seher Ellis and Lukas Zilka and David Vilar Torres and Lisa Patel and Sho Arora and Betty Chan and Jonas Adler and Kareem Ayoub and Jacky Liang and Fayaz Jamil and Jiepu Jiang and Simon Baumgartner and Haitian Sun and Yael Karov and Yaroslav Akulov and Hui Zheng and Irene Cai and Claudio Fantacci and James Rubin and Alex Rav Acha and Mengchao Wang and Nina D'Souza and Rohit Sathyanarayana and Shengyang Dai and Simon Rowe and Andrey Simanovsky and Omer Goldman and Yuheng Kuang and Xiaoyue Pan and Andrew Rosenberg and Tania Rojas-Esponda and Praneet Dutta and Amy Zeng and Irina Jurenka and Greg Farquhar and Yamini Bansal and Shariq Iqbal and Becca Roelofs and Ga-Young Joung and Parker Beak and Changwan Ryu and Ryan Poplin and Yan Wu and Jean-Baptiste Alayrac and Senaka Buthpitiya and Olaf Ronneberger and Caleb Habtegebriel and Wei Li and Paul Cavallaro and Aurora Wei and Guy Bensky and Timo Denk and Harish Ganapathy and Jeff Stanway and Pratik Joshi and Francesco Bertolini and Jessica Lo and Olivia Ma and Zachary Charles and Geta Sampemane and Himanshu Sahni and Xu Chen and Harry Askham and David Gaddy and Peter Young and Jiewen Tan and Matan Eyal and Arthur Bražinskas and Li Zhong and Zhichun Wu and Mark Epstein and Kai Bailey and Andrew Hard and Kamyu Lee and Sasha Goldshtein and Alex Ruiz and Mohammed Badawi and Matthias Lochbrunner and JK Kearns and Ashley Brown and Fabio Pardo and Theophane Weber and Haichuan Yang and Pan-Pan Jiang and Berkin Akin and Zhao Fu and Marcus Wainwright and Chi Zou and Meenu Gaba and Pierre-Antoine Manzagol and Wendy Kan and Yang Song and Karina Zainullina and Rui Lin and Jeongwoo Ko and Salil Deshmukh and Apoorv Jindal and James Svensson and Divya Tyam and Heri Zhao and Christine Kaeser-Chen and Scott Baird and Pooya Moradi and Jamie Hall and Qiuchen Guo and Vincent Tsang and Bowen Liang and Fernando Pereira and Suhas Ganesh and Ivan Korotkov and Jakub Adamek and Sridhar Thiagarajan and Vinh Tran and Charles Chen and Chris Tar and Sanil Jain and Ishita Dasgupta and Taylan Bilal and David Reitter and Kai Zhao and Giulia Vezzani and Yasmin Gehman and Pulkit Mehta and Lauren Beltrone and Xerxes Dotiwalla and Sergio Guadarrama and Zaheer Abbas and Stefani Karp and Petko Georgiev and Chun-Sung Ferng and Marc Brockschmidt and Liqian Peng and Christoph Hirnschall and Vikas Verma and Yingying Bi and Ying Xiao and Avigail Dabush and Kelvin Xu and Phil Wallis and Randall Parker and Qifei Wang and Yang Xu and Ilkin Safarli and Dinesh Tewari and Yin Zhang and Seungyeon Kim and Andrea Gesmundo and Mackenzie Thomas and Sergey Levi and Ahmed Chowdhury and Kanishka Rao and Peter Garst and Sam Conway-Rahman and Helen Ran and Kay McKinney and Zhisheng Xiao and Wenhao Yu and Rohan Agrawal and Axel Stjerngren and Catalin Ionescu and Jingjing Chen and Vivek Sharma and Justin Chiu and Fei Liu and Ken Franko and Clayton Sanford and Xingyu Cai and Paul Michel and Sanjay Ganapathy and Jane Labanowski and Zachary Garrett and Ben Vargas and Sean Sun and Bryan Gale and Thomas Buschmann and Guillaume Desjardins and Nimesh Ghelani and Palak Jain and Mudit Verma and Chulayuth Asawaroengchai and Julian Eisenschlos and Jitendra Harlalka and Hideto Kazawa and Don Metzler and Joshua Howland and Ying Jian and Jake Ades and Viral Shah and Tynan Gangwani and Seungji Lee and Roman Ring and Steven M. Hernandez and Dean Reich and Amer Sinha and Ashutosh Sathe and Joe Kovac and Ashleah Gill and Ajay Kannan and Andrea D'olimpio and Martin Sevenich and Jay Whang and Been Kim and Khe Chai Sim and Jilin Chen and Jiageng Zhang and Shuba Lall and Yossi Matias and Bill Jia and Abe Friesen and Sara Nasso and Ashish Thapliyal and Bryan Perozzi and Ting Yu and Anna Shekhawat and Safeen Huda and Peter Grabowski and Eric Wang and Ashwin Sreevatsa and Hilal Dib and Mehadi Hassen and Parker Schuh and Vedrana Milutinovic and Chris Welty and Michael Quinn and Ali Shah and Bangju Wang and Gabe Barth-Maron and Justin Frye and Natalie Axelsson and Tao Zhu and Yukun Ma and Irene Giannoumis and Hanie Sedghi and Chang Ye and Yi Luan and Kevin Aydin and Bilva Chandra and Vivek Sampathkumar and Ronny Huang and Victor Lavrenko and Ahmed Eleryan and Zhi Hong and Steven Hansen and Sara Mc Carthy and Bidisha Samanta and Domagoj Ćevid and Xin Wang and Fangtao Li and Michael Voznesensky and Matt Hoffman and Andreas Terzis and Vikash Sehwag and Gil Fidel and Luheng He and Mu Cai and Yanzhang He and Alex Feng and Martin Nikoltchev and Samrat Phatale and Jason Chase and Rory Lawton and Ming Zhang and Tom Ouyang and Manuel Tragut and Mehdi Hafezi Manshadi and Arjun Narayanan and Jiaming Shen and Xu Gao and Tolga Bolukbasi and Nick Roy and Xin Li and Daniel Golovin and Liviu Panait and Zhen Qin and Guangxing Han and Thomas Anthony and Sneha Kudugunta and Viorica Patraucean and Aniket Ray and Xinyun Chen and Xiaochen Yang and Tanuj Bhatia and Pranav Talluri and Alex Morris and Andrija Ražnatović and Bethanie Brownfield and James An and Sheng Peng and Patrick Kane and Ce Zheng and Nico Duduta and Joshua Kessinger and James Noraky and Siqi Liu and Keran Rong and Petar Veličković and Keith Rush and Alex Goldin and Fanny Wei and Shiva Mohan Reddy Garlapati and Caroline Pantofaru and Okwan Kwon and Jianmo Ni and Eric Noland and Julia Di Trapani and Françoise Beaufays and Abhijit Guha Roy and Yinlam Chow and Aybuke Turker and Geoffrey Cideron and Lantao Mei and Jon Clark and Qingyun Dou and Matko Bošnjak and Ralph Leith and Yuqing Du and Amir Yazdanbakhsh and Milad Nasr and Chester Kwak and Suraj Satishkumar Sheth and Alex Kaskasoli and Ankesh Anand and Balaji Lakshminarayanan and Sammy Jerome and David Bieber and Chun-Te Chu and Alexandre Senges and Tianxiao Shen and Mukund Sridhar and Ndaba Ndebele and Benjamin Beyret and Shakir Mohamed and Mia Chen and Markus Freitag and Jiaxian Guo and Luyang Liu and Paul Roit and Heng Chen and Shen Yan and Tom Stone and JD Co-Reyes and Jeremy Cole and Salvatore Scellato and Shekoofeh Azizi and Hadi Hashemi and Alicia Jin and Anand Iyer and Marcella Valentine and András György and Arun Ahuja and Daniel Hernandez Diaz and Chen-Yu Lee and Nathan Clement and Weize Kong and Drew Garmon and Ishaan Watts and Kush Bhatia and Khyatti Gupta and Matt Miecnikowski and Hugo Vallet and Ankur Taly and Edward Loper and Saket Joshi and James Atwood and Jo Chick and Mark Collier and Fotis Iliopoulos and Ryan Trostle and Beliz Gunel and Ramiro Leal-Cavazos and Arnar Mar Hrafnkelsson and Michael Guzman and Xiaoen Ju and Andy Forbes and Jesse Emond and Kushal Chauhan and Ben Caine and Li Xiao and Wenjun Zeng and Alexandre Moufarek and Daniel Murphy and Maya Meng and Nitish Gupta and Felix Riedel and Anil Das and Elijah Lawal and Shashi Narayan and Tiberiu Sosea and James Swirhun and Linda Friso and Behnam Neyshabur and Jing Lu and Sertan Girgin and Michael Wunder and Edouard Yvinec and Aroonalok Pyne and Victor Carbune and Shruti Rijhwani and Yang Guo and Tulsee Doshi and Anton Briukhov and Max Bain and Ayal Hitron and Xuanhui Wang and Ashish Gupta and Ke Chen and Cosmo Du and Weiyang Zhang and Dhruv Shah and Arjun Akula and Max Dylla and Ashyana Kachra and Weicheng Kuo and Tingting Zou and Lily Wang and Luyao Xu and Jifan Zhu and Justin Snyder and Sachit Menon and Orhan Firat and Igor Mordatch and Yuan Yuan and Natalia Ponomareva and Rory Blevins and Lawrence Moore and Weijun Wang and Phil Chen and Martin Scholz and Artur Dwornik and Jason Lin and Sicheng Li and Diego Antognini and Te I and Xiaodan Song and Matt Miller and Uday Kalra and Adam Raveret and Oscar Akerlund and Felix Wu and Andrew Nystrom and Namrata Godbole and Tianqi Liu and Hannah DeBalsi and Jewel Zhao and Buhuang Liu and Avi Caciularu and Lauren Lax and Urvashi Khandelwal and Victoria Langston and Eric Bailey and Silvio Lattanzi and Yufei Wang and Neel Kovelamudi and Sneha Mondal and Guru Guruganesh and Nan Hua and Ofir Roval and Paweł Wesołowski and Rishikesh Ingale and Jonathan Halcrow and Tim Sohn and Christof Angermueller and Bahram Raad and Eli Stickgold and Eva Lu and Alec Kosik and Jing Xie and Timothy Lillicrap and Austin Huang and Lydia Lihui Zhang and Dominik Paulus and Clement Farabet and Alex Wertheim and Bing Wang and Rishabh Joshi and Chu-ling Ko and Yonghui Wu and Shubham Agrawal and Lily Lin and XiangHai Sheng and Peter Sung and Tyler Breland-King and Christina Butterfield and Swapnil Gawde and Sumeet Singh and Qiao Zhang and Raj Apte and Shilpa Shetty and Adrian Hutter and Tao Li and Elizabeth Salesky and Federico Lebron and Jonni Kanerva and Michela Paganini and Arthur Nguyen and Rohith Vallu and Jan-Thorsten Peter and Sarmishta Velury and David Kao and Jay Hoover and Anna Bortsova and Colton Bishop and Shoshana Jakobovits and Alessandro Agostini and Alekh Agarwal and Chang Liu and Charles Kwong and Sasan Tavakkol and Ioana Bica and Alex Greve and Anirudh GP and Jake Marcus and Le Hou and Tom Duerig and Rivka Moroshko and Dave Lacey and Andy Davis and Julien Amelot and Guohui Wang and Frank Kim and Theofilos Strinopoulos and Hui Wan and Charline Le Lan and Shankar Krishnan and Haotian Tang and Peter Humphreys and Junwen Bai and Idan Heimlich Shtacher and Diego Machado and Chenxi Pang and Ken Burke and Dangyi Liu and Renga Aravamudhan and Yue Song and Ed Hirst and Abhimanyu Singh and Brendan Jou and Liang Bai and Francesco Piccinno and Chuyuan Kelly Fu and Robin Alazard and Barak Meiri and Daniel Winter and Charlie Chen and Mingda Zhang and Jens Heitkaemper and John Lambert and Jinhyuk Lee and Alexander Frömmgen and Sergey Rogulenko and Pranav Nair and Paul Niemczyk and Anton Bulyenov and Bibo Xu and Hadar Shemtov and Morteza Zadimoghaddam and Serge Toropov and Mateo Wirth and Hanjun Dai and Sreenivas Gollapudi and Daniel Zheng and Alex Kurakin and Chansoo Lee and Kalesha Bullard and Nicolas Serrano and Ivana Balazevic and Yang Li and Johan Schalkwyk and Mark Murphy and Mingyang Zhang and Kevin Sequeira and Romina Datta and Nishant Agrawal and Charles Sutton and Nithya Attaluri and Mencher Chiang and Wael Farhan and Gregory Thornton and Kate Lin and Travis Choma and Hung Nguyen and Kingshuk Dasgupta and Dirk Robinson and Iulia Comşa and Michael Riley and Arjun Pillai and Basil Mustafa and Ben Golan and Amir Zandieh and Jean-Baptiste Lespiau and Billy Porter and David Ross and Sujeevan Rajayogam and Mohit Agarwal and Subhashini Venugopalan and Bobak Shahriari and Qiqi Yan and Hao Xu and Taylor Tobin and Pavel Dubov and Hongzhi Shi and Adrià Recasens and Anton Kovsharov and Sebastian Borgeaud and Lucio Dery and Shanthal Vasanth and Elena Gribovskaya and Linhai Qiu and Mahdis Mahdieh and Wojtek Skut and Elizabeth Nielsen and CJ Zheng and Adams Yu and Carrie Grimes Bostock and Shaleen Gupta and Aaron Archer and Chris Rawles and Elinor Davies and Alexey Svyatkovskiy and Tomy Tsai and Yoni Halpern and Christian Reisswig and Bartek Wydrowski and Bo Chang and Joan Puigcerver and Mor Hazan Taege and Jian Li and Eva Schnider and Xinjian Li and Dragos Dena and Yunhan Xu and Umesh Telang and Tianze Shi and Heiga Zen and Kyle Kastner and Yeongil Ko and Neesha Subramaniam and Aviral Kumar and Pete Blois and Zhuyun Dai and John Wieting and Yifeng Lu and Yoel Zeldes and Tian Xie and Anja Hauth and Alexandru Ţifrea and Yuqi Li and Sam El-Husseini and Dan Abolafia and Howard Zhou and Wen Ding and Sahra Ghalebikesabi and Carlos Guía and Andrii Maksai and Ágoston Weisz and Sercan Arik and Nick Sukhanov and Aga Świetlik and Xuhui Jia and Luo Yu and Weiyue Wang and Mark Brand and Dawn Bloxwich and Sean Kirmani and Zhe Chen and Alec Go and Pablo Sprechmann and Nithish Kannen and Alen Carin and Paramjit Sandhu and Isabel Edkins and Leslie Nooteboom and Jai Gupta and Loren Maggiore and Javad Azizi and Yael Pritch and Pengcheng Yin and Mansi Gupta and Danny Tarlow and Duncan Smith and Desi Ivanov and Mohammad Babaeizadeh and Ankita Goel and Satish Kambala and Grace Chu and Matej Kastelic and Michelle Liu and Hagen Soltau and Austin Stone and Shivani Agrawal and Min Kim and Kedar Soparkar and Srinivas Tadepalli and Oskar Bunyan and Rachel Soh and Arvind Kannan and DY Kim and Blake JianHang Chen and Afief Halumi and Sudeshna Roy and Yulong Wang and Olcan Sercinoglu and Gena Gibson and Sijal Bhatnagar and Motoki Sano and Daniel von Dincklage and Qingchun Ren and Blagoj Mitrevski and Mirek Olšák and Jennifer She and Carl Doersch and Jilei and Wang and Bingyuan Liu and Qijun Tan and Tamar Yakar and Tris Warkentin and Alex Ramirez and Carl Lebsack and Josh Dillon and Rajiv Mathews and Tom Cobley and Zelin Wu and Zhuoyuan Chen and Jon Simon and Swaroop Nath and Tara Sainath and Alexei Bendebury and Ryan Julian and Bharath Mankalale and Daria Ćurko and Paulo Zacchello and Adam R. Brown and Kiranbir Sodhia and Heidi Howard and Sergi Caelles and Abhinav Gupta and Gareth Evans and Anna Bulanova and Lesley Katzen and Roman Goldenberg and Anton Tsitsulin and Joe Stanton and Benoit Schillings and Vitaly Kovalev and Corey Fry and Rushin Shah and Kuo Lin and Shyam Upadhyay and Cheng Li and Soroush Radpour and Marcello Maggioni and Jing Xiong and Lukas Haas and Jenny Brennan and Aishwarya Kamath and Nikolay Savinov and Arsha Nagrani and Trevor Yacovone and Ryan Kappedal and Kostas Andriopoulos and Li Lao and YaGuang Li and Grigory Rozhdestvenskiy and Kazuma Hashimoto and Andrew Audibert and Sophia Austin and Daniel Rodriguez and Anian Ruoss and Garrett Honke and Deep Karkhanis and Xi Xiong and Qing Wei and James Huang and Zhaoqi Leng and Vittal Premachandran and Stan Bileschi and Georgios Evangelopoulos and Thomas Mensink and Jay Pavagadhi and Denis Teplyashin and Paul Chang and Linting Xue and Garrett Tanzer and Sally Goldman and Kaushal Patel and Shixin Li and Jeremy Wiesner and Ivy Zheng and Ian Stewart-Binks and Jie Han and Zhi Li and Liangchen Luo and Karel Lenc and Mario Lučić and Fuzhao Xue and Ryan Mullins and Alexey Guseynov and Chung-Ching Chang and Isaac Galatzer-Levy and Adam Zhang and Garrett Bingham and Grace Hu and Ale Hartman and Yue Ma and Jordan Griffith and Alex Irpan and Carey Radebaugh and Summer Yue and Lijie Fan and Victor Ungureanu and Christina Sorokin and Hannah Teufel and Peiran Li and Rohan Anil and Dimitris Paparas and Todd Wang and Chu-Cheng Lin and Hui Peng and Megan Shum and Goran Petrovic and Demetra Brady and Richard Nguyen and Klaus Macherey and Zhihao Li and Harman Singh and Madhavi Yenugula and Mariko Iinuma and Xinyi Chen and Kavya Kopparapu and Alexey Stern and Shachi Dave and Chandu Thekkath and Florence Perot and Anurag Kumar and Fangda Li and Yang Xiao and Matthew Bilotti and Mohammad Hossein Bateni and Isaac Noble and Lisa Lee and Amelio Vázquez-Reina and Julian Salazar and Xiaomeng Yang and Boyu Wang and Ela Gruzewska and Anand Rao and Sindhu Raghuram and Zheng Xu and Eyal Ben-David and Jieru Mei and Sid Dalmia and Zhaoyi Zhang and Yuchen Liu and Gagan Bansal and Helena Pankov and Steven Schwarcz and Andrea Burns and Christine Chan and Sumit Sanghai and Ricky Liang and Ethan Liang and Antoine He and Amy Stuart and Arun Narayanan and Yukun Zhu and Christian Frank and Bahar Fatemi and Amit Sabne and Oran Lang and Indro Bhattacharya and Shane Settle and Maria Wang and Brendan McMahan and Andrea Tacchetti and Livio Baldini Soares and Majid Hadian and Serkan Cabi and Timothy Chung and Nikita Putikhin and Gang Li and Jeremy Chen and Austin Tarango and Henryk Michalewski and Mehran Kazemi and Hussain Masoom and Hila Sheftel and Rakesh Shivanna and Archita Vadali and Ramona Comanescu and Doug Reid and Joss Moore and Arvind Neelakantan and Michaël Sander and Jonathan Herzig and Aviv Rosenberg and Mostafa Dehghani and JD Choi and Michael Fink and Reid Hayes and Eric Ge and Shitao Weng and Chia-Hua Ho and John Karro and Kalpesh Krishna and Lam Nguyen Thiet and Amy Skerry-Ryan and Daniel Eppens and Marco Andreetto and Navin Sarma and Silvano Bonacina and Burcu Karagol Ayan and Megha Nawhal and Zhihao Shan and Mike Dusenberry and Shantanu Thakoor and Sagar Gubbi and Duc Dung Nguyen and Reut Tsarfaty and Samuel Albanie and Jovana Mitrović and Meet Gandhi and Bo-Juen Chen and Alessandro Epasto and Georgi Stephanov and Ye Jin and Samuel Gehman and Aida Amini and Jack Weber and Feryal Behbahani and Shawn Xu and Miltos Allamanis and Xi Chen and Myle Ott and Claire Sha and Michal Jastrzebski and Hang Qi and David Greene and Xinyi Wu and Abodunrinwa Toki and Daniel Vlasic and Jane Shapiro and Ragha Kotikalapudi and Zhe Shen and Takaaki Saeki and Sirui Xie and Albin Cassirer and Shikhar Bharadwaj and Tatsuya Kiyono and Srinadh Bhojanapalli and Elan Rosenfeld and Sam Ritter and Jieming Mao and João Gabriel Oliveira and Zoltan Egyed and Bernd Bandemer and Emilio Parisotto and Keisuke Kinoshita and Juliette Pluto and Petros Maniatis and Steve Li and Yaohui Guo and Golnaz Ghiasi and Jean Tarbouriech and Srimon Chatterjee and Julie Jin and Katrina and Xu and Jennimaria Palomaki and Séb Arnold and Madhavi Sewak and Federico Piccinini and Mohit Sharma and Ben Albrecht and Sean Purser-haskell and Ashwin Vaswani and Chongyan Chen and Matheus Wisniewski and Qin Cao and John Aslanides and Nguyet Minh Phu and Maximilian Sieb and Lauren Agubuzu and Anne Zheng and Daniel Sohn and Marco Selvi and Anders Andreassen and Krishan Subudhi and Prem Eruvbetine and Oliver Woodman and Tomas Mery and Sebastian Krause and Xiaoqi Ren and Xiao Ma and Jincheng Luo and Dawn Chen and Wei Fan and Henry Griffiths and Christian Schuler and Alice Li and Shujian Zhang and Jean-Michel Sarr and Shixin Luo and Riccardo Patana and Matthew Watson and Dani Naboulsi and Michael Collins and Sailesh Sidhwani and Emiel Hoogeboom and Sharon Silver and Emily Caveness and Xiaokai Zhao and Mikel Rodriguez and Maxine Deines and Libin Bai and Patrick Griffin and Marco Tagliasacchi and Emily Xue and Spandana Raj Babbula and Bo Pang and Nan Ding and Gloria Shen and Elijah Peake and Remi Crocker and Shubha Srinivas Raghvendra and Danny Swisher and Woohyun Han and Richa Singh and Ling Wu and Vladimir Pchelin and Tsendsuren Munkhdalai and Dana Alon and Geoff Bacon and Efren Robles and Jannis Bulian and Melvin Johnson and George Powell and Felipe Tiengo Ferreira and Yaoyiran Li and Frederik Benzing and Mihajlo Velimirović and Hubert Soyer and William Kong and Tony and Nguyên and Zhen Yang and Jeremiah Liu and Joost van Amersfoort and Daniel Gillick and Baochen Sun and Nathalie Rauschmayr and Katie Zhang and Serena Zhan and Tao Zhou and Alexey Frolov and Chengrun Yang and Denis Vnukov and Louis Rouillard and Hongji Li and Amol Mandhane and Nova Fallen and Rajesh Venkataraman and Clara Huiyi Hu and Jennifer Brennan and Jenny Lee and Jerry Chang and Martin Sundermeyer and Zhufeng Pan and Rosemary Ke and Simon Tong and Alex Fabrikant and William Bono and Jindong Gu and Ryan Foley and Yiran Mao and Manolis Delakis and Dhruva Bhaswar and Roy Frostig and Nick Li and Avital Zipori and Cath Hope and Olga Kozlova and Swaroop Mishra and Josip Djolonga and Craig Schiff and Majd Al Merey and Eleftheria Briakou and Peter Morgan and Andy Wan and Avinatan Hassidim and RJ Skerry-Ryan and Kuntal Sengupta and Mary Jasarevic and Praveen Kallakuri and Paige Kunkle and Hannah Brennan and Tom Lieber and Hassan Mansoor and Julian Walker and Bing Zhang and Annie Xie and Goran Žužić and Adaeze Chukwuka and Alex Druinsky and Donghyun Cho and Rui Yao and Ferjad Naeem and Shiraz Butt and Eunyoung Kim and Zhipeng Jia and Mandy Jordan and Adam Lelkes and Mark Kurzeja and Sophie Wang and James Zhao and Andrew Over and Abhishek Chakladar and Marcel Prasetya and Neha Jha and Sriram Ganapathy and Yale Cong and Prakash Shroff and Carl Saroufim and Sobhan Miryoosefi and Mohamed Hammad and Tajwar Nasir and Weijuan Xi and Yang Gao and Young Maeng and Ben Hora and Chin-Yi Cheng and Parisa Haghani and Yoad Lewenberg and Caden Lu and Martin Matysiak and Naina Raisinghani and Huiyu Wang and Lexi Baugher and Rahul Sukthankar and Minh Giang and John Schultz and Noah Fiedel and Minmin Chen and Cheng-Chun Lee and Tapomay Dey and Hao Zheng and Shachi Paul and Celine Smith and Andy Ly and Yicheng Wang and Rishabh Bansal and Bartek Perz and Susanna Ricco and Stasha Blank and Vaishakh Keshava and Deepak Sharma and Marvin Chow and Kunal Lad and Komal Jalan and Simon Osindero and Craig Swanson and Jacob Scott and Anastasija Ilić and Xiaowei Li and Siddhartha Reddy Jonnalagadda and Afzal Shama Soudagar and Yan Xiong and Bat-Orgil Batsaikhan and Daniel Jarrett and Naveen Kumar and Maulik Shah and Matt Lawlor and Austin Waters and Mark Graham and Rhys May and Sabela Ramos and Sandra Lefdal and Zeynep Cankara and Nacho Cano and Brendan O'Donoghue and Jed Borovik and Frederick Liu and Jordan Grimstad and Mahmoud Alnahlawi and Katerina Tsihlas and Tom Hudson and Nikolai Grigorev and Yiling Jia and Terry Huang and Tobenna Peter Igwe and Sergei Lebedev and Xiaodan Tang and Igor Krivokon and Frankie Garcia and Melissa Tan and Eric Jia and Peter Stys and Shikhar Vashishth and Yu Liang and Balaji Venkatraman and Chenjie Gu and Anastasios Kementsietsidis and Chen Zhu and Junehyuk Jung and Yunfei Bai and Mohammad Javad Hosseini and Faruk Ahmed and Aditya Gupta and Xin Yuan and Shereen Ashraf and Shitij Nigam and Gautam Vasudevan and Pranjal Awasthi and Adi Mayrav Gilady and Zelda Mariet and Ramy Eskander and Haiguang Li and Hexiang Hu and Guillermo Garrido and Philippe Schlattner and George Zhang and Rohun Saxena and Petar Dević and Kritika Muralidharan and Ashwin Murthy and Yiqian Zhou and Min Choi and Arissa Wongpanich and Zhengdong Wang and Premal Shah and Yuntao Xu and Yiling Huang and Stephen Spencer and Alice Chen and James Cohan and Junjie Wang and Jonathan Tompson and Junru Wu and Ruba Haroun and Haiqiong Li and Blanca Huergo and Fan Yang and Tongxin Yin and James Wendt and Michael Bendersky and Rahma Chaabouni and Javier Snaider and Johan Ferret and Abhishek Jindal and Tara Thompson and Andrew Xue and Will Bishop and Shubham Milind Phal and Archit Sharma and Yunhsuan Sung and Prabakar Radhakrishnan and Mo Shomrat and Reeve Ingle and Roopali Vij and Justin Gilmer and Mihai Dorin Istin and Sam Sobell and Yang Lu and Emily Nottage and Dorsa Sadigh and Jeremiah Willcock and Tingnan Zhang and Steve Xu and Sasha Brown and Katherine Lee and Gary Wang and Yun Zhu and Yi Tay and Cheolmin Kim and Audrey Gutierrez and Abhanshu Sharma and Yongqin Xian and Sungyong Seo and Claire Cui and Elena Pochernina and Cip Baetu and Krzysztof Jastrzębski and Mimi Ly and Mohamed Elhawaty and Dan Suh and Eren Sezener and Pidong Wang and Nancy Yuen and George Tucker and Jiahao Cai and Zuguang Yang and Cindy Wang and Alex Muzio and Hai Qian and Jae Yoo and Derek Lockhart and Kevin R. McKee and Mandy Guo and Malika Mehrotra and Artur Mendonça and Sanket Vaibhav Mehta and Sherry Ben and Chetan Tekur and Jiaqi Mu and Muye Zhu and Victoria Krakovna and Hongrae Lee and AJ Maschinot and Sébastien Cevey and HyunJeong Choe and Aijun Bai and Hansa Srinivasan and Derek Gasaway and Nick Young and Patrick Siegler and Dan Holtmann-Rice and Vihari Piratla and Kate Baumli and Roey Yogev and Alex Hofer and Hado van Hasselt and Svetlana Grant and Yuri Chervonyi and David Silver and Andrew Hogue and Ayushi Agarwal and Kathie Wang and Preeti Singh and Four Flynn and Josh Lipschultz and Robert David and Lizzetth Bellot and Yao-Yuan Yang and Long Le and Filippo Graziano and Kate Olszewska and Kevin Hui and Akanksha Maurya and Nikos Parotsidis and Weijie Chen and Tayo Oguntebi and Joe Kelley and Anirudh Baddepudi and Johannes Mauerer and Gregory Shaw and Alex Siegman and Lin Yang and Shravya Shetty and Subhrajit Roy and Yunting Song and Wojciech Stokowiec and Ryan Burnell and Omkar Savant and Robert Busa-Fekete and Jin Miao and Samrat Ghosh and Liam MacDermed and Phillip Lippe and Mikhail Dektiarev and Zach Behrman and Fabian Mentzer and Kelvin Nguyen and Meng Wei and Siddharth Verma and Chris Knutsen and Sudeep Dasari and Zhipeng Yan and Petr Mitrichev and Xingyu Wang and Virat Shejwalkar and Jacob Austin and Srinivas Sunkara and Navneet Potti and Yan Virin and Christian Wright and Gaël Liu and Oriana Riva and Etienne Pot and Greg Kochanski and Quoc Le and Gargi Balasubramaniam and Arka Dhar and Yuguo Liao and Adam Bloniarz and Divyansh Shukla and Elizabeth Cole and Jong Lee and Sheng Zhang and Sushant Kafle and Siddharth Vashishtha and Parsa Mahmoudieh and Grace Chen and Raphael Hoffmann and Pranesh Srinivasan and Agustin Dal Lago and Yoav Ben Shalom and Zi Wang and Michael Elabd and Anuj Sharma and Junhyuk Oh and Suraj Kothawade and Maigo Le and Marianne Monteiro and Shentao Yang and Kaiz Alarakyia and Robert Geirhos and Diana Mincu and Håvard Garnes and Hayato Kobayashi and Soroosh Mariooryad and Kacper Krasowiak and Zhixin and Lai and Shibl Mourad and Mingqiu Wang and Fan Bu and Ophir Aharoni and Guanjie Chen and Abhimanyu Goyal and Vadim Zubov and Ankur Bapna and Elahe Dabir and Nisarg Kothari and Kay Lamerigts and Nicola De Cao and Jeremy Shar and Christopher Yew and Nitish Kulkarni and Dre Mahaarachchi and Mandar Joshi and Zhenhai Zhu and Jared Lichtarge and Yichao Zhou and Hannah Muckenhirn and Vittorio Selo and Oriol Vinyals and Peter Chen and Anthony Brohan and Vaibhav Mehta and Sarah Cogan and Ruth Wang and Ty Geri and Wei-Jen Ko and Wei Chen and Fabio Viola and Keshav Shivam and Lisa Wang and Madeleine Clare Elish and Raluca Ada Popa and Sébastien Pereira and Jianqiao Liu and Raphael Koster and Donnie Kim and Gufeng Zhang and Sayna Ebrahimi and Partha Talukdar and Yanyan Zheng and Petra Poklukar and Ales Mikhalap and Dale Johnson and Anitha Vijayakumar and Mark Omernick and Matt Dibb and Ayush Dubey and Qiong Hu and Apurv Suman and Vaibhav Aggarwal and Ilya Kornakov and Fei Xia and Wing Lowe and Alexey Kolganov and Ted Xiao and Vitaly Nikolaev and Steven Hemingray and Bonnie Li and Joana Iljazi and Mikołaj Rybiński and Ballie Sandhu and Peggy Lu and Thang Luong and Rodolphe Jenatton and Vineetha Govindaraj and Hui and Li and Gabriel Dulac-Arnold and Wonpyo Park and Henry Wang and Abhinit Modi and Jean Pouget-Abadie and Kristina Greller and Rahul Gupta and Robert Berry and Prajit Ramachandran and Jinyu Xie and Liam McCafferty and Jianling Wang and Kilol Gupta and Hyeontaek Lim and Blaž Bratanič and Andy Brock and Ilia Akolzin and Jim Sproch and Dan Karliner and Duhyeon Kim and Adrian Goedeckemeyer and Noam Shazeer and Cordelia Schmid and Daniele Calandriello and Parul Bhatia and Krzysztof Choromanski and Ceslee Montgomery and Dheeru Dua and Ana Ramalho and Helen King and Yue Gao and Lynn Nguyen and David Lindner and Divya Pitta and Oleaser Johnson and Khalid Salama and Diego Ardila and Michael Han and Erin Farnese and Seth Odoom and Ziyue Wang and Xiangzhuo Ding and Norman Rink and Ray Smith and Harshal Tushar Lehri and Eden Cohen and Neera Vats and Tong He and Parthasarathy Gopavarapu and Adam Paszke and Miteyan Patel and Wouter Van Gansbeke and Lucia Loher and Luis Castro and Maria Voitovich and Tamara von Glehn and Nelson George and Simon Niklaus and Zach Eaton-Rosen and Nemanja Rakićević and Erik Jue and Sagi Perel and Carrie Zhang and Yuval Bahat and Angéline Pouget and Zhi Xing and Fantine Huot and Ashish Shenoy and Taylor Bos and Vincent Coriou and Bryan Richter and Natasha Noy and Yaqing Wang and Santiago Ontanon and Siyang Qin and Gleb Makarchuk and Demis Hassabis and Zhuowan Li and Mandar Sharma and Kumaran Venkatesan and Iurii Kemaev and Roxanne Daniel and Shiyu Huang and Saloni Shah and Octavio Ponce and Warren and Chen and Manaal Faruqui and Jialin Wu and Slavica Andačić and Szabolcs Payrits and Daniel McDuff and Tom Hume and Yuan Cao and MH Tessler and Qingze Wang and Yinan Wang and Ivor Rendulic and Eirikur Agustsson and Matthew Johnson and Tanya Lando and Andrew Howard and Sri Gayatri Sundara Padmanabhan and Mayank Daswani and Andrea Banino and Michael Kilgore and Jonathan Heek and Ziwei Ji and Alvaro Caceres and Conglong Li and Nora Kassner and Alexey Vlaskin and Zeyu Liu and Alex Grills and Yanhan Hou and Roykrong Sukkerd and Gowoon Cheon and Nishita Shetty and Larisa Markeeva and Piotr Stanczyk and Tejas Iyer and Yuan Gong and Shawn Gao and Keerthana Gopalakrishnan and Tim Blyth and Malcolm Reynolds and Avishkar Bhoopchand and Misha Bilenko and Dero Gharibian and Vicky Zayats and Aleksandra Faust and Abhinav Singh and Min Ma and Hongyang Jiao and Sudheendra Vijayanarasimhan and Lora Aroyo and Vikas Yadav and Sarah Chakera and Ashwin Kakarla and Vilobh Meshram and Karol Gregor and Gabriela Botea and Evan Senter and Dawei Jia and Geza Kovacs and Neha Sharma and Sebastien Baur and Kai Kang and Yifan He and Lin Zhuo and Marija Kostelac and Itay Laish and Songyou Peng and Louis O'Bryan and Daniel Kasenberg and Girish Ramchandra Rao and Edouard Leurent and Biao Zhang and Sage Stevens and Ana Salazar and Ye Zhang and Ivan Lobov and Jake Walker and Allen Porter and Morgan Redshaw and Han Ke and Abhishek Rao and Alex Lee and Hoi Lam and Michael Moffitt and Jaeyoun Kim and Siyuan Qiao and Terry Koo and Robert Dadashi and Xinying Song and Mukund Sundararajan and Peng Xu and Chizu Kawamoto and Yan Zhong and Clara Barbu and Apoorv Reddy and Mauro Verzetti and Leon Li and George Papamakarios and Hanna Klimczak-Plucińska and Mary Cassin and Koray Kavukcuoglu and Rigel Swavely and Alain Vaucher and Jeffrey Zhao and Ross Hemsley and Michael Tschannen and Heming Ge and Gaurav Menghani and Yang Yu and Natalie Ha and Wei He and Xiao Wu and Maggie Song and Rachel Sterneck and Stefan Zinke and Dan A. Calian and Annie Marsden and Alejandro Cruzado Ruiz and Matteo Hessel and Almog Gueta and Benjamin Lee and Brian Farris and Manish Gupta and Yunjie Li and Mohammad Saleh and Vedant Misra and Kefan Xiao and Piermaria Mendolicchio and Gavin Buttimore and Varvara Krayvanova and Nigamaa Nayakanti and Matthew Wiethoff and Yash Pande and Azalia Mirhoseini and Ni Lao and Jasmine Liu and Yiqing Hua and Angie Chen and Yury Malkov and Dmitry Kalashnikov and Shubham Gupta and Kartik Audhkhasi and Yuexiang Zhai and Sudhindra Kopalle and Prateek Jain and Eran Ofek and Clemens Meyer and Khuslen Baatarsukh and Hana Strejček and Jun Qian and James Freedman and Ricardo Figueira and Michal Sokolik and Olivier Bachem and Raymond Lin and Dia Kharrat and Chris Hidey and Pingmei Xu and Dennis Duan and Yin Li and Muge Ersoy and Richard Everett and Kevin Cen and Rebeca Santamaria-Fernandez and Amir Taubenfeld and Ian Mackinnon and Linda Deng and Polina Zablotskaia and Shashank Viswanadha and Shivanker Goel and Damion Yates and Yunxiao Deng and Peter Choy and Mingqing Chen and Abhishek Sinha and Alex Mossin and Yiming Wang and Arthur Szlam and Susan Hao and Paul Kishan Rubenstein and Metin Toksoz-Exley and Miranda Aperghis and Yin Zhong and Junwhan Ahn and Michael Isard and Olivier Lacombe and Florian Luisier and Chrysovalantis Anastasiou and Yogesh Kalley and Utsav Prabhu and Emma Dunleavy and Shaan Bijwadia and Justin Mao-Jones and Kelly Chen and Rama Pasumarthi and Emily Wood and Adil Dostmohamed and Nate Hurley and Jiri Simsa and Alicia Parrish and Mantas Pajarskas and Matt Harvey and Ondrej Skopek and Yony Kochinski and Javier Rey and Verena Rieser and Denny Zhou and Sun Jae Lee and Trilok Acharya and Guowang Li and Joe Jiang and Xiaofan Zhang and Bryant Gipson and Ethan Mahintorabi and Marco Gelmi and Nima Khajehnouri and Angel Yeh and Kayi Lee and Loic Matthey and Leslie Baker and Trang Pham and Han Fu and Alex Pak and Prakhar Gupta and Cristina Vasconcelos and Adam Sadovsky and Brian Walker and Sissie Hsiao and Patrik Zochbauer and Andreea Marzoca and Noam Velan and Junhao Zeng and Gilles Baechler and Danny Driess and Divya Jain and Yanping Huang and Lizzie Tao and John Maggs and Nir Levine and Jon Schneider and Erika Gemzer and Samuel Petit and Shan Han and Zach Fisher and Dustin Zelle and Courtney Biles and Eugene Ie and Asya Fadeeva and Casper Liu and Juliana Vicente Franco and Adrian Collister and Hao Zhang and Renshen Wang and Ruizhe Zhao and Leandro Kieliger and Kurt Shuster and Rui Zhu and Boqing Gong and Lawrence Chan and Ruoxi Sun and Sujoy Basu and Roland Zimmermann and Jamie Hayes and Abhishek Bapna and Jasper Snoek and Weel Yang and Puranjay Datta and Jad Al Abdallah and Kevin Kilgour and Lu Li and SQ Mah and Yennie Jun and Morgane Rivière and Abhijit Karmarkar and Tammo Spalink and Tao Huang and Lucas Gonzalez and Duc-Hieu Tran and Averi Nowak and John Palowitch and Martin Chadwick and Ellie Talius and Harsh Mehta and Thibault Sellam and Philipp Fränken and Massimo Nicosia and Kyle He and Aditya Kini and David Amos and Sugato Basu and Harrison Jobe and Eleni Shaw and Qiantong Xu and Colin Evans and Daisuke Ikeda and Chaochao Yan and Larry Jin and Lun Wang and Sachin Yadav and Ilia Labzovsky and Ramesh Sampath and Ada Ma and Candice Schumann and Aditya Siddhant and Rohin Shah and John Youssef and Rishabh Agarwal and Natalie Dabney and Alessio Tonioni and Moran Ambar and Jing Li and Isabelle Guyon and Benny Li and David Soergel and Boya Fang and Georgi Karadzhov and Cristian Udrescu and Trieu Trinh and Vikas Raunak and Seb Noury and Dee Guo and Sonal Gupta and Mara Finkelstein and Denis Petek and Lihao Liang and Greg Billock and Pei Sun and David Wood and Yiwen Song and Xiaobin Yu and Tatiana Matejovicova and Regev Cohen and Kalyan Andra and David D'Ambrosio and Zhiwei Deng and Vincent Nallatamby and Ebrahim Songhori and Rumen Dangovski and Andrew Lampinen and Pankil Botadra and Adam Hillier and Jiawei Cao and Nagabhushan Baddi and Adhi Kuncoro and Toshihiro Yoshino and Ankit Bhagatwala and Marcáurelio Ranzato and Rylan Schaeffer and Tianlin Liu and Shuai Ye and Obaid Sarvana and John Nham and Chenkai Kuang and Isabel Gao and Jinoo Baek and Shubham Mittal and Ayzaan Wahid and Anita Gergely and Bin Ni and Josh Feldman and Carrie Muir and Pascal Lamblin and Wolfgang Macherey and Ethan Dyer and Logan Kilpatrick and Víctor Campos and Mukul Bhutani and Stanislav Fort and Yanif Ahmad and Aliaksei Severyn and Kleopatra Chatziprimou and Oleksandr Ferludin and Mason Dimarco and Aditya Kusupati and Joe Heyward and Dan Bahir and Kevin Villela and Katie Millican and Dror Marcus and Sanaz Bahargam and Caglar Unlu and Nicholas Roth and Zichuan Wei and Siddharth Gopal and Deepanway Ghoshal and Edward Lee and Sharon Lin and Jennie Lees and Dayeong Lee and Anahita Hosseini and Connie Fan and Seth Neel and Marcus Wu and Yasemin Altun and Honglong Cai and Enrique Piqueras and Josh Woodward and Alessandro Bissacco and Salem Haykal and Mahyar Bordbar and Prasha Sundaram and Sarah Hodkinson and Daniel Toyama and George Polovets and Austin Myers and Anu Sinha and Tomer Levinboim and Kashyap Krishnakumar and Rachita Chhaparia and Tatiana Sholokhova and Nitesh Bharadwaj Gundavarapu and Ganesh Jawahar and Haroon Qureshi and Jieru Hu and Nikola Momchev and Matthew Rahtz and Renjie Wu and Aishwarya P S and Kedar Dhamdhere and Meiqi Guo and Umang Gupta and Ali Eslami and Mariano Schain and Michiel Blokzijl and David Welling and Dave Orr and Levent Bolelli and Nicolas Perez-Nieves and Mikhail Sirotenko and Aman Prasad and Arjun Kar and Borja De Balle Pigem and Tayfun Terzi and Gellért Weisz and Dipankar Ghosh and Aditi Mavalankar and Dhruv Madeka and Kaspar Daugaard and Hartwig Adam and Viraj Shah and Dana Berman and Maggie Tran and Steven Baker and Ewa Andrejczuk and Grishma Chole and Ganna Raboshchuk and Mahdi Mirzazadeh and Thais Kagohara and Shimu Wu and Christian Schallhart and Bernett Orlando and Chen Wang and Alban Rrustemi and Hao Xiong and Hao Liu and Arpi Vezer and Nolan Ramsden and Shuo-yiin Chang and Sidharth Mudgal and Yan Li and Nino Vieillard and Yedid Hoshen and Farooq Ahmad and Ambrose Slone and Amy Hua and Natan Potikha and Mirko Rossini and Jon Stritar and Sushant Prakash and Zifeng Wang and Xuanyi Dong and Alireza Nazari and Efrat Nehoran and Kaan Tekelioglu and Yinxiao Li and Kartikeya Badola and Tom Funkhouser and Yuanzhen Li and Varun Yerram and Ramya Ganeshan and Daniel Formoso and Karol Langner and Tian Shi and Huijian Li and Yumeya Yamamori and Amayika Panda and Alaa Saade and Angelo Scorza Scarpati and Chris Breaux and CJ Carey and Zongwei Zhou and Cho-Jui Hsieh and Sophie Bridgers and Alena Butryna and Nishesh Gupta and Vaibhav Tulsyan and Sanghyun Woo and Evgenii Eltyshev and Will Grathwohl and Chanel Parks and Seth Benjamin and Rina Panigrahy and Shenil Dodhia and Daniel De Freitas and Chris Sauer and Will Song and Ferran Alet and Jackson Tolins and Cosmin Paduraru and Xingyi Zhou and Brian Albert and Zizhao Zhang and Lei Shu and Mudit Bansal and Sarah Nguyen and Amir Globerson and Owen Xiao and James Manyika and Tom Hennigan and Rong Rong and Josip Matak and Anton Bakalov and Ankur Sharma and Danila Sinopalnikov and Andrew Pierson and Stephen Roller and Geoff Brown and Mingcen Gao and Toshiyuki Fukuzawa and Amin Ghafouri and Kenny Vassigh and Iain Barr and Zhicheng Wang and Anna Korsun and Rajesh Jayaram and Lijie Ren and Tim Zaman and Samira Khan and Yana Lunts and Dan Deutsch and Dave Uthus and Nitzan Katz and Masha Samsikova and Amr Khalifa and Nikhil Sethi and Jiao Sun and Luming Tang and Uri Alon and Xianghong Luo and Dian Yu and Abhishek Nayyar and Bryce Petrini and Will Truong and Vincent Hellendoorn and Nikolai Chinaev and Chris Alberti and Wei Wang and Jingcao Hu and Vahab Mirrokni and Ananth Balashankar and Avia Aharon and Aahil Mehta and Ahmet Iscen and Joseph Kready and Lucas Manning and Anhad Mohananey and Yuankai Chen and Anshuman Tripathi and Allen Wu and Igor Petrovski and Dawsen Hwang and Martin Baeuml and Shreyas Chandrakaladharan and Yuan Liu and Rey Coaguila and Maxwell Chen and Sally Ma and Pouya Tafti and Susheel Tatineni and Terry Spitz and Jiayu Ye and Paul Vicol and Mihaela Rosca and Adrià Puigdomènech and Zohar Yahav and Sanjay Ghemawat and Hanzhao Lin and Phoebe Kirk and Zaid Nabulsi and Sergey Brin and Bernd Bohnet and Ken Caluwaerts and Aditya Srikanth Veerubhotla and Dan Zheng and Zihang Dai and Petre Petrov and Yichong Xu and Ramin Mehran and Zhuo Xu and Luisa Zintgraf and Jiho Choi and Spurthi Amba Hombaiah and Romal Thoppilan and Sashank Reddi and Lukasz Lew and Li Li and Kellie Webster and KP Sawhney and Lampros Lamprou and Siamak Shakeri and Mayank Lunayach and Jianmin Chen and Sumit Bagri and Alex Salcianu and Ying Chen and Yani Donchev and Charlotte Magister and Signe Nørly and Vitor Rodrigues and Tomas Izo and Hila Noga and Joe Zou and Thomas Köppe and Wenxuan Zhou and Kenton Lee and Xiangzhu Long and Danielle Eisenbud and Anthony Chen and Connor Schenck and Chi Ming To and Peilin Zhong and Emanuel Taropa and Minh Truong and Omer Levy and Danilo Martins and Zhiyuan Zhang and Christopher Semturs and Kelvin Zhang and Alex Yakubovich and Pol Moreno and Lara McConnaughey and Di Lu and Sam Redmond and Lotte Weerts and Yonatan Bitton and Tiziana Refice and Nicolas Lacasse and Arthur Conmy and Corentin Tallec and Julian Odell and Hannah Forbes-Pollard and Arkadiusz Socala and Jonathan Hoech and Pushmeet Kohli and Alanna Walton and Rui Wang and Mikita Sazanovich and Kexin Zhu and Andrei Kapishnikov and Rich Galt and Matthew Denton and Ben Murdoch and Caitlin Sikora and Kareem Mohamed and Wei Wei and Uri First and Tim McConnell and Luis C. Cobo and James Qin and Thi Avrahami and Daniel Balle and Yu Watanabe and Annie Louis and Adam Kraft and Setareh Ariafar and Yiming Gu and Eugénie Rives and Charles Yoon and Andrei Rusu and James Cobon-Kerr and Chris Hahn and Jiaming Luo and Yuvein and Zhu and Niharika Ahuja and Rodrigo Benenson and Raphaël Lopez Kaufman and Honglin Yu and Lloyd Hightower and Junlin Zhang and Darren Ni and Lisa Anne Hendricks and Gabby Wang and Gal Yona and Lalit Jain and Pablo Barrio and Surya Bhupatiraju and Siva Velusamy and Allan Dafoe and Sebastian Riedel and Tara Thomas and Zhe Yuan and Mathias Bellaiche and Sheena Panthaplackel and Klemen Kloboves and Sarthak Jauhari and Canfer Akbulut and Todor Davchev and Evgeny Gladchenko and David Madras and Aleksandr Chuklin and Tyrone Hill and Quan Yuan and Mukundan Madhavan and Luke Leonhard and Dylan Scandinaro and Qihang Chen and Ning Niu and Arthur Douillard and Bogdan Damoc and Yasumasa Onoe and Fabian Pedregosa and Fred Bertsch and Chas Leichner and Joseph Pagadora and Jonathan Malmaud and Sameera Ponda and Andy Twigg and Oleksii Duzhyi and Jingwei Shen and Miaosen Wang and Roopal Garg and Jing Chen and Utku Evci and Jonathan Lee and Leon Liu and Koji Kojima and Masa Yamaguchi and Arunkumar Rajendran and AJ Piergiovanni and Vinodh Kumar Rajendran and Marco Fornoni and Gabriel Ibagon and Harry Ragan and Sadh MNM Khan and John Blitzer and Andrew Bunner and Guan Sun and Takahiro Kosakai and Scott Lundberg and Ndidi Elue and Kelvin Guu and SK Park and Jane Park and Arunachalam Narayanaswamy and Chengda Wu and Jayaram Mudigonda and Trevor Cohn and Hairong Mu and Ravi Kumar and Laura Graesser and Yichi Zhang and Richard Killam and Vincent Zhuang and Mai Giménez and Wael Al Jishi and Ruy Ley-Wild and Alex Zhai and Kazuki Osawa and Diego Cedillo and Jialu Liu and Mayank Upadhyay and Marcin Sieniek and Roshan Sharma and Tom Paine and Anelia Angelova and Sravanti Addepalli and Carolina Parada and Kingshuk Majumder and Avery Lamp and Sanjiv Kumar and Xiang Deng and Artiom Myaskovsky and Tea Sabolić and Jeffrey Dudek and Sarah York and Félix de Chaumont Quitry and Jiazhong Nie and Dee Cattle and Alok Gunjan and Bilal Piot and Waleed Khawaja and Seojin Bang and Simon Wang and Siavash Khodadadeh and Raghavender R and Praynaa Rawlani and Richard Powell and Kevin Lee and Johannes Griesser and GS Oh and Cesar Magalhaes and Yujia Li and Simon Tokumine and Hadas Natalie Vogel and Dennis Hsu and Arturo BC and Disha Jindal and Matan Cohen and Zi Yang and Junwei Yuan and Dario de Cesare and Tony Bruguier and Jun Xu and Monica Roy and Alon Jacovi and Dan Belov and Rahul Arya and Phoenix Meadowlark and Shlomi Cohen-Ganor and Wenting Ye and Patrick Morris-Suzuki and Praseem Banzal and Gan Song and Pranavaraj Ponnuramu and Fred Zhang and George Scrivener and Salah Zaiem and Alif Raditya Rochman and Kehang Han and Badih Ghazi and Kate Lee and Shahar Drath and Daniel Suo and Antonious Girgis and Pradeep Shenoy and Duy Nguyen and Douglas Eck and Somit Gupta and Le Yan and Joao Carreira and Anmol Gulati and Ruoxin Sang and Daniil Mirylenka and Emma Cooney and Edward Chou and Mingyang Ling and Cindy Fan and Ben Coleman and Guilherme Tubone and Ravin Kumar and Jason Baldridge and Felix Hernandez-Campos and Angeliki Lazaridou and James Besley and Itay Yona and Neslihan Bulut and Quentin Wellens and AJ Pierigiovanni and Jasmine George and Richard Green and Pu Han and Connie Tao and Geoff Clark and Chong You and Abbas Abdolmaleki and Justin Fu and Tongzhou Chen and Ashwin Chaugule and Angad Chandorkar and Altaf Rahman and Will Thompson and Penporn Koanantakool and Mike Bernico and Jie Ren and Andrey Vlasov and Sergei Vassilvitskii and Maciej Kula and Yizhong Liang and Dahun Kim and Yangsibo Huang and Chengxi Ye and Dmitry Lepikhin and Wesley Helmholz},
      year={2025},
      eprint={2507.06261},
      archivePrefix={arXiv},
      primaryClass={cs.CL},
      url={https://arxiv.org/abs/2507.06261}, 
}

@misc{openai2024gpt4technicalreport,
      title={GPT-4 Technical Report}, 
      author={OpenAI and Josh Achiam and Steven Adler and Sandhini Agarwal and Lama Ahmad and Ilge Akkaya and Florencia Leoni Aleman and Diogo Almeida and Janko Altenschmidt and Sam Altman and Shyamal Anadkat and Red Avila and Igor Babuschkin and Suchir Balaji and Valerie Balcom and Paul Baltescu and Haiming Bao and Mohammad Bavarian and Jeff Belgum and Irwan Bello and Jake Berdine and Gabriel Bernadett-Shapiro and Christopher Berner and Lenny Bogdonoff and Oleg Boiko and Madelaine Boyd and Anna-Luisa Brakman and Greg Brockman and Tim Brooks and Miles Brundage and Kevin Button and Trevor Cai and Rosie Campbell and Andrew Cann and Brittany Carey and Chelsea Carlson and Rory Carmichael and Brooke Chan and Che Chang and Fotis Chantzis and Derek Chen and Sully Chen and Ruby Chen and Jason Chen and Mark Chen and Ben Chess and Chester Cho and Casey Chu and Hyung Won Chung and Dave Cummings and Jeremiah Currier and Yunxing Dai and Cory Decareaux and Thomas Degry and Noah Deutsch and Damien Deville and Arka Dhar and David Dohan and Steve Dowling and Sheila Dunning and Adrien Ecoffet and Atty Eleti and Tyna Eloundou and David Farhi and Liam Fedus and Niko Felix and Simón Posada Fishman and Juston Forte and Isabella Fulford and Leo Gao and Elie Georges and Christian Gibson and Vik Goel and Tarun Gogineni and Gabriel Goh and Rapha Gontijo-Lopes and Jonathan Gordon and Morgan Grafstein and Scott Gray and Ryan Greene and Joshua Gross and Shixiang Shane Gu and Yufei Guo and Chris Hallacy and Jesse Han and Jeff Harris and Yuchen He and Mike Heaton and Johannes Heidecke and Chris Hesse and Alan Hickey and Wade Hickey and Peter Hoeschele and Brandon Houghton and Kenny Hsu and Shengli Hu and Xin Hu and Joost Huizinga and Shantanu Jain and Shawn Jain and Joanne Jang and Angela Jiang and Roger Jiang and Haozhun Jin and Denny Jin and Shino Jomoto and Billie Jonn and Heewoo Jun and Tomer Kaftan and Łukasz Kaiser and Ali Kamali and Ingmar Kanitscheider and Nitish Shirish Keskar and Tabarak Khan and Logan Kilpatrick and Jong Wook Kim and Christina Kim and Yongjik Kim and Jan Hendrik Kirchner and Jamie Kiros and Matt Knight and Daniel Kokotajlo and Łukasz Kondraciuk and Andrew Kondrich and Aris Konstantinidis and Kyle Kosic and Gretchen Krueger and Vishal Kuo and Michael Lampe and Ikai Lan and Teddy Lee and Jan Leike and Jade Leung and Daniel Levy and Chak Ming Li and Rachel Lim and Molly Lin and Stephanie Lin and Mateusz Litwin and Theresa Lopez and Ryan Lowe and Patricia Lue and Anna Makanju and Kim Malfacini and Sam Manning and Todor Markov and Yaniv Markovski and Bianca Martin and Katie Mayer and Andrew Mayne and Bob McGrew and Scott Mayer McKinney and Christine McLeavey and Paul McMillan and Jake McNeil and David Medina and Aalok Mehta and Jacob Menick and Luke Metz and Andrey Mishchenko and Pamela Mishkin and Vinnie Monaco and Evan Morikawa and Daniel Mossing and Tong Mu and Mira Murati and Oleg Murk and David Mély and Ashvin Nair and Reiichiro Nakano and Rajeev Nayak and Arvind Neelakantan and Richard Ngo and Hyeonwoo Noh and Long Ouyang and Cullen O'Keefe and Jakub Pachocki and Alex Paino and Joe Palermo and Ashley Pantuliano and Giambattista Parascandolo and Joel Parish and Emy Parparita and Alex Passos and Mikhail Pavlov and Andrew Peng and Adam Perelman and Filipe de Avila Belbute Peres and Michael Petrov and Henrique Ponde de Oliveira Pinto and Michael and Pokorny and Michelle Pokrass and Vitchyr H. Pong and Tolly Powell and Alethea Power and Boris Power and Elizabeth Proehl and Raul Puri and Alec Radford and Jack Rae and Aditya Ramesh and Cameron Raymond and Francis Real and Kendra Rimbach and Carl Ross and Bob Rotsted and Henri Roussez and Nick Ryder and Mario Saltarelli and Ted Sanders and Shibani Santurkar and Girish Sastry and Heather Schmidt and David Schnurr and John Schulman and Daniel Selsam and Kyla Sheppard and Toki Sherbakov and Jessica Shieh and Sarah Shoker and Pranav Shyam and Szymon Sidor and Eric Sigler and Maddie Simens and Jordan Sitkin and Katarina Slama and Ian Sohl and Benjamin Sokolowsky and Yang Song and Natalie Staudacher and Felipe Petroski Such and Natalie Summers and Ilya Sutskever and Jie Tang and Nikolas Tezak and Madeleine B. Thompson and Phil Tillet and Amin Tootoonchian and Elizabeth Tseng and Preston Tuggle and Nick Turley and Jerry Tworek and Juan Felipe Cerón Uribe and Andrea Vallone and Arun Vijayvergiya and Chelsea Voss and Carroll Wainwright and Justin Jay Wang and Alvin Wang and Ben Wang and Jonathan Ward and Jason Wei and CJ Weinmann and Akila Welihinda and Peter Welinder and Jiayi Weng and Lilian Weng and Matt Wiethoff and Dave Willner and Clemens Winter and Samuel Wolrich and Hannah Wong and Lauren Workman and Sherwin Wu and Jeff Wu and Michael Wu and Kai Xiao and Tao Xu and Sarah Yoo and Kevin Yu and Qiming Yuan and Wojciech Zaremba and Rowan Zellers and Chong Zhang and Marvin Zhang and Shengjia Zhao and Tianhao Zheng and Juntang Zhuang and William Zhuk and Barret Zoph},
      year={2024},
      eprint={2303.08774},
      archivePrefix={arXiv},
      primaryClass={cs.CL},
      url={https://arxiv.org/abs/2303.08774}, 
}

@misc{yang2025qwen3technicalreport,
      title={Qwen3 Technical Report}, 
      author={An Yang and Anfeng Li and Baosong Yang and Beichen Zhang and Binyuan Hui and Bo Zheng and Bowen Yu and Chang Gao and Chengen Huang and Chenxu Lv and Chujie Zheng and Dayiheng Liu and Fan Zhou and Fei Huang and Feng Hu and Hao Ge and Haoran Wei and Huan Lin and Jialong Tang and Jian Yang and Jianhong Tu and Jianwei Zhang and Jianxin Yang and Jiaxi Yang and Jing Zhou and Jingren Zhou and Junyang Lin and Kai Dang and Keqin Bao and Kexin Yang and Le Yu and Lianghao Deng and Mei Li and Mingfeng Xue and Mingze Li and Pei Zhang and Peng Wang and Qin Zhu and Rui Men and Ruize Gao and Shixuan Liu and Shuang Luo and Tianhao Li and Tianyi Tang and Wenbiao Yin and Xingzhang Ren and Xinyu Wang and Xinyu Zhang and Xuancheng Ren and Yang Fan and Yang Su and Yichang Zhang and Yinger Zhang and Yu Wan and Yuqiong Liu and Zekun Wang and Zeyu Cui and Zhenru Zhang and Zhipeng Zhou and Zihan Qiu},
      year={2025},
      eprint={2505.09388},
      archivePrefix={arXiv},
      primaryClass={cs.CL},
      url={https://arxiv.org/abs/2505.09388}, 
}

@misc{qwen2025qwen25technicalreport,
      title={Qwen2.5 Technical Report}, 
      author={Qwen and : and An Yang and Baosong Yang and Beichen Zhang and Binyuan Hui and Bo Zheng and Bowen Yu and Chengyuan Li and Dayiheng Liu and Fei Huang and Haoran Wei and Huan Lin and Jian Yang and Jianhong Tu and Jianwei Zhang and Jianxin Yang and Jiaxi Yang and Jingren Zhou and Junyang Lin and Kai Dang and Keming Lu and Keqin Bao and Kexin Yang and Le Yu and Mei Li and Mingfeng Xue and Pei Zhang and Qin Zhu and Rui Men and Runji Lin and Tianhao Li and Tianyi Tang and Tingyu Xia and Xingzhang Ren and Xuancheng Ren and Yang Fan and Yang Su and Yichang Zhang and Yu Wan and Yuqiong Liu and Zeyu Cui and Zhenru Zhang and Zihan Qiu},
      year={2025},
      eprint={2412.15115},
      archivePrefix={arXiv},
      primaryClass={cs.CL},
      url={https://arxiv.org/abs/2412.15115}, 
}

@misc{aryabumi2024aya23openweight,
      title={Aya 23: Open Weight Releases to Further Multilingual Progress}, 
      author={Viraat Aryabumi and John Dang and Dwarak Talupuru and Saurabh Dash and David Cairuz and Hangyu Lin and Bharat Venkitesh and Madeline Smith and Jon Ander Campos and Yi Chern Tan and Kelly Marchisio and Max Bartolo and Sebastian Ruder and Acyr Locatelli and Julia Kreutzer and Nick Frosst and Aidan Gomez and Phil Blunsom and Marzieh Fadaee and Ahmet Üstün and Sara Hooker},
      year={2024},
      eprint={2405.15032},
      archivePrefix={arXiv},
      primaryClass={cs.CL},
      url={https://arxiv.org/abs/2405.15032}, 
}

@misc{dang2024ayaexpansecombiningresearch,
      title={Aya Expanse: Combining Research Breakthroughs for a New Multilingual Frontier}, 
      author={John Dang and Shivalika Singh and Daniel D'souza and Arash Ahmadian and Alejandro Salamanca and Madeline Smith and Aidan Peppin and Sungjin Hong and Manoj Govindassamy and Terrence Zhao and Sandra Kublik and Meor Amer and Viraat Aryabumi and Jon Ander Campos and Yi-Chern Tan and Tom Kocmi and Florian Strub and Nathan Grinsztajn and Yannis Flet-Berliac and Acyr Locatelli and Hangyu Lin and Dwarak Talupuru and Bharat Venkitesh and David Cairuz and Bowen Yang and Tim Chung and Wei-Yin Ko and Sylvie Shang Shi and Amir Shukayev and Sammie Bae and Aleksandra Piktus and Roman Castagné and Felipe Cruz-Salinas and Eddie Kim and Lucas Crawhall-Stein and Adrien Morisot and Sudip Roy and Phil Blunsom and Ivan Zhang and Aidan Gomez and Nick Frosst and Marzieh Fadaee and Beyza Ermis and Ahmet Üstün and Sara Hooker},
      year={2024},
      eprint={2412.04261},
      archivePrefix={arXiv},
      primaryClass={cs.CL},
      url={https://arxiv.org/abs/2412.04261}, 
}

@inproceedings{wibowo-etal-2024-copal,
    title = "{COPAL}-{ID}: {I}ndonesian Language Reasoning with Local Culture and Nuances",
    author = "Wibowo, Haryo  and
      Fuadi, Erland  and
      Nityasya, Made  and
      Prasojo, Radityo Eko  and
      Aji, Alham",
    editor = "Duh, Kevin  and
      Gomez, Helena  and
      Bethard, Steven",
    booktitle = "Proceedings of the 2024 Conference of the North American Chapter of the Association for Computational Linguistics: Human Language Technologies (Volume 1: Long Papers)",
    month = jun,
    year = "2024",
    address = "Mexico City, Mexico",
    publisher = "Association for Computational Linguistics",
    url = "https://aclanthology.org/2024.naacl-long.77/",
    doi = "10.18653/v1/2024.naacl-long.77",
    pages = "1404--1422"
}

@misc{sakaguchi2019winograndeadversarialwinogradschema,
      title={WinoGrande: An Adversarial Winograd Schema Challenge at Scale}, 
      author={Keisuke Sakaguchi and Ronan Le Bras and Chandra Bhagavatula and Yejin Choi},
      year={2019},
      eprint={1907.10641},
      archivePrefix={arXiv},
      primaryClass={cs.CL},
      url={https://arxiv.org/abs/1907.10641}, 
}

@misc{clark2018arc,
      title={Think you have Solved Question Answering? Try ARC, the AI2 Reasoning Challenge}, 
      author={Peter Clark and Isaac Cowhey and Oren Etzioni and Tushar Khot and Ashish Sabharwal and Carissa Schoenick and Oyvind Tafjord},
      year={2018},
      eprint={1803.05457},
      archivePrefix={arXiv},
      primaryClass={cs.AI},
      url={https://arxiv.org/abs/1803.05457}, 
}

@misc{ismayilzada2023crow,
      title={CRoW: Benchmarking Commonsense Reasoning in Real-World Tasks}, 
      author={Mete Ismayilzada and Debjit Paul and Syrielle Montariol and Mor Geva and Antoine Bosselut},
      year={2023},
      eprint={2310.15239},
      archivePrefix={arXiv},
      primaryClass={cs.CL},
      url={https://arxiv.org/abs/2310.15239}, 
}

@misc{mmlu,
      title={Measuring Massive Multitask Language Understanding}, 
      author={Dan Hendrycks and Collin Burns and Steven Basart and Andy Zou and Mantas Mazeika and Dawn Song and Jacob Steinhardt},
      year={2021},
      eprint={2009.03300},
      archivePrefix={arXiv},
      primaryClass={cs.CY},
      url={https://arxiv.org/abs/2009.03300}, 
}

@misc{mmlupro,
      title={MMLU-Pro: A More Robust and Challenging Multi-Task Language Understanding Benchmark}, 
      author={Yubo Wang and Xueguang Ma and Ge Zhang and Yuansheng Ni and Abhranil Chandra and Shiguang Guo and Weiming Ren and Aaran Arulraj and Xuan He and Ziyan Jiang and Tianle Li and Max Ku and Kai Wang and Alex Zhuang and Rongqi Fan and Xiang Yue and Wenhu Chen},
      year={2024},
      eprint={2406.01574},
      archivePrefix={arXiv},
      primaryClass={cs.CL},
      url={https://arxiv.org/abs/2406.01574}, 
}

@misc{fabbri2025multinrcchallengingnativemultilingual,
      title={MultiNRC: A Challenging and Native Multilingual Reasoning Evaluation Benchmark for LLMs}, 
      author={Alexander R. Fabbri and Diego Mares and Jorge Flores and Meher Mankikar and Ernesto Hernandez and Dean Lee and Bing Liu and Chen Xing},
      year={2025},
      eprint={2507.17476},
      archivePrefix={arXiv},
      primaryClass={cs.CL},
      url={https://arxiv.org/abs/2507.17476}, 
}

@inproceedings{myung2024blend,
  title     = {BLEnD: A Benchmark for LLMs on Everyday Knowledge in Diverse Cultures and Languages},
  author    = {Myung, Junho and others},
  booktitle = {Advances in Neural Information Processing Systems (Datasets and Benchmarks Track)},
  year      = {2024}
}

\appendix

\section{Data Statistics in More Detail}
\label{more-data}

This section provides additional breakdowns of template coverage across cultural aspects and reasoning categories, supplementing Table~\ref{tab:data_stats} in the main paper.

\subsection{Cultural-aspect coverage}
\begin{figure}[H]
  \centering
  \includegraphics[width=\columnwidth]{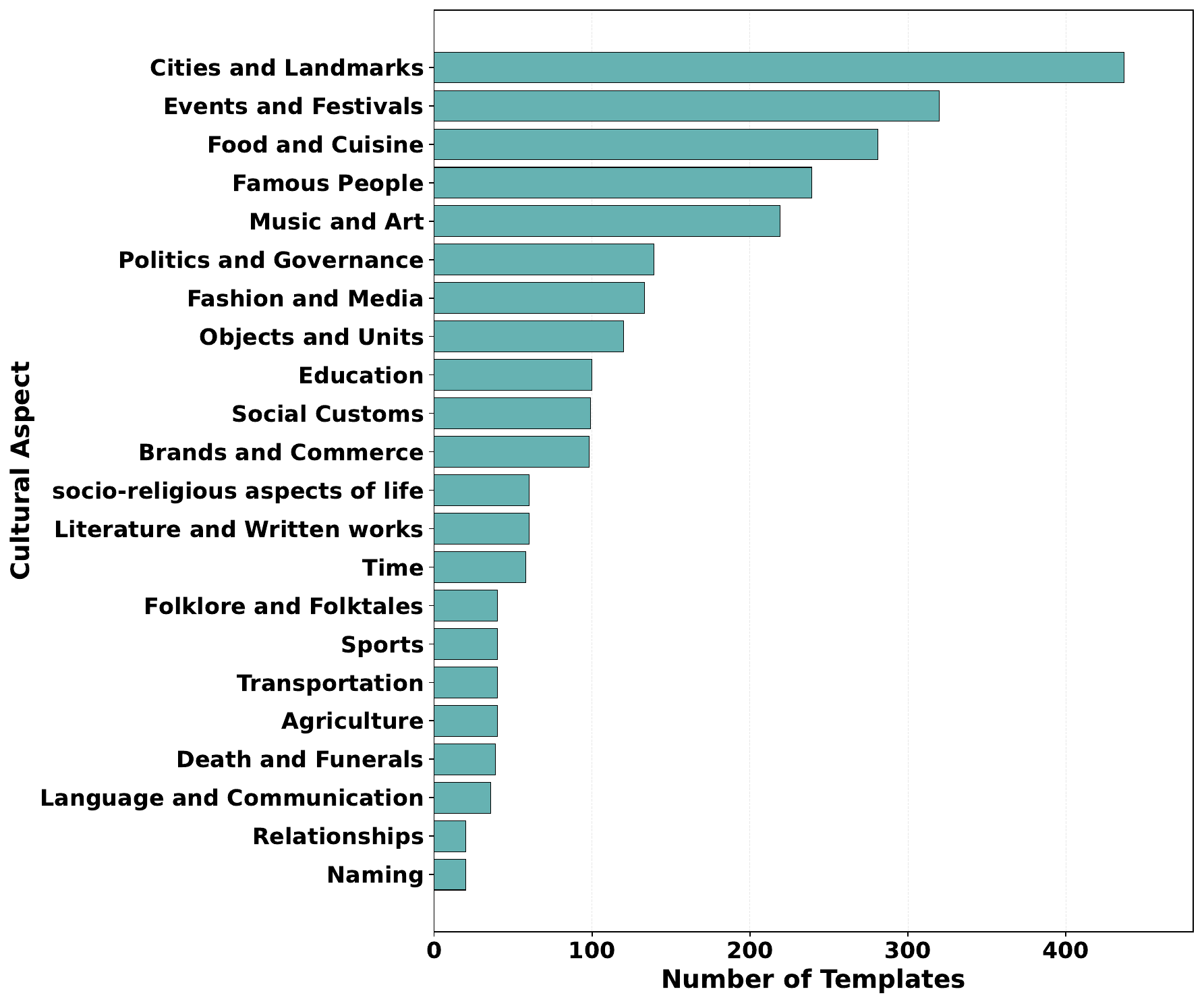}
  \caption{\textbf{Distribution of cultural-aspect tags in the benchmark.}
  Bars report the number of \emph{template instantiations} tagged with each of the 22 cultural aspects.
  Aspects are not mutually exclusive, so a single item may contribute to multiple bars.}
  \label{fig:culture}
\end{figure}

\subsection{Reasoning-category coverage}
\begin{figure}[H]
  \centering
  \includegraphics[width=\columnwidth]{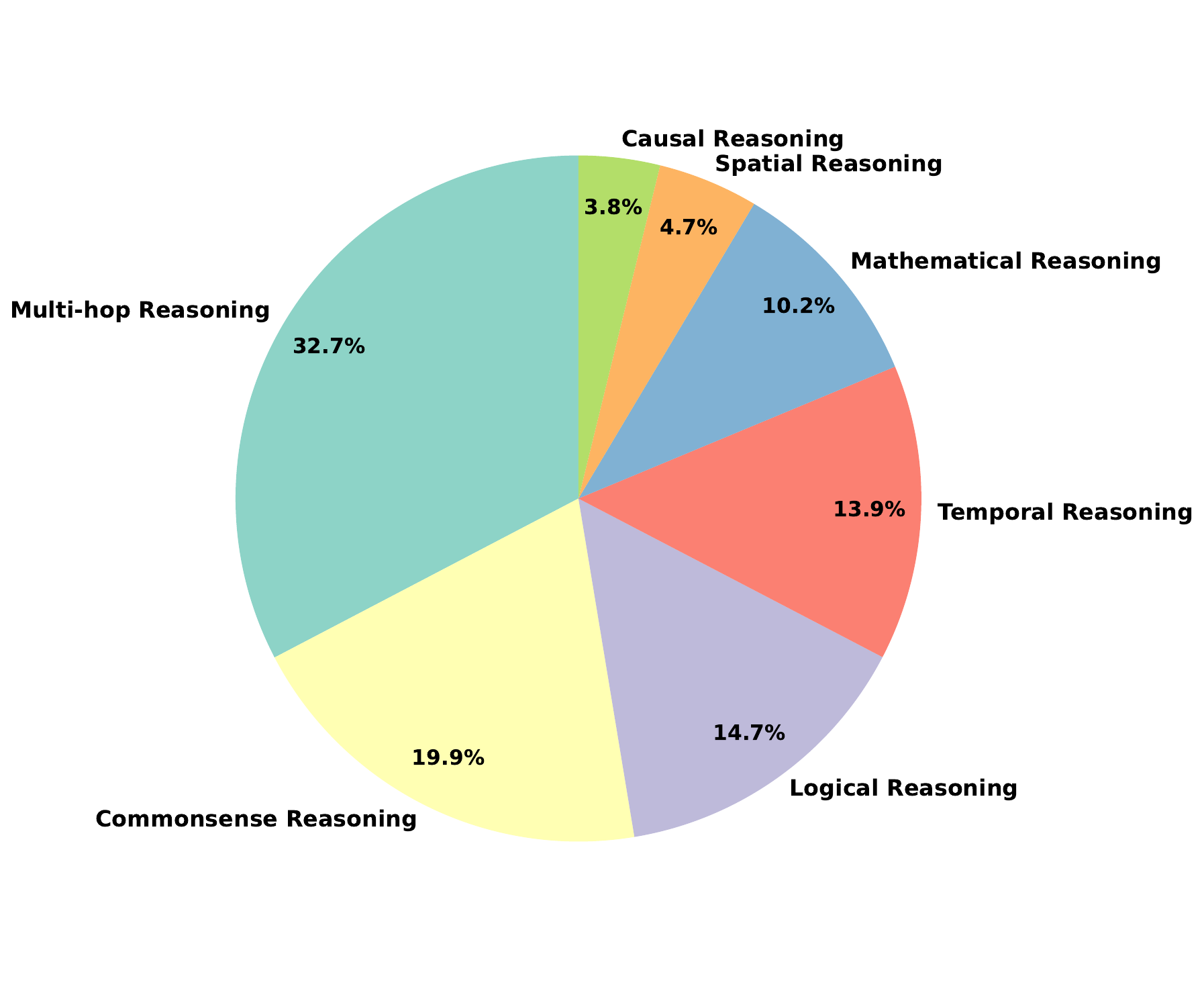}
  \caption{\textbf{Distribution of reasoning-category tags in the template set.}
  Percentages are normalized over tag assignments (multi-label items contribute multiple assignments).}
  \label{fig:reasoning}
\end{figure}


\section{Verifier Agreement}
\label{app:iaa}

To verify the factual correctness of cultural content and the clarity of local-language translations, we recruited three independent native verifiers per cultural context, distinct from the original annotators.
Each verifier independently answered the MCQ version of every item in the local language, without access to the original annotations.
Inter-annotator agreement (IAA) is defined as the proportion of items on which all three verifiers selected the same answer.
When verifiers disagreed, we applied majority voting to determine the accepted answer and flagged the item for further review; items identified by a majority as factually incorrect or ambiguous were revised accordingly.

Table~\ref{tab:iaa} reports per-context IAA scores.
Scores range from 87.0\% (South Africa) to 94.8\% (China), with a mean of 90.9\%, confirming high factual accuracy and translation clarity across all 20 cultural contexts.

\begin{table}[H]
\centering
\small
\setlength{\tabcolsep}{5pt}
\renewcommand{\arraystretch}{1.05}
\begin{tabular}{lc|lc}
\toprule
\textbf{Country} & \textbf{IAA (\%)} & \textbf{Country} & \textbf{IAA (\%)} \\
\midrule
Brazil       & 92.00 & Japan        & 91.80 \\
China        & 94.80 & Kyrgyzstan   & 92.50 \\
Egypt        & 91.30 & Mexico       & 93.43 \\
Ethiopia     & 89.10 & Morocco      & 90.46 \\
Georgia      & 92.30 & Nigeria      & 92.80 \\
Greece       & 90.21 & Philippines  & 88.00 \\
India        & 91.30 & South Africa & 87.00 \\
Indonesia    & 90.00 & Thailand     & 88.60 \\
Italy        & 88.97 & Tunisia      & 89.70 \\
Turkey       & 90.11 & Yemen        & 89.80 \\
\bottomrule
\end{tabular}
\caption{Inter-annotator agreement (IAA) across all 20 cultural contexts, defined as the proportion of items on which all three independent native verifiers selected the same answer.}
\label{tab:iaa}
\end{table}

\section{Annotator Demographics}
\label{app:annotators}

Table~\ref{tab:annotator_demographics} reports demographic information for the two annotators recruited per cultural context via Upwork, including gender, age group, duration of residence in the target culture, and education level.
Annotators were required to be native speakers of the target language with substantial lived experience in the corresponding cultural context.

\begin{table*}[t]
\centering
\scriptsize
\setlength{\tabcolsep}{3pt}
\renewcommand{\arraystretch}{1.05}

\resizebox{\textwidth}{!}{
\begin{tabular}{lcccccccccccccccccccc}
\toprule
 & EG & PH & IN & ET & MX & TN & GR & BR & KG & ZA & IT & TH & TR & GE & CN & ID & YE & NG & MA & JP \\
\midrule
\textbf{No. of annotators} & 2 & 2 & 2 & 2 & 2 & 2 & 2 & 2 & 2 & 2 & 2 & 2 & 2 & 2 & 2 & 2 & 2 & 2 & 2 & 2 \\
\midrule

\multicolumn{21}{l}{\textbf{Gender (\%)}} \\
Female & 100 & 100 & 50 & 0 & 100 & 100 & 50 & 50 & 50 & 50 & 50 & 100 & 50 & 100 & 50 & 0 & 50 & 50 & 50 & 50 \\
Male   & 0   & 0   & 50 & 100 & 0   & 0   & 50 & 50 & 50 & 50 & 50 & 0   & 50 & 0   & 50 & 100 & 50 & 50 & 50 & 50 \\
\midrule

\multicolumn{21}{l}{\textbf{Age (\%)}} \\
20--24 & 50 & 50 & 0  & 50 & 0  & 0   & 0   & 0   & 0   & 0   & 0   & 0  & 0   & 0   & 0   & 0   & 50 & 0   & 50 & 0 \\
25--35 & 50 & 50 & 50 & 50 & 50 & 100 & 0   & 100 & 100 & 50  & 50  & 50 & 100 & 0   & 100 & 100 & 50 & 100 & 0  & 100 \\
36--45 & 0  & 0  & 50 & 0  & 0  & 0   & 100 & 0   & 0   & 50  & 0   & 50 & 0   & 100 & 0   & 0   & 0  & 0   & 50 & 0 \\
46+    & 0  & 0  & 0  & 0  & 50 & 0   & 0   & 0   & 0   & 0   & 50  & 0  & 0   & 0   & 0   & 0   & 0  & 0   & 0  & 0 \\
\midrule

\multicolumn{21}{l}{\textbf{Duration of residence (years) (\%)}} \\
Native / whole life & 0   & 100 & 100 & 100 & 100 & 100 & 100 & 100 & 100 & 100 & 50 & 100 & 50 & 100 & 0   & 0   & 0   & 100 & 100 & 100 \\
$\ge$20 years       & 100 & 0   & 0   & 0   & 0   & 0   & 0   & 0   & 0   & 0   & 50 & 0   & 0  & 0   & 100 & 100 & 100 & 0   & 0   & 0 \\
10--19 years        & 0   & 0   & 0   & 0   & 0   & 0   & 0   & 0   & 0   & 0   & 0  & 0   & 50 & 0   & 0   & 0   & 0   & 0   & 0   & 0 \\
$<$10 years         & 0   & 0   & 0   & 0   & 0   & 0   & 0   & 0   & 0   & 0   & 0  & 0   & 0  & 0   & 0   & 0   & 0   & 0   & 0   & 0 \\
\midrule

\multicolumn{21}{l}{\textbf{Education level (\%)}} \\
High school or below & 0 & 0 & 0 & 0 & 0 & 0 & 50 & 0 & 0 & 50 & 0 & 0 & 0 & 0 & 0 & 0 & 0 & 0 & 0 & 0 \\
Diploma / college    & 0 & 0 & 0 & 0 & 0 & 0 & 0  & 0 & 0 & 50 & 0 & 0 & 0 & 0 & 0 & 0 & 0 & 0 & 0 & 0 \\
Bachelor's           & 0 & 100 & 100 & 100 & 50 & 0 & 0 & 50 & 100 & 0 & 0 & 0 & 100 & 50 & 0 & 0 & 100 & 50 & 50 & 50 \\
Master's / postgrad  & 100 & 0 & 0 & 0 & 50 & 100 & 50 & 50 & 0 & 0 & 100 & 100 & 0 & 50 & 0 & 0 & 0 & 50 & 50 & 50 \\
Doctorate            & 0 & 0 & 0 & 0 & 0 & 0 & 0  & 0  & 0  & 0  & 0  & 0  & 0  & 0  & 100 & 100 & 0  & 0  & 0  & 0 \\
\bottomrule
\end{tabular}
}

\caption{\textbf{Annotator demographics by cultural context (2 annotators per context).}
Percentages are multiples of 50\% due to $n=2$.
Abbreviations: EG=Egypt, PH=Philippines, IN=India, ET=Ethiopia, MX=Mexico, TN=Tunisia, GR=Greece, BR=Brazil, KG=Kyrgyzstan, ZA=South Africa, IT=Italy, TH=Thailand, TR=Turkey, GE=Georgia, CN=China, ID=Indonesia, YE=Yemen, NG=Nigeria, MA=Morocco, JP=Japan.}
\label{tab:annotator_demographics}
\end{table*}


\section{Log-Probability vs.\ Generation Scoring}
\label{app:logprob-vs-gen}

For open-weight models we primarily report log-probability scoring, which selects the answer option with the highest token-level log-likelihood.
To confirm this choice does not introduce systematic bias, we re-evaluated all open-weight models using direct generation and compared results.
Table~\ref{tab:logprob_vs_gen} reports both protocols side by side.
Differences are small and go in both directions, confirming the absence of systematic bias.
The most notable exception is \texttt{meta-llama-llama-3.1-8b-instruct}, which drops substantially under generation in MC-EN (54.19\,$\to$\,38.75) --- a result consistent with smaller models being more prone to output-format failures rather than knowledge failures.
This validates our choice to use log-probability scoring for open-weight models.

\begin{table}[H]
\centering
\scriptsize
\setlength{\tabcolsep}{3pt}
\renewcommand{\arraystretch}{1.05}
\begin{tabular}{@{}l@{ ~ }l@{ ~ }rrrr@{}}
\toprule
\textbf{Model} & \textbf{Mode} & \textbf{MC-EN} & \textbf{MC-L} & \textbf{TF-EN} & \textbf{TF-L} \\
\midrule
\multirow{2}{*}{coherelabs-aya-23-8b}
 & logprob   & 43.79 & 39.15 & 50.35 & 50.45 \\
 & gen & 40.00 & 36.00 & 50.00 & 50.00 \\
\midrule
\multirow{2}{*}{coherelabs-aya-expanse-8b}
 & logprob   & 52.67 & 48.74 & 51.74 & 52.47 \\
 & gen & 54.02 & 47.75 & 56.73 & 54.91 \\
\midrule
\multirow{2}{*}{meta-llama-llama-3.1-8b-instruct}
 & logprob   & 54.19 & 43.39 & 56.66 & 53.36 \\
 & gen & 38.75 & 46.25 & 53.12 & 51.88 \\
\midrule
\multirow{2}{*}{meta-llama-llama-3.2-3b-instruct}
 & logprob   & 47.38 & 36.33 & 54.87 & 51.64 \\
 & gen & 44.86 & 33.87 & 47.75 & 49.64 \\
\midrule
\multirow{2}{*}{qwen-qwen2.5-7b-instruct}
 & logprob   & 56.96 & 46.92 & 52.65 & 52.62 \\
 & gen & 55.54 & 48.15 & 58.27 & 55.49 \\
\midrule
\multirow{2}{*}{qwen-qwen3-4b-instruct-2507}
 & logprob   & 52.62 & 45.56 & 55.25 & 53.68 \\
 & gen & 53.16 & 46.13 & 55.16 & 53.95 \\
\bottomrule
\end{tabular}
\caption{Log-probability vs.\ generation scoring for open-weight models (accuracy, \%). Differences are small and unsystematic, supporting log-probability scoring as the primary protocol.}
\label{tab:logprob_vs_gen}
\end{table}

\section{Answer Extraction Error Rates}
\label{app:extraction-errors}

For generation-based models, Table~\ref{tab:extraction_errors} reports three error types: (i) \textbf{Empty Output} --- the model produced no output; (ii) \textbf{Answer Not Extracted} --- the model produced output but no valid answer letter could be parsed; (iii) \textbf{Correct but Mis-scored} --- the model's response contained the correct answer but was marked wrong due to an extraction failure.
Error rates are near zero across all models, confirming that extraction failures do not meaningfully affect benchmark scores or conclusions.

\begin{table}[H]
\centering
\scriptsize
\setlength{\tabcolsep}{3pt}
\renewcommand{\arraystretch}{1.05}
\begin{tabular}{@{}l@{}rrr@{}}
\toprule
\textbf{Model} & \textbf{Empty} & \textbf{Not Extracted} & \textbf{Correct but} \\
               & \textbf{Output} &  & \textbf{Mis-scored} \\
\midrule
meta-llama-llama-3.2-3b-instruct  & 0.00 & 3.56 & 0.37 \\
meta-llama-llama-3.1-8b-instruct  & 0.00 & 0.14 & 1.27 \\
coherelabs-aya-23-8b              & 0.07 & 0.35 & 0.01 \\
anthropic-claude-haiku-4.5        & 0.00 & 0.14 & 0.00 \\
deepseek-deepseek-chat-v3.1       & 0.00 & 0.11 & 0.00 \\
meta-llama-llama-3.3-70b-instruct & 0.00 & 0.07 & 0.00 \\
openai-gpt-5-chat                 & 0.00 & 0.04 & 0.00 \\
google-gemini-3-flash-preview     & 0.00 & 0.03 & 0.00 \\
anthropic-claude-opus-4.5         & 0.00 & 0.02 & 0.00 \\
qwen-qwen2.5-7b-instruct          & 0.00 & 0.02 & 0.00 \\
meta-llama-llama-4-maverick       & 0.00 & 0.02 & 0.00 \\
qwen-qwen3-4b-instruct-2507       & 0.00 & 0.01 & 0.00 \\
google-gemini-2.5-flash           & 0.00 & 0.01 & 0.00 \\
coherelabs-aya-expanse-8b         & 0.00 & 0.00 & 0.00 \\
qwen-qwen3-235b-a22b-2507         & 0.00 & 0.00 & 0.00 \\
openai-gpt-4o-mini                & 0.00 & 0.00 & 0.00 \\
\bottomrule
\end{tabular}
\caption{Answer extraction error rates (\%) for generation-based models. The ``Correct but Mis-scored'' rate reaches at most 1.27\% (Llama 3.1 8B), confirming extraction failures do not significantly affect conclusions.}
\label{tab:extraction_errors}
\end{table}

\section{Evaluation Prompts}
\label{app:eval-prompts}

We evaluate each model under two prompting regimes. For \textbf{standard (non-thinking)} models, we request a direct answer without eliciting explanations. For \textbf{thinking-capable} models, we allow step-by-step reasoning but enforce a strict final-line JSON output format for automatic parsing.

\subsection{Non-thinking models: Multiple-choice prompt}
\begin{promptbox}[title={Non-Thinking Models: MCQ Prompt}]
Question: {question}

A) {option1}
B) {option2}
C) {option3}
D) {option4}

Answer directly without any explanation or reasoning.
Output a single line containing only a JSON object of the form:
{"answer": "X"}

Replace X with the correct letter (A, B, C, or D).
Do not include anything else.
\end{promptbox}

\subsection{Non-thinking models: True/False prompt}
\begin{promptbox}[title={Non-Thinking Models: True/False Prompt}]
Statement: {question}

Answer directly without any explanation or reasoning.
Output a single line containing only a JSON object of the form:
{"answer": "ANSWER"}

Replace ANSWER with either "T" for True or "F" for False.
Do not include anything else.
\end{promptbox}

\subsection{Thinking models: Multiple-choice prompt}
\begin{promptbox}[title={Thinking Models: MCQ Prompt}]
Question: {question}

A) {option1}
B) {option2}
C) {option3}
D) {option4}

First, you may think step by step to solve the question.
At the very end, output a single line containing only a JSON object of the form:
{"answer": "X"}

Replace X with the correct letter (A, B, C, or D).
Do not include anything else on that final line.
\end{promptbox}

\subsection{Thinking models: True/False prompt}
\begin{promptbox}[title={Thinking Models: True/False Prompt}]
Statement: {question}

First, you may think step by step.
At the very end, output a single line containing only a JSON object of the form:
{"answer": "ANSWER"}

Replace ANSWER with either "T" for True or "F" for False.
Do not include anything else on that final line.
\end{promptbox}

\subsection{Data verification prompt (GPT-5.2-chat)}
To improve linguistic quality while preserving meaning, we run an automated proofreading pass using GPT-5.2-chat that is restricted to spelling and grammar fixes only:
\begin{promptbox}[title={Data Verification Prompt (GPT-5.2-chat)}]
You are a careful proofreader. Your task is to check this {context} for ONLY grammatical and spelling mistakes.

**Instructions:**
1. Fix ONLY spelling errors and grammatical mistakes
2. Preserve the exact writing style, tone, and word choices
3. Keep all cultural references and proper nouns exactly as they are
4. Do NOT rephrase or improve the writing - only correct errors
5. If there are no errors, return the text exactly as-is

**Common errors to check for:**
- **Spelling mistakes**: "recieve" -> "receive", "occured" -> "occurred"
- **Plural/singular agreement**: "1 foods" -> "1 food", "2 dish" -> "2 dishes"
- **Subject-verb agreement**: "they was" -> "they were", "he don't" -> "he doesn't", "that are" with singular -> "that is"
- **Verb tense for past events**: "Is X established in 1950?" -> "Was X established in 1950?" (BE CONSISTENT across similar questions)
- **Pronoun agreement**: Singular "their" is acceptable as gender-neutral. Keep it unless clearly wrong. Be CONSISTENT within the same question set.
- **Unnecessary apostrophes**: "apple's" (plural) -> "apples"
- **Capitalization**: Only fix clear errors like "i" -> "I" or well-known proper nouns (countries, famous people). DO NOT automatically capitalize cultural terms, food names, or words that might be intentionally lowercase.
- **Article errors**: "a apple" -> "an apple", "an hour" not "a hour"

**Text to check:**
{text}

**Response format:**
If there are errors:
{
  "has_errors": true,
  "corrected": "the corrected text here",
  "explanation": "brief explanation of what was fixed (e.g., 'Fixed plural/singular agreement')"
}

If no errors:
{
  "has_errors": false,
  "corrected": "{text}",
  "explanation": "No errors found"
}

Respond ONLY with valid JSON, nothing else.
\end{promptbox}

\section{Example Templates by Cultural Aspect}
\label{app:templates}

Table~\ref{tab:cultural_aspect_example_templates} lists one representative template prompt per cultural aspect, illustrating how each aspect is operationalized across the 100 language-agnostic templates.


\begin{table*}[ht!]
\centering
\scriptsize
\setlength{\tabcolsep}{3pt}
\renewcommand{\arraystretch}{1.12}

\begin{tabular}{p{2.6cm} p{5.2cm} || p{2.6cm} p{5.2cm}}
\toprule
\multicolumn{2}{c||}{\textbf{Cultural Aspect: Example Templates (I)}} &
\multicolumn{2}{c}{\textbf{Cultural Aspect: Example Templates (II)}} \\
\midrule
\textbf{Aspect} & \textbf{Template prompt} & \textbf{Aspect} & \textbf{Template prompt} \\
\midrule

Food and Cuisine &
\texttt{Which dish would [FAMOUS PERSON] probably not recognize from their childhood?}
&
Sports &
\texttt{In [CULTURE SPORT] tradition, what happens when [CONDITION]?}
\\[-1pt]
\cmidrule(lr){1-2}\cmidrule(lr){3-4}

Music and Art &
\texttt{Which traditional musical instrument from [COUNTRY/REGION] has the earliest recorded history?}
&
Fashion and Media &
\texttt{I couldn't stop laughing watching a/an [CULTURE/REGION] series with [ACTOR NAME]. Which of the following is most likely the name of the series?}
\\[-1pt]
\cmidrule(lr){1-2}\cmidrule(lr){3-4}

Cities and Landmarks &
\texttt{Among all the provinces in [SET OF PROVINCES/LOCATION], how many provinces have an area smaller than [PROVINCE]?}
&
Transportation &
\texttt{If I live in [LOCATION 1/RESIDENTIAL AREA 1] and I want to go to [LOCATION 2/RESIDENTIAL AREA 2], how much time would it take on average if I traveled by [TRANSPORTATION METHOD]?}
\\[-1pt]
\cmidrule(lr){1-2}\cmidrule(lr){3-4}

Famous People &
\texttt{Who among these [NATIONALITY] [FAMOUS PEOPLE TYPE] does NOT share the key trait of [COMMON TRAIT]?}
&
Education &
\texttt{Which [ACADEMIC PERIOD] would a [AGE]-year-old typically be in according to [COUNTRY]'s education system?}
\\[-1pt]
\cmidrule(lr){1-2}\cmidrule(lr){3-4}

Politics and Governance &
\texttt{What is the name of the first child of the [Nth] president/leader of [COUNTRY]?}
&
Agriculture &
\texttt{In [REGION] during [MONTH], which crop is typically being [AGRICULTURAL ACTIVITY]?}
\\[-1pt]
\cmidrule(lr){1-2}\cmidrule(lr){3-4}

Events and Festivals &
\texttt{Which of the following special days is the closest to [EVENT]?}
&
Naming &
\texttt{If my friend has the last name "[LAST NAME]", which country is most likely their birthplace?}
\\[-1pt]
\cmidrule(lr){1-2}\cmidrule(lr){3-4}

Objects and Units &
\texttt{In [CULTURE] traditional measurements, how many [UNIT] equal one [LARGER UNIT]?}
&
Folklore and Folktales &
\texttt{In [CULTURE]'s folk tales, which character would be considered out of place if it appears alongside [CHARACTER]?}
\\[-1pt]
\cmidrule(lr){1-2}\cmidrule(lr){3-4}

Socio-religious Aspects of Life &
\texttt{According to [NATIONALITY] cultural superstition, what should one do after [ACTION] to avoid bad luck?}
&
Brands and Commerce &
\texttt{Among these local brands in [CULTURE/REGION], which one would a typical middle-income person be most likely to use?}
\\[-1pt]
\cmidrule(lr){1-2}\cmidrule(lr){3-4}

Language and Communication &
\texttt{In [COUNTRY], which age group or social category is LEAST likely to use the expression "[COMMON PHRASE]" to describe [MEANING OF EXPRESSION TO THEM]?}
&
Death and Funerals &
\texttt{In [CULTURE/REGION], how many days after death is [RITUAL/EVENT] traditionally performed?}
\\[-1pt]
\cmidrule(lr){1-2}\cmidrule(lr){3-4}

Social Customs &
\texttt{In [COUNTRY/REGION], when [CONDITION/ACTIVITIES], which of the following actions is considered a taboo?}
&
Time &
\texttt{If you convert [DATE] in the [CALENDAR SYSTEM] calendar to the Gregorian calendar in [YEAR], which month would it fall in?}
\\[-1pt]
\cmidrule(lr){1-2}\cmidrule(lr){3-4}

Relationships &
\texttt{If I call my [NATIONALITY] father [TERM], would my children call him [OPTION A]?}
&
Literature and Written works &
\texttt{What age-appropriate book can I buy for my [AGE]-year-old son?}
\\

\bottomrule
\end{tabular}
\caption{Cultural aspects covered in the benchmark, with one example template per aspect.}
\label{tab:cultural_aspect_example_templates}
\end{table*}

\section{Detailed Benchmark Results}
\label{app:detailed-benchmark}

This section reports accuracy on \emph{local-language} evaluation instances (MC--L, T--L, F--L), aggregated either (i) by writing system (macro-averaging across languages that share a script) or (ii) by benchmark language/cultural context.

\begin{table*}[t]
\centering
\scriptsize
\setlength{\tabcolsep}{3pt}
\renewcommand{\arraystretch}{1.05}
\resizebox{\textwidth}{!}{%
\begin{tabular}{p{2.4cm}p{4cm}rrrrrrrrrr}
\toprule
\textbf{Category} & \textbf{Model} &
\textbf{Arabic} & \textbf{Cyrillic} & \textbf{Devanagari} & \textbf{Georgian} &
\textbf{Ge'ez} & \textbf{Greek} & \textbf{Han} & \textbf{Japanese} & \textbf{Latin} & \textbf{Thai} \\
\midrule
\multirow{4}{*}{\textbf{Closed (thinking)}}
& Gemini 3 Flash Preview (thinking) & 88.7\% & 85.0\% & 97.0\% & 90.4\% & 87.5\% & 88.5\% & 89.7\% & 85.9\% & 90.0\% & 87.9\% \\
& Gemini 2.5 Pro (thinking)         & 86.3\% & 85.0\% & 95.5\% & 90.9\% & 80.5\% & 84.0\% & 87.1\% & 85.5\% & 90.0\% & 83.8\% \\
& DeepSeek V3.1 (thinking)          & 67.9\% & 67.5\% & 94.0\% & 76.3\% & 55.5\% & 82.5\% & 85.1\% & 79.3\% & 79.0\% & 75.3\% \\
& Gemini 2.5 Flash (thinking)       & 67.4\% & 67.5\% & 76.5\% & 60.1\% & 70.5\% & 64.0\% & 70.1\% & 76.3\% & 66.8\% & 71.2\% \\
\multicolumn{2}{l}{\textit{Average (Closed thinking)}}
& 77.6\% & 76.2\% & 90.8\% & 79.4\% & 73.5\% & 79.8\% & 83.0\% & 81.8\% & 81.4\% & 79.6\% \\
\midrule
\multirow{7}{*}{\textbf{Closed (standard)}}
& Gemini 3 Flash Preview & 85.1\% & 82.0\% & 96.5\% & 91.4\% & 81.5\% & 88.0\% & 88.1\% & 84.8\% & 87.2\% & 88.9\% \\
& Claude Opus 4.5        & 75.6\% & 74.0\% & 94.0\% & 80.3\% & 67.0\% & 81.5\% & 87.1\% & 82.3\% & 83.9\% & 72.9\% \\
& GPT-5 Chat             & 75.9\% & 72.5\% & 89.5\% & 79.3\% & 50.0\% & 81.5\% & 86.1\% & 80.8\% & 81.3\% & 79.3\% \\
& Gemini 2.5 Flash       & 75.1\% & 71.0\% & 95.5\% & 81.8\% & 64.5\% & 78.5\% & 86.6\% & 83.8\% & 82.5\% & 78.3\% \\
& DeepSeek V3.1          & 61.3\% & 60.5\% & 87.5\% & 65.7\% & 46.5\% & 71.5\% & 84.5\% & 78.3\% & 76.6\% & 73.2\% \\
& Claude Haiku 4.5       & 62.8\% & 57.5\% & 85.5\% & 64.1\% & 39.0\% & 71.0\% & 83.0\% & 77.8\% & 70.1\% & 59.1\% \\
& GPT-4o mini            & 64.9\% & 58.5\% & 81.5\% & 64.6\% & 38.0\% & 67.0\% & 79.9\% & 76.8\% & 70.7\% & 68.2\% \\
\multicolumn{2}{l}{\textit{Average (Closed standard)}}
& 71.5\% & 68.0\% & 90.0\% & 75.3\% & 55.2\% & 77.0\% & 85.0\% & 80.7\% & 78.9\% & 74.3\% \\
\midrule
\multirow{10}{*}{\textbf{Open-weight}}
& Qwen3-235B     & 67.2\% & 58.0\% & 85.5\% & 64.1\% & 45.0\% & 72.0\% & 84.0\% & 80.3\% & 73.4\% & 70.7\% \\
& Llama 3.3 70B  & 59.0\% & 60.5\% & 79.5\% & 58.6\% & 41.5\% & 73.5\% & 79.9\% & 78.3\% & 69.7\% & 62.6\% \\
& Llama 4 Maverick & 54.4\% & 61.0\% & 84.5\% & 66.7\% & 49.5\% & 70.0\% & 80.9\% & 67.6\% & 70.0\% & 71.7\% \\
& Llama 3.1 8B   & 42.6\% & 45.5\% & 61.5\% & 44.4\% & 29.0\% & 55.0\% & 65.5\% & 58.1\% & 50.1\% & 47.0\% \\
& Qwen3-4B       & 41.7\% & 36.5\% & 63.0\% & 42.4\% & 26.5\% & 52.0\% & 75.3\% & 62.1\% & 51.0\% & 50.5\% \\
& InternLM3-8B   & 42.0\% & 34.0\% & 53.0\% & 39.9\% & 36.0\% & 44.5\% & 77.8\% & 58.1\% & 49.8\% & 50.0\% \\
& Qwen2.5-7B     & 47.5\% & 42.0\% & 56.5\% & 40.4\% & 34.5\% & 53.0\% & 73.2\% & 70.7\% & 53.3\% & 53.0\% \\
& Aya Expanse 8B & 47.4\% & 36.5\% & 66.5\% & 39.4\% & 31.5\% & 60.5\% & 59.8\% & 62.1\% & 53.2\% & 43.4\% \\
& Llama 3.2 3B   & 35.3\% & 31.5\% & 52.5\% & 31.3\% & 37.5\% & 46.5\% & 51.5\% & 49.0\% & 44.8\% & 38.4\% \\
& Aya 23 8B      & 41.1\% & 33.5\% & 50.0\% & 31.8\% & 24.0\% & 46.5\% & 54.1\% & 53.0\% & 42.0\% & 37.4\% \\
\multicolumn{2}{l}{\textit{Average (Open-weight)}}
& 47.8\% & 43.9\% & 65.2\% & 45.9\% & 35.5\% & 57.4\% & 70.2\% & 63.9\% & 55.7\% & 52.5\% \\
\midrule
\multicolumn{2}{l}{\textbf{Average (All)}}
& 61.4\% & 58.1\% & 78.4\% & 62.1\% & 49.3\% & 68.2\% & 77.6\% & 72.9\% & 68.4\% & 64.9\% \\
\bottomrule
\end{tabular}%
}
\caption{Local-language accuracy (\%) by writing system. Each cell reports a model's overall accuracy on local-language evaluation instances (MC--L, T--L, F--L) aggregated over all benchmark contexts sharing a given script. For scripts used by multiple languages (e.g., Arabic, Latin), scores are macro-averaged across those languages. \textit{Average} rows report the mean over models within each group; \textbf{Average (All)} averages over all models.}
\label{tab:script_results}
\end{table*}

\begin{table*}[t]
\centering
\scriptsize
\setlength{\tabcolsep}{2pt}
\renewcommand{\arraystretch}{1.05}
\resizebox{\textwidth}{!}{%
\begin{tabular}{p{2.4cm}p{3.5cm}*{20}{r}}
\toprule
\textbf{Category} & \textbf{Model} &
\textbf{Amh.} & \textbf{Pt-BR} & \textbf{Zho.} & \textbf{Egy-Ar} &
\textbf{Geo.} & \textbf{Gre.} & \textbf{Hin.} & \textbf{Ind.} & \textbf{Ita.} &
\textbf{Jpn.} & \textbf{Kyr.} & \textbf{Mx-Es} & \textbf{Mor-Ar} &
\textbf{Tgl.} & \textbf{Tha.} & \textbf{Tun-Ar} & \textbf{Tur.} &
\textbf{Yem-Ar} & \textbf{Yor.} & \textbf{Zul.} \\
\midrule
\multirow{4}{*}{\textbf{Closed (thinking)}}
& Gemini 3 Flash Preview (thinking) & 87.5\% & 94.0\% & 89.7\% & 92.9\% & 90.4\% & 88.5\% & 97.0\% & 93.2\% & 90.3\% & 85.9\% & 85.0\% & 91.9\% & 90.5\% & 88.9\% & 87.9\% & 83.0\% & 94.0\% & 88.5\% & 88.9\% & 78.5\% \\
& Gemini 2.5 Pro (thinking)         & 80.5\% & 93.3\% & 87.1\% & 89.9\% & 90.9\% & 84.0\% & 95.5\% & 90.8\% & 89.3\% & 85.5\% & 85.0\% & 90.4\% & 88.0\% & 88.4\% & 83.8\% & 80.0\% & 94.0\% & 87.5\% & 83.7\% & 84.5 \\
& DeepSeek V3.1 (thinking)          & 55.5\% & 88.5\% & 85.1\% & 73.2\% & 76.3\% & 82.5\% & 94.0\% & 84.2\% & 69.4\% & 79.3\% & 67.5\% & 84.3\% & 71.0\% & 79.3\% & 75.3\% & 52.5\% & 85.0\% & 75.0\% & 75.3\% & 66.0\% \\
& Gemini 2.5 Flash (thinking)       & 70.5\% & 65.0\% & 70.1\% & 66.2\% & 60.1\% & 64.0\% & 76.5\% & 69.5\% & 62.8\% & 76.3\% & 67.5\% & 73.2\% & 69.5\% & 66.7\% & 71.2\% & 65.0\% & 77.5\% & 69.0\% & 59.5\% & 60.5\% \\
\multicolumn{2}{l}{\textit{Average (Closed thinking)}} &
73.5\% & 85.2\% & 83.0\% & 80.5\% & 79.4\% & 79.8\% & 90.8\% & 84.4\% & 78.0\% & 81.8\% & 76.2\% & 85.0\% & 79.8\% & 80.8\% & 79.6\% & 70.1\% & 87.6\% & 80.0\% & 76.8\% & 68.3\% \\
\midrule
\multirow{7}{*}{\textbf{Closed (standard)}}
& Gemini 3 Flash Preview & 81.5\% & 90.0\% & 88.1\% & 83.8\% & 91.4\% & 88.0\% & 96.5\% & 94.2\% & 88.3\% & 84.8\% & 82.0\% & 87.4\% & 88.0\% & 84.8\% & 88.9\% & 82.0\% & 90.5\% & 86.5\% & 83.2\% & 79.0\% \\
& Claude Opus 4.5        & 67.0\% & 89.6\% & 87.1\% & 71.1\% & 80.3\% & 81.5\% & 94.0\% & 91.6\% & 87.2\% & 82.3\% & 74.0\% & 86.9\% & 75.3\% & 78.8\% & 72.9\% & 76.5\% & 86.0\% & 79.6\% & 76.1\% & 75.0\% \\
& GPT-5 Chat             & 50.0\% & 87.0\% & 86.1\% & 74.7\% & 79.3\% & 81.5\% & 89.5\% & 81.6\% & 82.7\% & 80.8\% & 72.5\% & 83.3\% & 77.0\% & 79.8\% & 79.3\% & 74.0\% & 83.5\% & 78.0\% & 78.9\% & 73.5\% \\
& Gemini 2.5 Flash       & 64.5\% & 83.5\% & 86.6\% & 74.2\% & 81.8\% & 78.5\% & 95.5\% & 86.8\% & 89.8\% & 83.8\% & 71.0\% & 86.4\% & 79.5\% & 80.3\% & 78.3\% & 76.5\% & 84.0\% & 70.0\% & 78.4\% & 71.0\% \\
& DeepSeek V3.1          & 46.5\% & 84.5\% & 84.5\% & 59.1\% & 65.7\% & 71.5\% & 87.5\% & 77.4\% & 86.7\% & 78.3\% & 60.5\% & 77.3\% & 64.0\% & 75.8\% & 73.2\% & 56.5\% & 79.0\% & 65.5\% & 70.0\% & 62.0\% \\
& Claude Haiku 4.5       & 39.0\% & 77.5\% & 83.0\% & 62.1\% & 64.1\% & 71.0\% & 85.5\% & 68.9\% & 81.1\% & 77.8\% & 57.5\% & 71.2\% & 64.0\% & 69.7\% & 59.1\% & 57.0\% & 72.5\% & 68.0\% & 61.8\% & 58.0\% \\
& GPT-4o mini            & 38.0\% & 75.5\% & 79.9\% & 63.6\% & 64.6\% & 67.0\% & 81.5\% & 71.1\% & 82.1\% & 76.8\% & 58.5\% & 72.7\% & 66.0\% & 70.2\% & 68.2\% & 62.5\% & 75.0\% & 67.5\% & 60.5\% & 58.5\% \\
\multicolumn{2}{l}{\textit{Average (Closed standard)}} &
55.2\% & 83.9\% & 85.0\% & 69.8\% & 75.3\% & 77.0\% & 90.0\% & 81.7\% & 85.4\% & 80.7\% & 68.0\% & 80.7\% & 73.4\% & 77.1\% & 74.3\% & 69.3\% & 81.5\% & 73.6\% & 72.7\% & 68.1\% \\
\midrule
\multirow{10}{*}{\textbf{Open-weight}}
& Qwen3-235B     & 45.0\% & 77.5\% & 84.0\% & 67.7\% & 64.1\% & 72.0\% & 85.5\% & 75.8\% & 80.6\% & 80.3\% & 58.0\% & 79.8\% & 62.5\% & 76.3\% & 70.7\% & 67.0\% & 75.5\% & 71.5\% & 62.6\% & 59.0\% \\
& Llama 3.3 70B  & 41.5\% & 71.5\% & 79.9\% & 59.1\% & 58.6\% & 73.5\% & 79.5\% & 73.7\% & 80.6\% & 78.3\% & 60.5\% & 70.2\% & 60.0\% & 72.2\% & 62.6\% & 54.5\% & 72.5\% & 62.5\% & 61.1\% & 56.0\% \\
& Llama 4 Maverick & 49.5\% & 75.0\% & 80.9\% & 55.6\% & 66.7\% & 70.0\% & 84.5\% & 70.8\% & 78.1\% & 67.6\% & 61.0\% & 70.7\% & 65.5\% & 74.7\% & 71.7\% & 61.5\% & 71.0\% & 35.0\% & 56.8\% & 62.5\% \\
& Llama 3.1 8B   & 29.0\% & 42.5\% & 65.5\% & 40.4\% & 44.4\% & 55.0\% & 61.5\% & 52.6\% & 60.7\% & 58.1\% & 45.5\% & 54.5\% & 44.5\% & 46.5\% & 47.0\% & 41.5\% & 55.5\% & 44.0\% & 45.3\% & 43.0\% \\
& Qwen3-4B       & 26.5\% & 64.0\% & 75.3\% & 43.4\% & 42.4\% & 52.0\% & 63.0\% & 44.7\% & 64.3\% & 62.1\% & 36.5\% & 52.5\% & 46.0\% & 45.5\% & 50.5\% & 40.0\% & 48.0\% & 37.5\% & 43.2\% & 46.0\% \\
& InternLM3-8B   & 36.0\% & 46.5\% & 77.8\% & 33.8\% & 39.9\% & 44.5\% & 53.0\% & 47.9\% & 63.3\% & 58.1\% & 34.0\% & 57.1\% & 46.0\% & 49.0\% & 50.0\% & 45.0\% & 46.0\% & 43.0\% & 45.8\% & 42.5\% \\
& Qwen2.5-7B     & 34.5\% & 62.5\% & 73.2\% & 50.0\% & 40.4\% & 53.0\% & 56.5\% & 47.9\% & 62.2\% & 70.7\% & 42.0\% & 61.6\% & 48.0\% & 51.5\% & 53.0\% & 45.5\% & 50.0\% & 46.5\% & 46.8\% & 43.5\% \\
& Aya Expanse 8B & 31.5\% & 60.0\% & 59.8\% & 44.4\% & 39.4\% & 60.5\% & 66.5\% & 51.6\% & 66.8\% & 62.1\% & 36.5\% & 55.1\% & 50.0\% & 50.0\% & 43.4\% & 45.0\% & 55.5\% & 50.0\% & 45.3\% & 41.5\% \\
& Llama 3.2 3B   & 37.5\% & 44.0\% & 51.5\% & 36.9\% & 31.3\% & 46.5\% & 52.5\% & 42.1\% & 54.1\% & 49.0\% & 31.5\% & 48.0\% & 36.5\% & 46.0\% & 38.4\% & 32.5\% & 44.5\% & 35.5\% & 37.4\% & 42.5\% \\
& Aya 23 8B      & 24.0\% & 45.5\% & 54.1\% & 34.8\% & 31.8\% & 46.5\% & 50.0\% & 42.6\% & 54.1\% & 53.0\% & 33.5\% & 54.5\% & 43.5\% & 37.4\% & 37.4\% & 42.0\% & 40.5\% & 44.0\% & 35.8\% & 25.5\% \\
\multicolumn{2}{l}{\textit{Average (Open-weight)}} &
35.5\% & 58.9\% & 70.2\% & 46.6\% & 45.9\% & 57.4\% & 65.2\% & 55.0\% & 66.5\% & 63.9\% & 43.9\% & 60.4\% & 50.3\% & 54.9\% & 52.5\% & 47.5\% & 55.9\% & 46.9\% & 48.0\% & 46.2\% \\
\midrule
\multicolumn{2}{l}{\textbf{Average (All)}} &
49.3\% & 72.3\% & 77.6\% & 60.8\% & 62.1\% & 67.2\% & 78.4\% & 68.4\% & 73.0\% & 72.9\% & 57.2\% & 71.7\% & 61.9\% & 65.4\% & 62.7\% & 59.0\% & 70.5\% & 62.1\% & 61.7\% & 57.2\% \\
\bottomrule
\end{tabular}%
}
\caption{Local-language accuracy (\%) by benchmark language/cultural context. Each cell reports a model's overall accuracy on local-language instances (MC--L, T--L, F--L) for the corresponding context, aggregated over all templates. Missing entries ({--}) are ignored in averages. \textit{Average} rows report the mean over models within each group; \textbf{Average (All)} averages over all models. Column headers are abbreviated: Amh.=Amharic, Pt-BR=Brazilian Portuguese, Zho.=Chinese, Egy-Ar=Egyptian Arabic, Geo.=Georgian, Gre.=Greek, Hin.=Hindi, Ind.=Indonesian, Ita.=Italian, Jpn.=Japanese, Kyr.=Kyrgyz, Mx-Es=Mexican Spanish, Mor-Ar=Moroccan Arabic, Tgl.=Tagalog, Tha.=Thai, Tun-Ar=Tunisian Arabic, Tur.=Turkish, Yem-Ar=Yemeni Arabic, Yor.=Yoruba, Zul.=Zulu.}
\label{tab:language_results}
\end{table*}

\section{Annotation Platform}
\label{app:annotation}

\begin{figure*}[t]
  \centering
  \includegraphics[width=\textwidth,height=0.8\textheight,keepaspectratio]{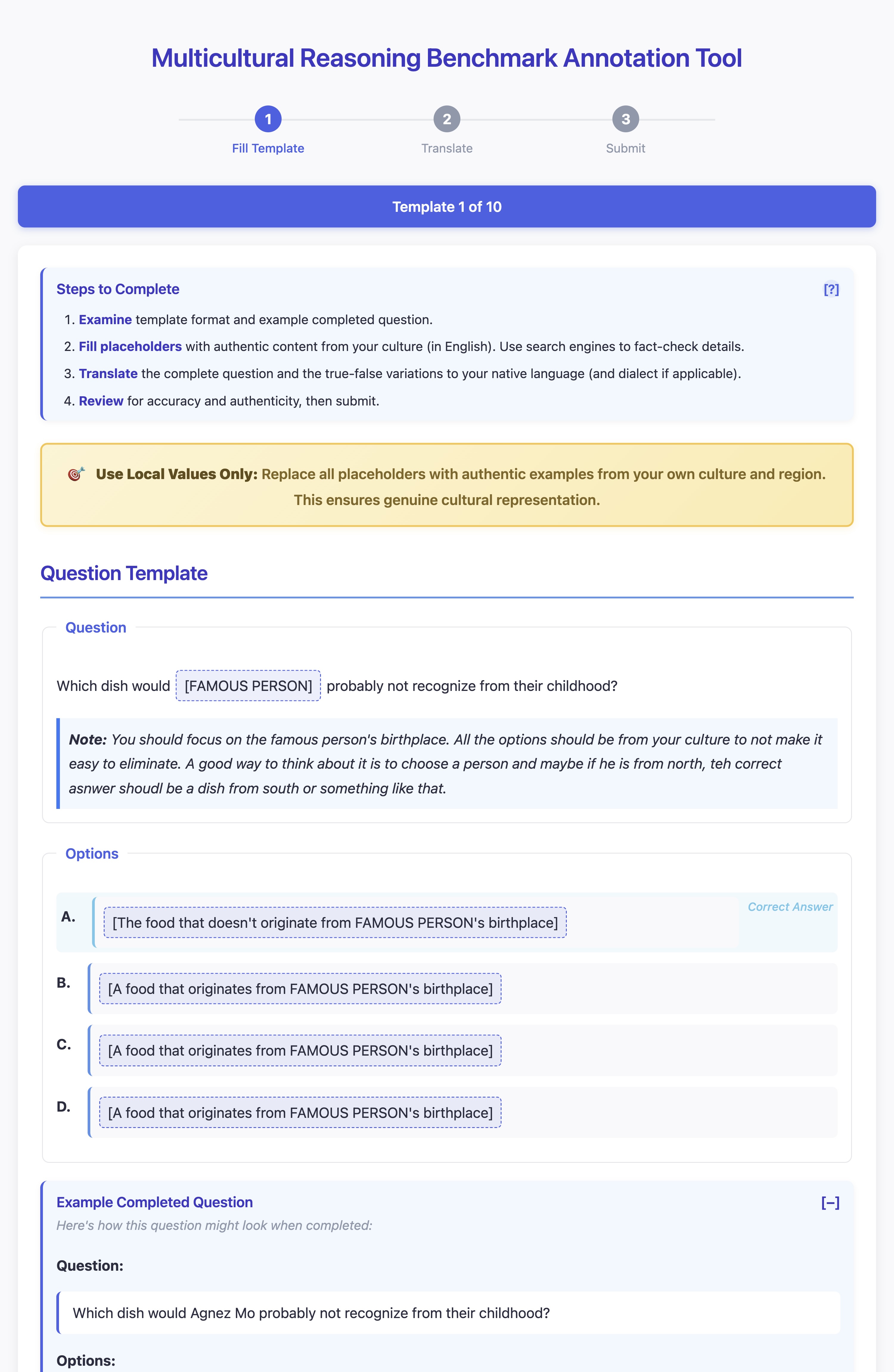}
  \caption{Annotation interface (Part 1 of 2): \textbf{Contextualization}.}
  \label{fig:platform_tool_top}
\end{figure*}

\begin{figure*}[t]
  \centering
  \includegraphics[width=\textwidth,height=0.8\textheight,keepaspectratio]{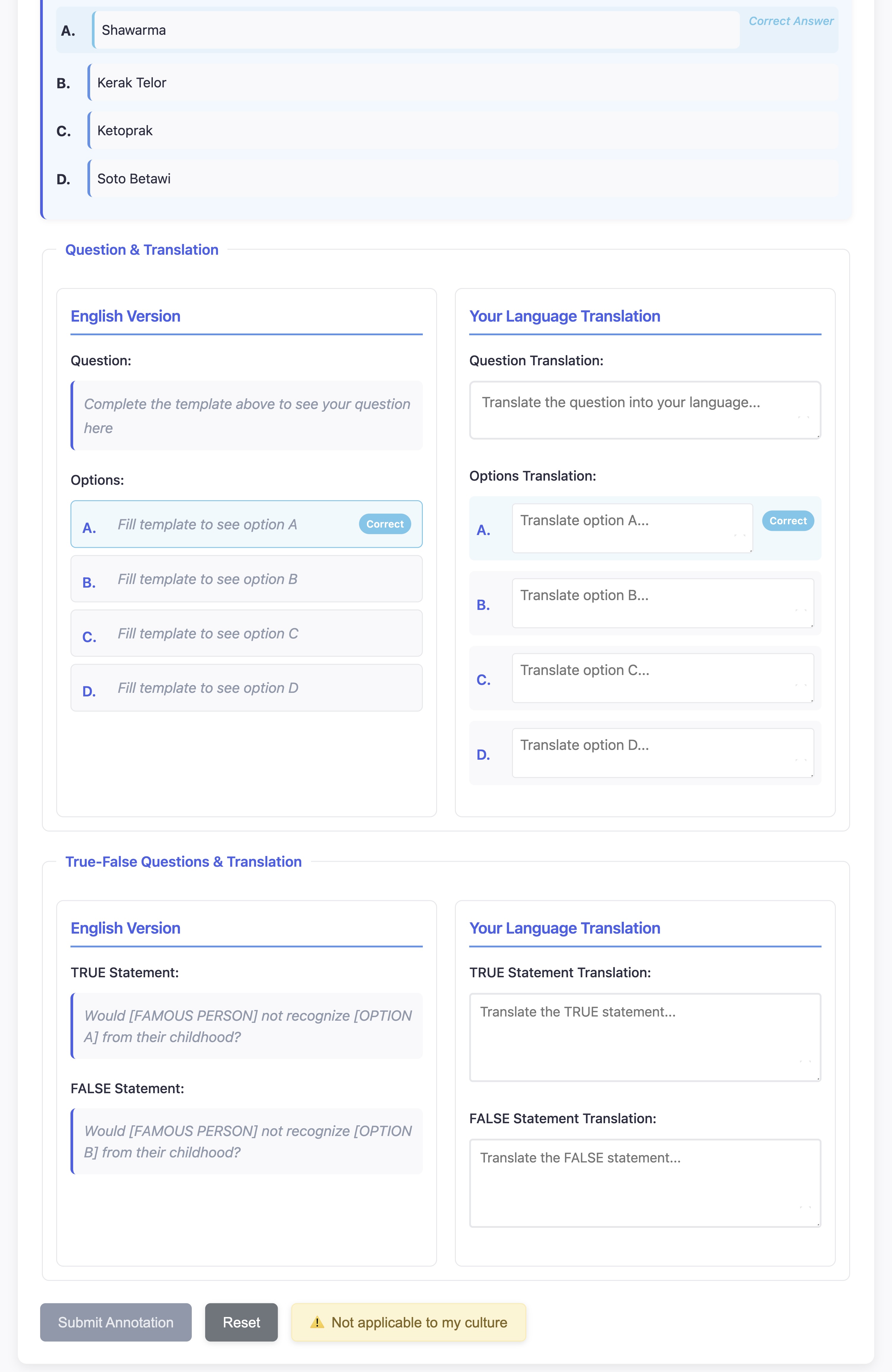}
  \caption{Annotation interface (Part 2 of 2): \textbf{Instantiation and translation}.}
  \label{fig:platform_tool_bottom}
\end{figure*}

Figure~\ref{fig:platform_home} shows the onboarding flow presented to annotators, including requirements for cultural authenticity and guidance for handling culturally specific items. Figure~\ref{fig:platform_tool_top} and Figure~\ref{fig:platform_tool_bottom} show the main annotation interface used to instantiate templates in English, translate them into the local language, and generate verification statements.

\begin{figure*}[t]
  \centering
  \includegraphics[width=\textwidth,height=0.8\textheight,keepaspectratio]{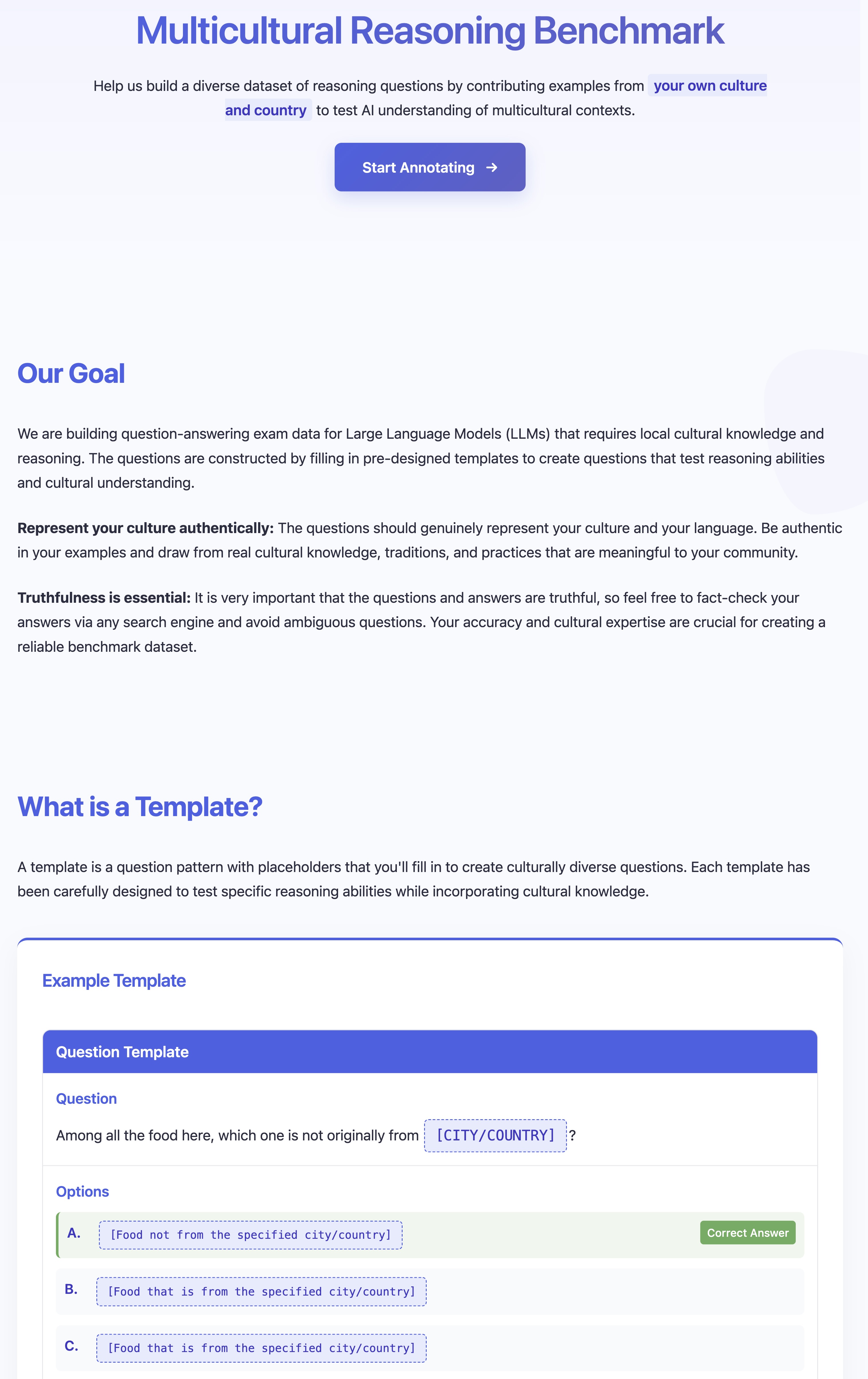}
  \caption{Onboarding landing page (Part 1 of 3).}
  \label{fig:platform_home}
\end{figure*}

\begin{figure*}[t]\ContinuedFloat
  \centering
  \includegraphics[width=\textwidth,height=0.8\textheight,keepaspectratio]{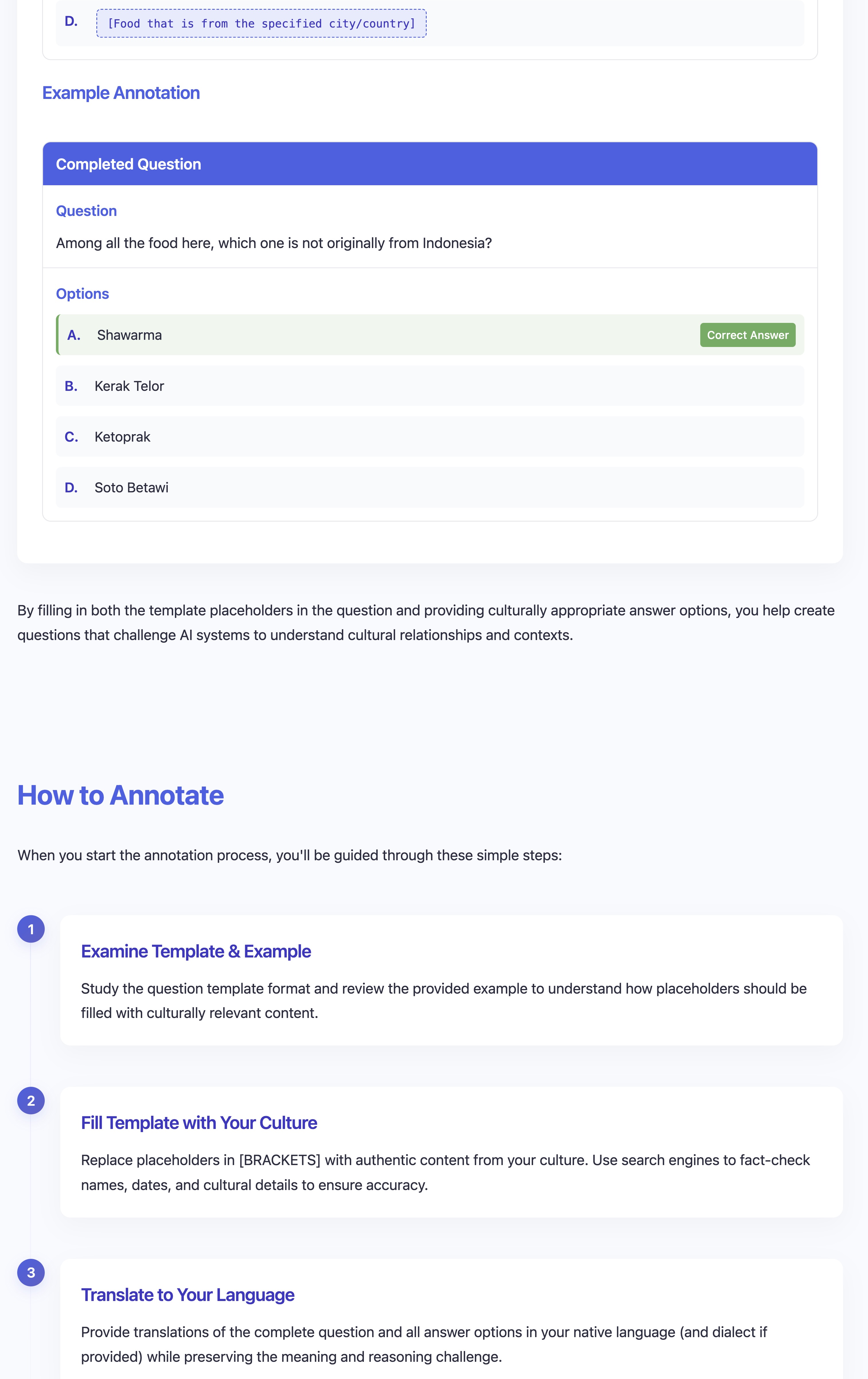}
  \caption{Onboarding landing page (Part 2 of 3).}
\end{figure*}

\begin{figure*}[t]\ContinuedFloat
  \centering
  \includegraphics[width=\textwidth,height=0.8\textheight,keepaspectratio]{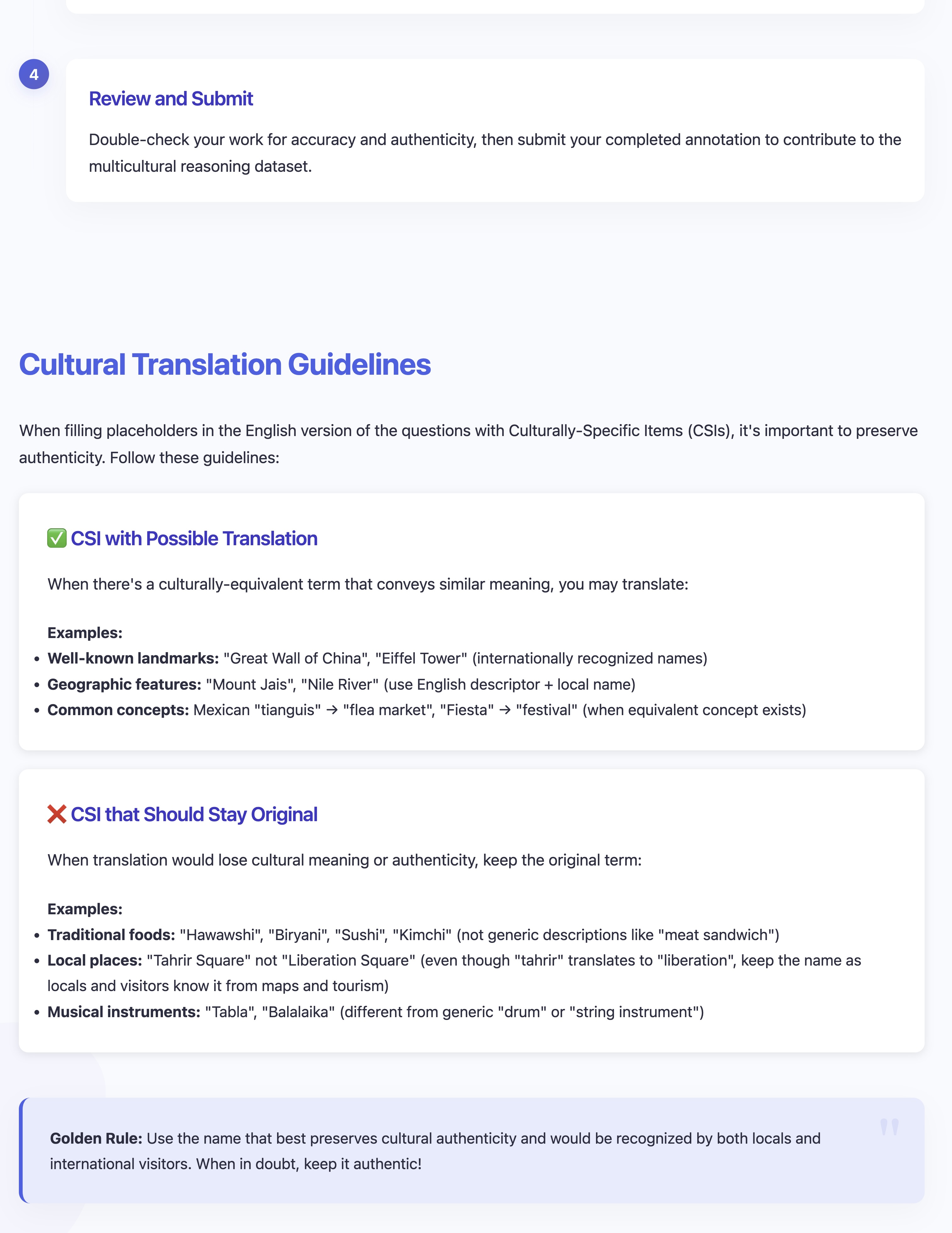}
  \caption{Onboarding landing page (Part 3 of 3).}
\end{figure*}
\clearpage

\end{document}